\pdfoutput=1
\documentclass[journal]{IEEEtran}
\usepackage{amsmath,amssymb}
\usepackage{algorithm}
\usepackage{algorithmic}
\usepackage{booktabs}
\usepackage{array}
\usepackage{graphicx}
\usepackage{tikz}
\usetikzlibrary{arrows.meta,positioning,calc}
\usepackage[numbers,sort&compress]{natbib}
\usepackage{bibunits}  %
\makeatletter\@namedef{b@IEEEexample:BSTcontrol}{{}{}{{}}{{}}}\makeatother%
\usepackage{url}
\defaultbibliographystyle{IEEEtranN}
\defaultbibliography{paper}

\newcommand{\csv}{\textsc{csv}}

\newcommand{\Tref}{T_{\mathrm{ref}}}
\newcommand{\esign}{\epsilon_{\mathrm{sign}}}
\newcommand{\Pst}{P\!\left(\mathrm{SS}(C)\mid I(T)\right)}
\newcommand{\Trem}{\tau_{\mathrm{rem}}}
\newcommand{\Treint}{\tau_{\mathrm{int}}}

\makeatletter
\def\subsubsection{\@startsection{subsubsection}{3}{\parindent}{0ex plus 0.1ex minus 0.1ex}%
{0ex}{\normalfont\normalsize\bfseries}}
\def\paragraph{\@startsection{paragraph}{4}{2\parindent}{0ex plus 0.1ex minus 0.1ex}%
{0ex}{\normalfont\normalsize\bfseries}}
\def\@IEEEsectpunct{\ }
\makeatother

\newcommand{\startsupplement}[1]{%
  \clearpage
  \setcounter{page}{1}\renewcommand{\thepage}{S\arabic{page}}%
  \setcounter{section}{0}\renewcommand{\thesection}{S\arabic{section}}%
  \setcounter{table}{0}\renewcommand{\thetable}{S\arabic{table}}%
  \setcounter{figure}{0}\renewcommand{\thefigure}{S\arabic{figure}}%
  \setcounter{equation}{0}\renewcommand{\theequation}{S\arabic{equation}}%
  \setcounter{algorithm}{0}\renewcommand{\thealgorithm}{S\arabic{algorithm}}%
  \setcounter{footnote}{0}%
  \twocolumn[{\centering{\Large Supplementary Material for}\\[0.5em]{\large ``#1''}\\[1.5em]}]}

\begin{document}

\title{Multi-Scale Structural Features for Continual, Comprehensible\\Visual Recognition in a Developmental Learning Framework}

\author{Zeki~Doruk~Erden\\Sabanc{\i} University\\doruk.erden@sabanciuniv.edu\vspace{-1.2\baselineskip}%
\thanks{This work has been submitted to the IEEE for possible publication. Copyright may be transferred without notice, after which this version may no longer be accessible. Code will be released publicly upon publication.}}

\markboth{}{}

\maketitle

\begin{abstract}
Machine learning struggles to learn continually, to reuse prior knowledge, and to expose comprehensible internal structure. A developmental, gradient-free framework addresses these by learning a discrete, topological model of its inputs through local variation and selection, with an inherent continual-learning guarantee: new observations refine existing structure without overwriting past knowledge, and without replay or task boundaries. An earlier visual extension demonstrated the principle but was limited by its feature representation. We extend this developmental design with a multi-scale structural representation of shape---contour change points and their cross-scale spatial relations in a single network---integrated with the developmental learning process so as to improve both the learning dynamics and the read-out, while preserving the learner's continual-learning properties. On class-incremental MNIST and the harder EMNIST-Letters, the system learns each class with no replay and no task boundaries, matching or exceeding every replay-free neural baseline; its internal representations are inherently comprehensible. What sets it apart is retention. Across both benchmarks, a previously seen class survives the learning of others far better than in regularisation-based networks---on MNIST, about as well as with an 80-image replay buffer, without storing one. When regular re-exposure is withdrawn, the regularisation baselines collapse toward chance, exposed as relearning on contact, while our system degrades gracefully, keeping most of what it learned; on EMNIST-Letters the separation is larger still. The result is a more accurate visual extension of a developmental learning approach, decisively superior to replay-free baselines in retention and inherently comprehensible.
\end{abstract}

\begin{IEEEkeywords}
Continual learning, developmental learning, structural representation, computer vision, interpretability.
\end{IEEEkeywords}

\begin{bibunit}
\nocite{IEEEexample:BSTcontrol}%

\section{Introduction} \label{sec:intro}

\IEEEPARstart{C}{ontinual} learning asks a system to learn from a non-stationary stream without overwriting what it already knows. Gradient-trained networks struggle by construction: past and present competence are carried by overlapping parameters in a shared weight space, and an update reducing error on the current data carries no constraint to preserve those supporting past competence \citep{mccloskey1989catastrophic,vandeven2024continual}. We call this \emph{destructive adaptation}.\footnote{We prefer this to ``catastrophic forgetting,'' which connotes an unintended, passive loss; the process is an \emph{active overwriting} of past information by the update on the current data.} The standard remedies---replay buffers, regularization anchors, task boundaries---mitigate the symptom while conceding the premise, each importing an assumption sequential experience does not offer: retained data, known boundaries, or a routing signal.

A different design avoids the problem at its root. A recently proposed \emph{developmental}, gradient-free framework grows a discrete, topological model of its observations by local variation and selection, so integrating an observation adds or refines structure locally and---under a formal condition---leaves the responses to all past observations locally unchanged \citep{erden2025foundations,erden2025thesis}. That method, termed the \emph{Modeller} (an instance of \emph{variation and selection}, or \emph{varsel}, networks---a proposed class of learning systems \citep{erden2025thesis}), integrates each observation exactly once, stores no past data, needs no task boundaries, and yields models that are readily and inherently comprehensible---a \emph{property set} statistical learning does not provide at any task scale. This reflects a \emph{developmental} view of learning---competence accumulated and retained through local variation and selection over experience, grown rather than fitted and overwritten \citep{weng2001autonomous}---and, for vision, a \emph{structural} one, on which recognition rests on explicit structural descriptions rather than stored appearances \citep{marr1978representation,biederman1987recognition}; both are long-standing positions in the study of natural cognition that gradient-trained systems set aside. Its extension to vision reconceives an observation as a relational network and learns by refining it, but the representation it used was minimal and capped accuracy \citep{erden2025foundations}.

This paper advances the visual extension of this developmental learning method, in both its representation and its learning process, while keeping its learning principle and preservation guarantee intact. Our central contribution is a \emph{multi-scale structural feature representation}: an image becomes a single relational network describing the shape at \emph{every} level of a coarsening hierarchy at once, so long-range and local relations are present together over shared nodes. This matters twice over. It gives refinement material at more than one granularity, so whichever scale recurs across a class's instances is retained while the rest is reduced away---structure accumulates rather than eroding. And it makes matching tractable: the atomic features are generic and recur many times within and across shapes, so the matcher needs relations constraining placement over the whole shape. We also improve the \emph{read-out}: a local, gradient-free procedure consuming the spatial statistics the matcher already computes, which treats the \emph{absence} of an expected unit as evidence. The construction rests on a general principle---\emph{a continuous entity can be represented by its points of change, typed by the order and character of that change}---of which we build the simplest instantiation: first-order change points on a binarized contour, restricting the system to 2D shape. This 2D restriction comes from the step that converts images into networks, \emph{not} from the learner, which is indifferent to what its nodes and edges stand for; we adopt it deliberately so that the learning machinery can be studied on inputs whose structure is easy to extract, without having to first solve the separate, hard problem of recovering structure from more complex (e.g.\ 3D) images (Sec.~\ref{sec:representation}). Figure~\ref{fig:overview} summarizes the system as a whole: a shape image becomes one multi-scale structural network. \emph{During training}, structural learning grows a per-class model from that network, one sample at a time; \emph{at test time}, a separate, read-only read-out matches the learned model against a new observation and predicts a class. The two share the same matching step but are distinct operations---the read-out never modifies the model.

On class-incremental MNIST the system learns all ten classes from a replay-free, boundary-free stream---in a stronger sense than ``continual'' usually carries, since learning proceeds strictly sample by sample with no batching---reaching $0.87$ held-out accuracy after 20 cycles (well above the $\approx\!0.50$ previously reported \citep{erden2025foundations,erden2025thesis}), matching or exceeding every replay-free neural baseline under this class-incremental stream while storing no past observations, and producing models that can simply be drawn. The property that sets it apart is retention: it keeps what it learns where the replay-free baselines do not---about as well as an $80$-image replay buffer, without storing one---and unlike them it continues to do so when the stream's regular recurrence is withdrawn---under randomised class order, or when each class is presented once and never again, the regularisation baselines that had been level with the Modeller collapse to chance, exposed as relearning through re-exposure, while the Modeller degrades gracefully. The same pattern holds, with wider margins, on the harder EMNIST-Letters (Sec.~\ref{sec:exp-emnist}).

Our overall contribution is the development of the framework of \citet{erden2025foundations} (reviewed in Sec.~\ref{sec:method}) for visual recognition; along that path we contribute the multi-scale representation the learner refines directly (Sec.~\ref{sec:representation}), an explicit account of the matching step both learning and prediction rest on (Sec.~\ref{sec:matching}), the local absence-aware read-out (Sec.~\ref{sec:readout}), and an evaluation covering retention, neural baselines under their respective assumptions, retention under irregular and single-pass stream schedules, a multi-scale ablation, transformation sensitivity, model-size growth, a direct preservation audit and readable models (Sec.~\ref{sec:experiments}); Section~\ref{sec:related} positions the work within the broader literature.

Because this work develops an existing framework rather than introducing one from scratch, Table~\ref{tab:provenance} states explicitly which elements are inherited unchanged and which are new to this paper; Sec.~\ref{app:differences} of the Supplementary Material (SM; its sections, tables, figures and algorithms are numbered with an S) gives the mechanism-level detail.

\begin{table}[tbp]
\centering
\caption{Provenance of the components. Inherited elements carry over from the
framework of \citet{erden2025foundations,erden2025thesis} unchanged in important parts, so their assumptions
(including the local preservation guarantee) hold here as stated there; modified elements are
adapted; new elements are contributions of this paper. Detail:
Sec.~\ref{app:differences}.}
\label{tab:provenance}
\footnotesize
\setlength{\tabcolsep}{4pt}
\begin{tabular}{@{}>{\raggedright\arraybackslash}p{0.44\columnwidth} >{\raggedright\arraybackslash}p{0.52\columnwidth}@{}}
\toprule
Component & Provenance \\
\midrule
Gradient-free variation-and-selection learning & \textbf{Inherited} --- creation, refinement, removal \\
Local preservation guarantee (removal-only refinement) & \textbf{Inherited} --- Thm.~1; refinement here is strictly more conservative (Sec.~\ref{app:differences}) \\
\csv{} representation, presence statistics, reliability, $P(s\mid t)$ & \textbf{Inherited} \\
Single-pass, replay-free, boundary-free setting & \textbf{Inherited} \\
Single-target \csv{}s, externalised negatives & \textbf{Inherited} --- from the prior \emph{visual} formulation \\
Network refinement & \textbf{Modified} --- rerelation removed; continuity from the representation \\
Feature extraction from contours & \textbf{Modified} --- raw change-points, no polygonal approximation \\
Chain structure and matching & \textbf{Modified} --- composed sources with anchors; chain-aware pinning \\
Multi-scale structural representation & \textbf{New} --- Sec.~\ref{sec:representation} \\
Read-out mechanism & \textbf{New} --- Sec.~\ref{sec:readout} \\
Reintegration maintenance operation & \textbf{New} --- no counterpart in the prior work \\
\bottomrule
\end{tabular}
\end{table}

\begin{figure}[t]
\centering
\resizebox{0.80\columnwidth}{!}{%
\begin{tikzpicture}[
  x=1cm,y=1cm, font=\small,
  box/.style={draw,rounded corners=2pt,line width=0.6pt,align=center,inner sep=4pt},
  flow/.style={-{Stealth[length=5pt]},line width=0.7pt},
  lbl/.style={font=\footnotesize,inner sep=2pt},
]
\node[box,fill=gray!8,minimum width=2.5cm,minimum height=0.9cm] (img) at (1.7,3.3) {shape image};
\node[box,fill=teal!12,minimum width=4.2cm,minimum height=0.9cm] (net) at (6.6,3.3) {multi-scale\\observation network};
\node[box,fill=violet!10,minimum width=3.0cm,minimum height=0.75cm] (mod) at (2.6,1.65) {learned model};
\node[box,fill=orange!14,minimum width=2.2cm,minimum height=0.75cm] (ro) at (8.8,1.65) {read-out};
\node[box,fill=gray!12,minimum width=1.4cm,minimum height=0.7cm] (cls) at (8.8,0.5) {class};
\draw[flow] (img.east) -- (net.west) node[lbl,midway,above,sloped]{encode};
\draw[flow] (net.south) -- (mod.north) node[lbl,pos=0.52,above,sloped]{learn};
\draw[flow] (net.south) -- (ro.north) node[lbl,pos=0.52,above,sloped]{observe};
\draw[flow] (mod.east) -- (ro.west) node[lbl,midway,above,sloped]{consult};
\draw[flow] (ro.south) -- (cls.north) node[lbl,midway,right]{argmax};
\end{tikzpicture}}
\caption{Overview. A shape image is encoded into one \emph{multi-scale structural network}
(Sec.~\ref{sec:representation}). The same network drives gradient-free structural
learning---variation, refinement, differentiation, reintegration---that grows a per-class model
of conditioning state variables (\csv{}s) and their negative suppressors (Sec.~\ref{sec:method});
because refinement only ever \emph{removes} structure, responses to already-accounted observations
are preserved. A read-only read-out matches that model against a new observation and sums signed
per-\csv{} contributions to a class (Sec.~\ref{sec:readout}). Matching (Sec.~\ref{sec:matching}) is
the step shared by learning and read-out. Learning is one sample at a time, with no replay and no
task boundaries.}
\label{fig:overview}
\end{figure}

\section{A Brief Review of the Modeller} \label{sec:method}

We give a compact, self-contained account of the learning methodology, the \textit{Modeller}, as instantiated here; it follows the design of \citet{erden2025foundations}, to which we refer for the conceptual grounding, and Sec.~\ref{app:differences} records where this instantiation differs.

\subsubsection{Definition of ``learning''.} Learning here differs fundamentally from the statistical process. There, learning extracts statistics from a \emph{batch}---a gradient estimate followed by a small step, repeated over epochs. Here there is no batch and no gradient: the model sees \emph{one sample at a time} and responds with discrete, local \emph{structural} changes---creating a unit, removing part of one, spawning a more specific one---that encode what the observation revealed. Each sample is processed once and never revisited. This is not a batch procedure adapted to receive one sample at a time; continual learning is not a setting it tolerates but the setting it is native to. The aim, consequently, is not to approximate a target optimization function by continuous parameter changes, but to make structural modifications to the current model that \textit{explain} any unexplained observation, and to update the internal statistical evidence within that structure.

\subsubsection{State variables.} The model is built from \emph{state variables} (SVs), whose state is \emph{active} ($1$), \emph{inactive} ($-1$) or \emph{unobserved} ($0$). Classes are \emph{target} SVs. Internal structure consists of \emph{conditioning} SVs (\csv{}s); each carries a \emph{source} pattern extracted from the image, a single \emph{target} (a target SV or another \csv{}; multiple \csv{}s conditioning one another thereby form a hierarchical, multi-layer representation)\footnote{The single-target constraint is not in-principle necessary---in \citet{erden2025foundations} it is relaxed in some settings---but we keep it here as a simplifying design choice, since relaxing it in the visual setting would require implementational changes unrelated to the question this paper studies. The main limitation it introduces is not functional but a model larger than necessary, with redundant representations; this is discussed further with our scaling results.}, and a signed \emph{polarity}: a positive \csv{} represents a configuration whose presence supports its target's activity, a negative one a configuration accounting for its inactivity. Each \csv{} accumulates presence statistics ($n_{\mathrm{active}}, n_{\mathrm{inactive}}$), from which its \emph{reliability} $P(\text{target}\,|\,\text{source present})$ and per-upstream $P(s\,|\,t)$ derive. 

\subsubsection{Observations as networks, and refinement.} An observation is a \emph{state network}: a directed graph whose typed nodes are feature instances and whose edges are relations, several relation types kept as parallel layers (a \emph{state polynetwork}). When a target is active and no \csv{} accounts for it, a new positive \csv{} is created whose source is the observed network itself (\emph{variation}). Learning then reduces each source toward what recurs (\emph{selection}): when the source matches (above a threshold) a new observation in which the target recurs, source elements absent from that observation are removed. This \textit{refinement} operation carries the property the framework is built around. \textbf{Because it only ever \emph{removes} requirements from a source and never adds any, an observation that satisfied a \csv{} before a modification still satisfies it afterwards.} The unit's response to what it has already accounted for is preserved exactly---the \textit{local preservation} guaranteed by Theorem~1 of \citet{erden2025foundations}, from which system-wide continual learning (observed empirically) follows \textit{emergently and naturally} rather than via any mechanism added on top.

This also exposes one central perceptual challenge: source (an internal representation) and observation are two \emph{separate} networks, so the learner must establish \emph{where} the source sits inside the observation---a \textit{matching} searched for at every learning and prediction step (Secs.~\ref{sec:representation} and~\ref{sec:matching}).

\subsubsection{Differentiation and structure formation.} Refinement alone would collapse each class to its least common denominator, so the methodology also grows structure: when a \csv{} fully matches but its target contradicts it, or a partial match leaves an unexplained residual, \emph{refinement upstreams} are spawned---new \csv{}s conditioned on the existing one that carve out the residual (Algorithm~\ref{alg:refine}; Figs.~1 and~2 of~\citep{erden2025foundations} illustrate the refinement operation and upstream creation respectively), forming chains of progressively more specific \csv{}s atop more general ones. Two maintenance operations use the statistics, acting probabilistically on opposite ends of how often a conditioner is satisfied when its target occurs: \emph{removal} discards one almost never satisfied, \emph{reintegration} merges one almost always satisfied back into the downstream it conditions; and removal weights this by chain depth so deep upstreams that carve nothing are pruned more readily (Sec.~\ref{app:differences} states both bands exactly). A chain is evaluated coherently---a specific upstream assessed in the context of where its downstream matched---by the matching procedure of Sec.~\ref{sec:matching}, the same at learning and prediction time. For simplicity each \csv{} conditions a single target and negative evidence is externalized into separate negative \csv{}s  (Sec.~\ref{app:differences}).

Algorithm~\ref{alg:learn} sets out one complete learning step, composing these operations---variation, refinement, differentiation, and the two statistics-driven maintenance operations (removal and reintegration); matching throughout is chain-aware and anchored (Sec.~\ref{sec:matching}).

\begin{algorithm}[!tb]
\caption{One learning step (Modeller with network refinement)}
\label{alg:learn}
\begin{algorithmic}[1]
\REQUIRE observed multi-scale network $O$; target states $\{S_T\}$
\STATE reset each \csv{}'s per-step fields, and apply the target states $\{S_T\}$
\STATE \COMMENT{\textbf{traversal} --- upstream from the targets, taken in order of activity}
\WHILE{some \csv{} is still to be processed}
  \STATE take the next \csv{} $C$ on that path
  \STATE match $C$'s source against $O$, its anchors pinned to where its downstream
  matched (Alg.~\ref{alg:match})
  \STATE resolve $C$'s state from whether its sources are satisfied and from its
  target's state
  \IF{$C$ matched only in part}
    \STATE refine $C$ and spawn the upstream carrying its residual
    (Alg.~\ref{alg:refine}), and add that upstream to the traversal
    \COMMENT{so it is processed within this same step}
  \ENDIF
\ENDWHILE
\STATE \COMMENT{\textbf{generation} --- only now, with the whole frontier resolved}
\FORALL{events left unexplained by the traversal}
  \STATE for an active target or \csv{} with no active positive conditioner, create a
  positive \csv{} whose source is $O$ itself
  \STATE for an inactive \csv{} with no active negative conditioner, create a negative
  one likewise
\ENDFOR
\STATE fold this step into every \csv{}'s counts, the just-created ones included, so
that their statistics begin from this observation
\STATE \COMMENT{\textbf{structure dynamics} --- two mirror-form bands, each
sampled per step, neither a hard test; $\hat p$ is the chain-effective significance
(Sec.~\ref{app:differences}), equal to $\Pst$ for a top-level \csv{}}
\STATE \textbf{removal:} if $\hat p<\Trem$, discard the conditioner with probability
$\rho\,(\Trem-\hat p)/\Trem$
\STATE \textbf{reintegration:} if $\Pst>\Treint$, and the conditioner is positive and
conditions another \csv{}, merge it into that downstream with probability
$\rho\,(\Pst-\Treint)/(1-\Treint)$
\STATE in both: never act on a \csv{} on its birth step, and down-weight one whose
relation gained no evidence this step
\end{algorithmic}
\end{algorithm}

\subsubsection{Novelty in learning process relative to the prior instantiation.} In terms of the learning process on the model, this instantiation differs from the prior description \citep{erden2025foundations,erden2025thesis} at four points; Sec.~\ref{app:differences} states each exactly. (i)~The refinement algorithm is \emph{plain} removal of non-recurring structure, without the bridging edges (``rerelation'') of the prior version: relational continuity across a removal is now supplied by the representation itself, since a coarser-scale edge already spans any removed fine structure (Sec.~\ref{sec:representation}), and coverage-thresholded matching replaces the prior absence-ratio noise criterion. (ii)~A chain is a \emph{composition} of distinct sources rather than a nest of ever-larger ones: an upstream stores only the structure it adds, joined to its downstream at \emph{anchor} nodes it references but does not own, so structure is stored once (the \csv{} counts of Sec.~\ref{sec:exp-structure} are disjoint contributions), matching binds anchors downstream-first at no extra search, and each \csv{} is readable on its own (Fig.~\ref{fig:interp}). (iii)~Features are read directly off the raw contour---no polygonal approximation---and the multi-scale coarsening into one augmented network is new. (iv) \emph{Reintegration}, which folds an upstream that has ceased to carve any distinction back into its downstream, has no counterpart in the prior work, whereas removal and the presence statistics follow it. Single-target \csv{}s and externalized negative structure are \emph{inherited} from the prior visual extension---a simplifying design choice that sidesteps an implementational detour tangential to this paper's contributions, at the cost of inflating model size and forgoing cross-class reuse (Sec.~\ref{sec:exp-structure}).

\section{Multi-Scale Structural Features} \label{sec:representation}

We now describe the representational contribution: the conversion of a shape image into the network the learner consumes. In the context of this work, we take the figure as binary (thresholding at $0.5$) and describe the first-order instantiation for 2D shape. It is a proof of concept for a general principle that extends to higher orders of change and to 3D shape; we leave those out here to keep the study focused, and regard first-order 2D shape as sufficiently illustrative of the power of the methodology.

\subsubsection{The principle: regions of constancy as points of change.} \label{sec:changepoint} The representation embodies a single general principle: \emph{regions of constancy are represented by their points of change}. An extended stretch continuing flatly collapses to the change points bounding it, and those points---typed by the order, direction and character of the change---become the nodes of a network. Nothing about this is image-specific: it applies to any signal representable as a continuous entity, and extends to higher orders of change where more detail is needed, the informativeness of each order diminishing rapidly. A good analogy is the general method of sketching a function by hand, where one marks its extreme points and interpolates between them, computing the extreme points of progressively higher-order derivatives as more detail is wanted. Our features are in this sense ``designed,'' but designed \emph{once}, from \textit{first principles}, and universally applicable---a candidate primitive for arbitrarily complex visual structure rather than a task-tuned descriptor, \emph{which is why we treat the change-point construction as part of the methodology itself rather than as engineering around it}. Two boundaries of the present instantiation are boundaries of the instantiation, not of the principle: (1)~we read change points off the \emph{contour} of a binarized figure, i.e.\ we apply the construction to a one-dimensional signal traced through the plane; the same construction applies directly to the un-thresholded image treated as a 2D intensity \emph{surface} or a 3D shape, identifying its directional extrema and change lines instead of a curve's turning points---as one sketches a surface by its ridges and saddles; and (2)~we use only the first order of change, where higher orders would enter as additional node types wherever finer discrimination is wanted. Both keep the principle intact and neither is trivial to implement well, so we defer them and develop the first-order, contour-based case here, which suffices to study the learning machinery. The claims this paper makes concern the learning \emph{process}---each observation integrated once, every past response preserved, no replay, readable models---and hold whatever the representation's reach; what a deliberately minimal feature vocabulary bounds is the absolute accuracy a given benchmark allows, not those process properties. This is why a limited representation is a suitable setting in which to study the process, and why expanding the representation (to intensity surfaces, 3D shape, or higher orders of change) is left as orthogonal future work (Sec.~\ref{sec:conclusion}) rather than pursued here.

\subsubsection{From contour to features.} The image is binarized and its contours extracted (outer boundaries and holes distinguished). Each contour is \emph{traversed} and the points where the traversal direction changes sign along either image axis---the directional extrema---are recorded as nodes, typed by the axis and direction of the change and the local convexity (Fig.~\ref{fig:multiscale}b); no smoothing is applied. Nodes are connected by directed edges kept as parallel layers: \emph{contour} edges linking successive change points, and \emph{spatial} edges expressing relative position of directly-reachable nodes, split into horizontal and vertical variants whose direction encodes order along that axis. Every node also carries the \emph{position} observed and every edge the \emph{displacement} between its endpoints; these give matching additional information beyond type compatibility and the read-out its class-conditional evidence, and in a \emph{learned} source these are running estimates over the instances matched.

\subsubsection{Multi-scale coarsening into one network.} 
\begin{figure}[tbp]
\centering
\includegraphics[width=0.96\columnwidth]{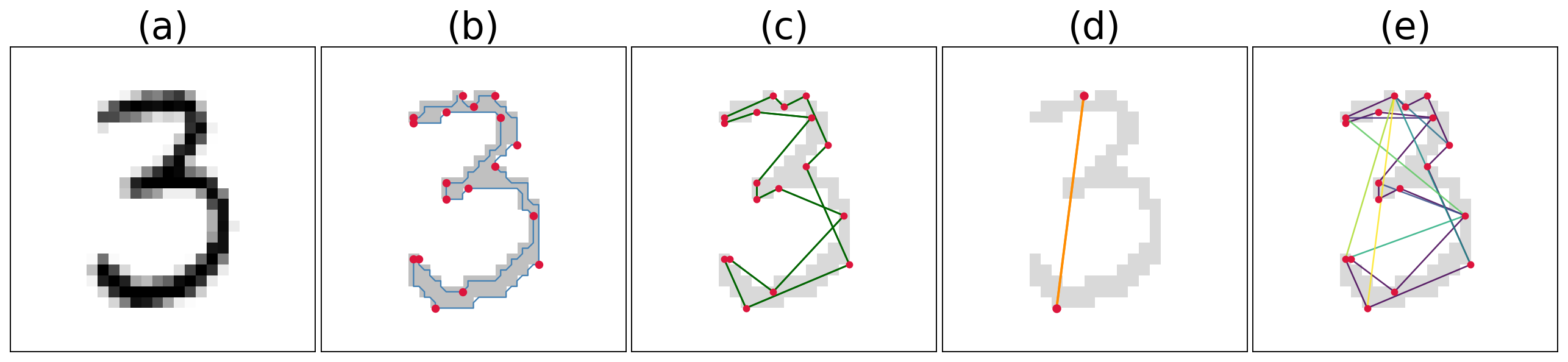}
\caption{Construction of the multi-scale representation (actual pipeline). (a)
Input. (b) Binarized contours with orientation-change points (red) that become
nodes. (c) The finest contour network. (d) The coarsest level: one long-range
relation summarizing the shape. (e) The unified augmented network: edges of
\emph{every} level coexist over the shared node set (color = level). Spatial
layers omitted for legibility.}
\label{fig:multiscale}
\end{figure}

A single-scale network describes the shape at one granularity only, and fixing it in advance decides, before any learning happens, which distinctions the learner can make (Fig.~\ref{fig:schematic}). We therefore do not fix one. The network is coarsened iteratively---at each level the least-extent same-type maximum/minimum pair is removed, contour edges are created between the removed elements' surviving neighbours, and the spatial layers are recomputed---and the representation the learner receives is the \emph{accumulation} of all levels into one augmented network, nodes shared across levels and the edges of every level coexisting as parallel structure (Fig.~\ref{fig:multiscale}). What has been outlined here is a brief form, given the space; the procedure in full is Algorithm~\ref{alg:coarsen} in Sec.~\ref{app:algs}.

Relations of every extent thus coexist over the same nodes---long-range ones stating how distant parts of the shape stand to each other (``this turning point precedes that concavity''), short-range ones stating local detail---with \emph{no} priority attached to either: which a class comes to depend on is settled by learning, and the intuitive reading (longer relations carrying what generalizes, shorter ones what separates) is an \emph{outcome} of refinement, where applicable, rather than an assumption.

Carrying all levels at once has three consequences. \emph{Refinement gains a choice of granularity}: relations failing to recur are reduced away and those recurring retained, so a mature \csv{} is expressed at whatever extents proved reliable. \emph{The algorithm becomes simpler}: when an element is refined away, longer-range edges spanning the gap are already present, so relational continuity survives without a repair step---which makes the base methodology's \emph{rerelation} operation \citep{erden2025foundations} unnecessary here (Sec.~\ref{app:differences}). And \emph{matching becomes well-determined}, as shape-spanning relations constrain a placement local information cannot settle.

\subsubsection{Scope and claims.} The learner is blind to the dimensionality and semantics of the network it refines---a network over 3D feature points would be processed identically---but the conversion from images to networks is not: obtaining a structured 3D description from photographs is a substantial problem in its own right, and the alternative of learning appearance across many stored 2D views is what this methodology exists to avoid. We take our motivation from the structural account of recognition introduced above---that it rests on structural descriptions rather than accumulated 2D appearances \citep{marr1978representation,biederman1987recognition}. We therefore restrict this work to intrinsically 2D shape, for which MNIST is canonical, and treat structured 3D feature extraction as the enabling step for future extensions (see also~\citep{erden2025foundations}). The representation is correspondingly minimal and is not offered as a competitive digit descriptor; the claim is the \emph{abstract design}, that multi-scale visual structure and its cross-scale relations can be expressed as one relational network on which a structural continual learner operates directly. We do, however, defend the underlying principle more strongly than we can demonstrate it here: representing a continuous entity by its points of change is in our view capable of furnishing a \emph{complete} feature description of shape to whatever fidelity is wanted, as further orders are admitted, and likewise for the surface formulation. We do not claim this as an experimental result, but state it as the position the representation is built on, and one we consider both defensible and testable.

\section{Matching a Pattern to an Observation} \label{sec:matching}

Both learning (Sec.~\ref{sec:method}) and prediction (Sec.~\ref{sec:readout}) rest on the same operation: locating where a learned pattern (a \csv{}'s source, or a chain of them) sits within an observed network. We describe it here, before either use, because it is the step at which the representation of Sec.~\ref{sec:representation} does the most work.

\subsubsection{The problem.} 
\begin{figure}[tbp]
\centering
\includegraphics[width=\columnwidth]{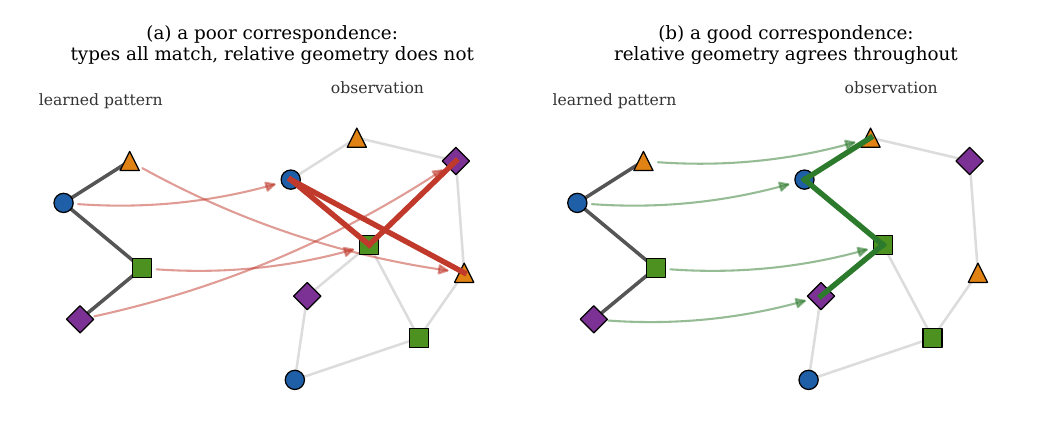}
\caption{The matching problem. A learned pattern (left in each panel) must be
placed within an observation (right) whose nodes carry the same feature types; marker
shape and colour denote type. \textbf{(a)} A poor match: every pattern node
is matched to an observation node of the correct type, yet the placed configuration
(red) bears no resemblance to the pattern's own geometry---the relative displacements
disagree. \textbf{(b)} A good match: the placed configuration (green)
reproduces the pattern's arrangement, so every relation agrees. Type compatibility
alone does not distinguish these; agreement of the relations does.}
\label{fig:matching}
\end{figure}

The pattern (a learned internal representation) and the observation are separate networks of typed nodes. A candidate match assigns pattern nodes to observation nodes of compatible type and must be consistent with the pattern's edges. Because the node types are generic---a given kind of orientation change occurs many times in one digit---type compatibility alone leaves many candidates, and short-range relations do not reduce them much: a small configuration of neighbouring change points typically recurs, in near-identical local geometry, at several places in the same shape, and is likewise prone to noise. What distinguishes the correct placement from these near-duplicates is how the configuration sits relative to the rest of the shape (Fig.~\ref{fig:matching}), across \textit{all} scales the learning process has found \textit{relevant}---which is exactly what the long-range relations of the augmented network express. This is where the prior instantiation fell short: matching was decided by ranking candidates on positional proximity and judging them by how much structure survived the assignment, preserving the largest fit \citep{erden2025foundations}---which, on a single-scale network, admits assignments locally plausible everywhere and globally wrong---visible in the ${\approx}0.50$ ten-class accuracy reported there. Rather than searching harder for the largest fit, we make the observation carry relations whose agreement a correct placement must reproduce.

\subsubsection{The procedure.} Every node of a pattern carries a running estimate of its position and every edge one of its \emph{displacement}, maintained online as means over the observations in which that element matched. The matcher grows a match \emph{incrementally} (Algorithm~\ref{alg:match}): it seeds a pairing of compatible type and extends to a neighbour only when the observed displacement agrees, within a spatial tolerance, with the learned one. Relations of every extent participate, so a placement is constrained by local detail and by relations reaching across the shape at once. The tolerance is swept strict to lenient and nothing is rejected outright on distance---agreement enters as a score falling off with disagreement, so the sweep \emph{orders} the search rather than filtering it.

A match is \emph{full} when the entire pattern is placed and \emph{partial} otherwise. Matching is \emph{chain-aware} (pinned) in the following sense: A \emph{mid-chain} \csv{} is one whose target is another \csv{} rather than a class: its own source is the additional structure carved on top of the configuration the downstream already represents (Sec.~\ref{sec:method}; in the base methodology such an upstream instead \emph{contained} its downstream's source, whereas here the two are separate sources joined at \emph{anchor} nodes, Sec.~\ref{app:differences}). It is placed relative to where that downstream matched, so a chain is matched as one coherent object rather than piecewise---which is what lets deep chains discriminate on topology. At learning time the same procedure drives refinement and updates the surviving geometry toward the observed values.

\begin{algorithm}[!tb]
\caption{Matching a pattern to an observation}
\label{alg:match}
\begin{algorithmic}[1]
\REQUIRE pattern $P$ (a \csv{} source or chain, with learned node positions and edge
displacements); observation $O$; tolerances $\tau_1<\dots<\tau_m$, strict to lenient
\IF{$P$ is mid-chain}
  \STATE pin $P$'s anchor nodes to where its downstream matched in $O$
\ENDIF
\STATE $\mathcal{M} \gets$ a population of seed matches, each pairing a node of
$P$ with a node of $O$ of compatible type, ordered by positional proximity
\FORALL{tolerances $\tau$ in increasing order}
  \STATE let every match in $\mathcal{M}$ reconsider the extensions it
  previously declined \COMMENT{a looser $\tau$ may now admit them}
  \REPEAT
    \FORALL{$M\in\mathcal{M}$, and unplaced nodes $n$ of $P$ adjacent to a node $M$
    has placed}
      \STATE for each $O$-node of compatible type, score placing $n$ there by how
      closely its displacement from that neighbour agrees with the learned one, the
      score falling off over $\tau$
      \STATE admit each candidate with probability given by its score
      \COMMENT{nothing is rejected outright on distance}
    \ENDFOR
    \STATE where admitted placements compete for the same $O$-node, fork $M$ into
    alternatives; cap the population, setting the weakest aside
  \UNTIL{a round admits nothing}
\ENDFOR
\STATE $M^\star \gets$ the surviving match of greatest coverage
\STATE $\mathrm{degree} \gets$ the smaller of the fraction of $P$'s own nodes placed
and the fraction of its own edges placed
\COMMENT{anchors excluded: they are pinned, not searched for}
\STATE \textbf{return} $M^\star$, $\mathrm{degree}$, and
$\textsc{full} = [\mathrm{degree} = 1]$
\end{algorithmic}
\end{algorithm}

\section{Read-Out: Structure to Decision} \label{sec:readout}

Learning produces, per class, positive \csv{} chains and negative suppressive structure. Classification requires a \emph{read-out}: a procedure mapping a trained model and a new observation to a class. As it reads the structure and statistics and mutates nothing, it is strictly separate from learning.

The read-out consumes what matching (Sec.~\ref{sec:matching}) makes available: which \csv{}s are present, and, for those that are, where their nodes and edges were placed (Sec.~\ref{app:readout-details}). The learned spatial statistics are the informative part here. Position is unreliable as a criterion for \emph{deciding} a match, but once a match has been fixed on structural grounds, the positions and orientations at which a \csv{}'s elements landed are class-conditional evidence: a \csv{} of class $k$ typically matches its own class at a characteristic place and orientation, and matches other classes elsewhere. Our read-out is built on this distinction, and on treating the \emph{absence} of a \csv{} as informative rather than as no evidence at all.

\subsubsection{Accumulated \csv{} statistics.} Each \csv{} $c$ accumulates statistics for its own class $k_c$ only, together with a class-agnostic \emph{pool} that serves as a background: two firing counters; a streaming mean and variance of each matched node's \emph{position} (a Welford accumulator, one per node for the own class and one for the pool); and a circular mean of each matched edge's \emph{orientation}. A \csv{}'s class $k_c$ is read off its immediate target, resolved one hop at a time. These accumulators are distinct from the matcher's own estimates---which keep one unconditioned mean per element to \emph{decide} a match, whereas these keep a mean \emph{and} a spread to weigh one already decided---and from the presence statistics $n_{\mathrm{active}},n_{\mathrm{inactive}}$ the learner maintains (Sec.~\ref{sec:method}), which are defined relative to a \csv{}'s immediate target and drive the maturity filter of Sec.~\ref{sec:exp-structure}.

\subsubsection{Three kinds of evidence.} \emph{Firing.} Write $m^{\mathrm{own}}_c$ and $m^{\mathrm{tot}}_c$ for the number of steps at which $c$ existed in the model with its own class active, and at which it existed at all; and $n^{\mathrm{own}}_c$, $n^{\mathrm{tot}}_c$ for the subsets of those steps at which $c$ additionally \emph{fired}, that is, matched the observation in full (the condition is the activity of the class $c$ conditions toward, not the state of its immediate target, which for a mid-chain \csv{} may be inactive while the class is present). The two firing rates are frequencies smoothed by adding $\alpha=\tfrac12$ (Jeffreys),
\begin{equation*}
P^{o}_c=\frac{n^{\mathrm{own}}_c+\alpha}{m^{\mathrm{own}}_c+2\alpha},
\qquad
P^{f}_c=\frac{(n^{\mathrm{tot}}_c-n^{\mathrm{own}}_c)+\alpha}
             {(m^{\mathrm{tot}}_c-m^{\mathrm{own}}_c)+2\alpha},
\end{equation*}
clipped away from $0$ and $1$; the foreign rate is an \emph{aggregate} over all classes other than $c$'s own, and the smoothing keeps a \csv{} that has never yet fired from contributing an unbounded term.

\emph{Position.} A position anchor is a streaming Welford accumulator $a=(n,\mu_x,\mu_y,M_{2x},M_{2y})$ over the positions at which that source node has been placed, one per node for the \csv{}'s own class and one for the pool. Its per-axis spread is $\sigma=\sqrt{M_2/n+\varphi^2}$ with $\varphi=1.5$\,px, and
\begin{equation*}
\ell(p;a)=-\tfrac12\Big[\Big(\tfrac{p_x-\mu_x}{\sigma_x}\Big)^{\!2}
                        +\Big(\tfrac{p_y-\mu_y}{\sigma_y}\Big)^{\!2}\Big]
          -\log(\sigma_x\sigma_y),
\end{equation*}
a diagonal-Gaussian log-density up to an additive constant that cancels in the difference $\ell(p;a^{o})-\ell(p;a^{\mathrm{pool}})$. The floor $\varphi$ makes the spread self-calibrating: without it a node placed almost identically in every match would have $\sigma\!\to\!0$ and dominate the sum, so $\varphi$ sets the resolution below which positional agreement is not treated as informative.

\emph{Orientation.} Orientation is \emph{axial}---a segment and its reverse are the same orientation---so an edge angle $\theta$ enters through its doubled angle. An orientation anchor accumulates $(n,\sum\cos 2\theta,\sum\sin 2\theta)$, giving a mean direction $\bar\theta=\tfrac12\arctan2(\sum\sin 2\theta,\sum\cos 2\theta)$, and the deviation of an observed $\theta$ from it is the circular distance $\Delta=\min\!\big(\delta,\pi-\delta\big)$ with $\delta=|\theta-\bar\theta|\bmod\pi$. The term $\kappa(\cos 2\Delta^{o}-\cos 2\Delta^{\mathrm{pool}})$ is the log-ratio of two von Mises densities of common concentration $\kappa$ up to their (identical) normalizers. Both geometric terms are gated on having at least two observations in the anchor and in the pool, so a single placement never sets a class-conditional expectation; their weights are fixed for all reported runs at $w_p=3.0$, $w_o=1.0$ and $\kappa=2.0$ (Sec.~\ref{app:repro}).

\subsubsection{Assembling the score.} These combine by one principle: each \csv{} contributes to the score of its own class $k_c$ alone, weighing its own-class evidence against the pooled background, and the only operation comparing classes is a final $\arg\max$---no global classifier, learned weights or gradient. If $c$ is \emph{absent} it still contributes the absence term $s_{k_c}\mathrel{+}=\log\frac{1-P^{o}_c}{1-P^{f}_c}$, so a \csv{} that usually fires for its class but is silent here \emph{demotes} that class. If \emph{present} it contributes the firing term plus the position and orientation terms above:
\begin{align}
s_{k_c}\mathrel{+}=\ &\log\tfrac{P^{o}_c}{P^{f}_c}
 + w_p\!\!\sum_{n}\!\big[\ell(p_n;a^{o}_{c,n})-\ell(p_n;a^{\text{pool}}_{c,n})\big]\nonumber\\
 &+ w_o\,\kappa\!\!\sum_{e}\!\big[\cos 2\Delta^{o}_{c,e}-\cos 2\Delta^{\text{pool}}_{c,e}\big].
\end{align}
Every component is textbook; the contribution is that this information is present in the learned structure and that a read-out respecting the paradigm's locality recovers it with no global classifier or fitted weights, and, being read-only, does not affect the learning process (Algorithm~\ref{alg:readout}). Sec.~\ref{app:readout-details} covers the (model-independent) cost of the presence pass and two minor policies; Sec.~\ref{app:readout} reports a simpler read-out consulting \emph{only} which \csv{}s matched, ablating the geometry.

\begin{algorithm}[!tb]
\caption{Read-out for one observation}
\label{alg:readout}
\begin{algorithmic}[1]
\REQUIRE trained model; observation $O$; weights $w_p,w_o,\kappa$
\STATE $s_k \gets 0$ for every class $k$
\STATE compute the presence of every \csv{} against $O$ (Alg.~\ref{alg:match},
downstream-first)
\FORALL{\csv{}s $c$}
  \STATE $k_c \gets$ class of $c$'s immediate target, resolved one hop at a time
  \STATE \textbf{if} $k_c$ is undefined \textbf{then} skip $c$
  \STATE $P^{o}_c,P^{f}_c \gets$ smoothed own-class and aggregate-foreign firing rates
  \IF{$c$ is not fully present}
    \STATE $s_{k_c} \mathrel{+}= \log\frac{1-P^{o}_c}{1-P^{f}_c}$
    \COMMENT{absence counts as evidence}
  \ELSE
    \STATE $s_{k_c} \mathrel{+}= \log\frac{P^{o}_c}{P^{f}_c}$
    \FORALL{source nodes $n$ of $c$, placed at $p_n$}
      \STATE \textbf{if} both the own-class and pooled anchors of $n$ have $\ge 2$
      observations \textbf{then}
      $s_{k_c} \mathrel{+}= w_p\big[\ell(p_n;a^{o}_{c,n})-\ell(p_n;a^{\mathrm{pool}}_{c,n})\big]$
    \ENDFOR
    \FORALL{source edges $e$ of $c$, placed at angle $\theta_e$}
      \STATE \textbf{if} both orientation anchors of $e$ have $\ge 2$ observations
      \textbf{then}
      $s_{k_c} \mathrel{+}= w_o\kappa\big[\cos 2\Delta^{o}_{c,e}-\cos 2\Delta^{\mathrm{pool}}_{c,e}\big]$
    \ENDFOR
  \ENDIF
\ENDFOR
\STATE \textbf{return} $\arg\max_k s_k$
\end{algorithmic}
\end{algorithm}

\section{Experiments} \label{sec:experiments}

\subsection{Protocol} \label{sec:exp-protocol} We use two class-incremental shape benchmarks run through the same pipeline: MNIST \citep{lecun1998gradient} (ten digit classes) and the harder EMNIST-Letters \citep{cohen2017emnist} (twenty-six classes, each merging a letter's upper- and lower-case forms, so a single class spans two distinct shapes). Both are filtered for topological consistency to single-connected-component glyphs (for MNIST we additionally require the loop structure expected of each digit, a criterion specific to digits); this keeps about $85\%$ of MNIST and about $97\%$ of EMNIST. The filtering is a scope simplification, not a requirement of the method: the present feature construction reads structure off clean contours and does not yet model topological \emph{gaps} within a stroke (breaks or disconnections), just as it does not model smooth intensity transitions---both handleable in principle by the same change-point construction (Sec.~\ref{sec:changepoint}) but a lengthy extension orthogonal to what is tested here. We therefore keep only single-component glyphs---the large majority of each dataset (exact criteria in Sec.~\ref{app:repro})---and train and evaluate \emph{all} methods on the identical subset. MNIST is the primary benchmark and carries the full set of analyses; EMNIST-Letters is a harder second benchmark on which we verify that the continual-learning result persists, and that the learned models and representations remain comprehensible, as the class count more than doubles (Sec.~\ref{sec:exp-emnist}). A \emph{cycle} presents the classes in fixed order, five samples each, and we run twenty cycles. For the Modeller, three properties define what is being tested: \textbf{exactly one online learning step per sample} (no batches), \textbf{no re-presentation and no replay buffer}, and \textbf{no task- or cycle-boundary signal} (all neural baselines violate at least one, detailed below). This is not difficulty for its own sake: it is what a natural stream of experience looks like before it is collected into a dataset. The comparison in Sec.~\ref{sec:exp-regimes} is therefore between methods under \emph{different} assumptions, stated explicitly there. Held-out evaluation uses twenty images per class on MNIST and ten per class on EMNIST, at a disjoint offset, after each class block of every cycle; the main results average ten seeds (the ablation, scaling and additional sweeps use fewer, stated where they appear), and the \emph{baseline} configuration for ablations is the full system with the local geometric read-out. The Modeller's learning parameters (Sec.~\ref{app:repro}) were fixed in advance and never selected on evaluation data---their sensitivity is characterised afterwards (Sec.~\ref{app:sensitivity})---while its read-out weights were set during development on MNIST and carried to EMNIST unchanged; each baseline, in contrast, was tuned by a sweep on the held-out set, which can only favour it.

\subsection{Continual Learning Across Stream Regimes (MNIST)} \label{sec:exp-regimes} \label{sec:exp-Modeller} We evaluate continual learning on MNIST by presenting the identical data---$200$ class-blocks, twenty per class, five samples per block---under a spectrum of stream regimes that weakens recurrence from a regular cycle to a strict single pass, changing only the order of the blocks (Sec.~\ref{app:regimes}): \emph{cyclic} (the ten classes recur in a fixed order), \emph{shuffled} (each cycle an independent random permutation of the classes), \emph{permuted} (one global shuffle, no cycle structure), \emph{longgap} (four rounds in reshuffled class order; within a round each class is trained as one consecutive run, then absent for ${\sim}45$ blocks until its next run), and \emph{blocked} (each class trained once and \emph{never revisited}---the strict single-pass limit). Regularity matters because when every class is re-encountered on a short, fixed schedule a method that merely relearns quickly is hard to tell from one that retains; the spectrum separates the two.

We compare against representative neural continual-learning baselines under their best settings, spanning the three families that need no visible task boundary: \emph{regularisation} and distillation, storing only per-parameter statistics or a model snapshot (SI \citep{zenke2017si}, MAS \citep{aljundi2018mas}, LwF \citep{li2018lwf}); \emph{replay}, storing and re-presenting past examples (naive experience replay; DER++ \citep{buzzega2020derpp}, which also stores the logits emitted when each example was inserted; ER-ACE \citep{caccia2022erace}); and \emph{expansion}, an autoencoder-gated expert model \citep{erden2025aec}, standing in for the task-based \emph{expansion} family, but spawning and routing its experts from reconstruction error rather than from predefined task boundaries. Each is at its best configuration over a sweep of its own hyper-parameters (Sec.~\ref{app:clbaselines}). The comparison is between methods under \emph{different} assumptions, which we state explicitly (Sec.~\ref{app:assumptions}): replay stores and re-presents data; the regularisers store none but each needs a \emph{consolidation schedule}, which ours does not; and none learns strictly sample-by-sample---each block is a batch of five revisited for twenty gradient updates, an advantage they hold and ours does not (rerun one sample at a time, every replay-free baseline loses accuracy; Sec.~\ref{app:clbaselines}). We report three metrics: \emph{final accuracy}; \emph{forgetting} (per-class peak minus final); and \emph{within-cycle retention} (WCR), the fraction of a class's just-trained accuracy still held when it next returns (WCR$_{\mathrm{end}}$) and at its worst point in the interval (WCR$_{\mathrm{wst}}$), measured across each class's true absence, with a $\theta=0.3$ floor on the just-trained accuracy and each ratio capped at $1$ (Sec.~\ref{app:wcr}). Forgetting and WCR are direct measures of continual learning: how much of a class survives the learning of others.

\subsubsection{Accuracy, forgetting, and retention across the spectrum.} Tables~\ref{tab:regimes} and~\ref{tab:wcr-main} report the three metrics for every method across the spectrum (the final-accuracy trend is also plotted in Fig.~\ref{fig:regimes}, Sec.~\ref{app:regimes}). Under the regular regimes---cyclic, shuffled and permuted---the Modeller and the two strongest regularisers reach \emph{close final accuracy}: under cyclic the Modeller ($0.874$, ten seeds) is statistically level with SI ($0.879$) and MAS ($0.865$) ($p{=}0.82$ and $p{=}0.22$ respectively; every other difference is significant at $p<0.01$, two-sided Wilcoxon signed-rank, Sec.~\ref{app:stats}). This much is expected: on a re-exposing stream every class returns often, so as cycles accumulate the procedure approaches stochastic gradient descent with a slow time constant, under which the regularisers eventually reach similar accuracy.

Where these methods differ under the regular regimes is in \emph{forgetting} and \emph{within-cycle retention}. The regularisers give up much of a class between its returns and relearn it on each return, so their forgetting is higher and their WCR far lower than the Modeller's---WCR$_{\mathrm{end}}$ ${\approx}0.75$ against ${\approx}0.95$ across all cycles under cyclic, even where final accuracy is close. The gap is widest in each class's \emph{early episodes} (Table~\ref{tab:wcr-main}, top), where the regularisers hold only ${\approx}0.2$--$0.3$ against the Modeller's ${\approx}0.9$ (WCR$_{\mathrm{end}}$). It is those early episodes that test continual learning most sharply: as cycles accumulate the stream becomes effectively i.i.d.---slow-timescale stochastic gradient descent under data abundance, where the regularisers recover---so the all-episode figures, which average the easy later cycles in, \emph{understate} the difference. The Modeller's WCR is indistinguishable across cyclic, shuffled and permuted, so its retention survives random and non-cyclic order as well.

As recurrence weakens, the difference reaches final accuracy too. By \emph{longgap} the Modeller holds $0.786$ against SI $0.619$ and MAS $0.534$; under \emph{blocked} it retains $0.661$ while SI ($0.104$), MAS ($0.106$) and the plain network ($0.100$) collapse to chance---forgetting essentially everything (${\sim}0.90$, against the Modeller's $0.296$) and retaining nothing (WCR$_{\mathrm{end}}$/WCR$_{\mathrm{wst}}$ ${\approx}0.10/0.10$, against $0.72/0.52$). \textit{Even with no re-exposure at all}, the Modeller preserves a considerable portion of what it learned and degrades gracefully---its central strength. It does retain more \emph{with} re-exposure than without, so if anything the regular regimes \emph{conceal} its relative advantage compared to classical approaches, which widens as recurrence weakens. Among replay-free methods the Modeller alone retains \emph{and} stays accurate across the whole spectrum: the expert model retains within each cycle but at far lower accuracy (retention without an accurate read-out; the cyclic close-up below explains why), and only DER++ with an $80$-image buffer (eight percent of the stream, re-presented at every step) does both---by replaying stored data, ordinary training on a growing pool rather than continual learning in the sense meant.

\begin{table}[tbp]
\centering
\caption{Final accuracy (top) and forgetting (bottom) as the stream's
recurrence is weakened (MNIST, same data reordered; mean, ten seeds). The
regularisation baselines, level with the Modeller under cyclic recurrence, collapse to chance
once classes stop recurring (forgetting ${\sim}0.9$), showing their retention to depend on
re-exposure; the Modeller degrades gracefully, while among the baselines only the $80$-image replay
buffer stays strong under blocked (the expert model degrades heavily, the regularisers to
chance). Bold: best replay-free method in each column; the replay method is in italics. All ten
methods with standard deviations, and model size by regime: Sec.~\ref{app:regimes}. $^\star$This work.}
\label{tab:regimes}
\small
\setlength{\tabcolsep}{4pt}
\begin{tabular}{lccccc}
\toprule
Method & cyclic & shuffled & permuted & longgap & blocked \\
\midrule
\multicolumn{6}{l}{\emph{Final accuracy (higher is better)}} \\
Modeller$^\star$ & $.874$ & $.859$ & $\mathbf{.845}$ & $\mathbf{.786}$ & $\mathbf{.661}$ \\
SI & $\mathbf{.879}$ & $.859$ & $.829$ & $.619$ & $.104$ \\
MAS & $.865$ & $\mathbf{.863}$ & $.832$ & $.534$ & $.106$ \\
Plain NN & $.661$ & $.651$ & $.584$ & $.180$ & $.100$ \\
Expert-AEC & $.607$ & $.597$ & $.540$ & $.489$ & $.350$ \\
\textit{DER++ (80)} & $\mathit{.911}$ & $\mathit{.894}$ & $\mathit{.886}$ & $\mathit{.876}$ & $\mathit{.754}$ \\
\midrule
\multicolumn{6}{l}{\emph{Forgetting, peak$-$final (lower is better)}} \\
Modeller$^\star$ & $\mathbf{.100}$ & $\mathbf{.112}$ & $\mathbf{.118}$ & $\mathbf{.171}$ & $\mathbf{.296}$ \\
SI & $.120$ & $.141$ & $.170$ & $.381$ & $.896$ \\
MAS & $.135$ & $.137$ & $.167$ & $.466$ & $.894$ \\
Plain NN & $.339$ & $.349$ & $.416$ & $.820$ & $.900$ \\
Expert-AEC & $.245$ & $.232$ & $.303$ & $.321$ & $.426$ \\
\textit{DER++ (80)} & $\mathit{.071}$ & $\mathit{.090}$ & $\mathit{.100}$ & $\mathit{.112}$ & $\mathit{.235}$ \\
\bottomrule
\end{tabular}
\end{table}

\begin{table}[tbp]
\centering
\caption{Within-cycle retention (WCR$_{\mathrm{end}}$/WCR$_{\mathrm{wst}}$) on MNIST, by regime, over each
class's early episodes (its first three) and over all episodes; ten-seed means. WCR is
the fraction of a class's just-trained accuracy still held when it next returns (and at its worst
point in between)---the direct continual-learning measure, which separates methods where final
accuracy does not. The early window isolates the period of genuine stress: as cycles accumulate,
re-exposure turns the stream effectively i.i.d.\ and every method recovers, so the replay-free gap
is widest early. Among replay-free methods only the Modeller retains and stays
accurate; the expert model also retains (expert isolation) but at far lower accuracy
(Table~\ref{tab:regimes}), and the regularisers collapse by blocked; DER++ (80) retains
comparably (more under blocked) but by replaying stored data. Bold: best replay-free method in each
column; the replay method is in italics. All methods, with standard deviations: Table~\ref{tab:wcr-full}. $^\star$This work.}
\label{tab:wcr-main}
\small
\setlength{\tabcolsep}{2.5pt}
\begin{tabular}{lccccc}
\toprule
Method & cyclic & shuffled & permuted & longgap & blocked \\
\midrule
\multicolumn{6}{l}{\emph{Early episodes (each class's first three)}} \\
Modeller$^\star$ & $\mathbf{.89}/\mathbf{.84}$ & $\mathbf{.91}/\mathbf{.84}$ & $\mathbf{.88}/\mathbf{.81}$ & $\mathbf{.88}/\mathbf{.76}$ & $\mathbf{.72}/\mathbf{.52}$ \\
SI & $.29/.13$ & $.24/.16$ & $.28/.22$ & $.17/.09$ & $.10/.10$ \\
MAS & $.18/.02$ & $.13/.06$ & $.19/.14$ & $.14/.06$ & $.11/.10$ \\
Plain NN & $.10/.00$ & $.06/.02$ & $.11/.07$ & $.01/.00$ & $.10/.10$ \\
Expert-AEC & $.63/.60$ & $.75/.72$ & $.68/.67$ & $.60/.56$ & $.53/.47$ \\
\textit{DER++ (80)} & $\mathit{.88}/\mathit{.81}$ & $\mathit{.85}/\mathit{.77}$ & $\mathit{.86}/\mathit{.80}$ & $\mathit{.82}/\mathit{.76}$ & $\mathit{.76}/\mathit{.71}$ \\
\midrule
\multicolumn{6}{l}{\emph{All episodes}} \\
Modeller$^\star$ & $\mathbf{.95}/\mathbf{.92}$ & $\mathbf{.96}/\mathbf{.92}$ & $\mathbf{.95}/\mathbf{.92}$ & $\mathbf{.90}/\mathbf{.79}$ & $\mathbf{.72}/\mathbf{.52}$ \\
SI & $.75/.67$ & $.74/.68$ & $.72/.67$ & $.29/.20$ & $.10/.10$ \\
MAS & $.75/.68$ & $.76/.73$ & $.73/.70$ & $.24/.16$ & $.11/.10$ \\
Plain NN & $.41/.17$ & $.48/.37$ & $.46/.37$ & $.05/.04$ & $.10/.10$ \\
Expert-AEC & $.92/\mathbf{.92}$ & $.92/.91$ & $.90/.90$ & $.70/.67$ & $.53/.47$ \\
\textit{DER++ (80)} & $\mathit{.94}/\mathit{.92}$ & $\mathit{.94}/\mathit{.91}$ & $\mathit{.93}/\mathit{.91}$ & $\mathit{.84}/\mathit{.79}$ & $\mathit{.76}/\mathit{.71}$ \\
\bottomrule
\end{tabular}
\end{table}

\subsubsection{Detailed analysis of cyclic regime and failure of baselines.} 
\begin{figure}[t]
\centering
\includegraphics[width=0.80\columnwidth]{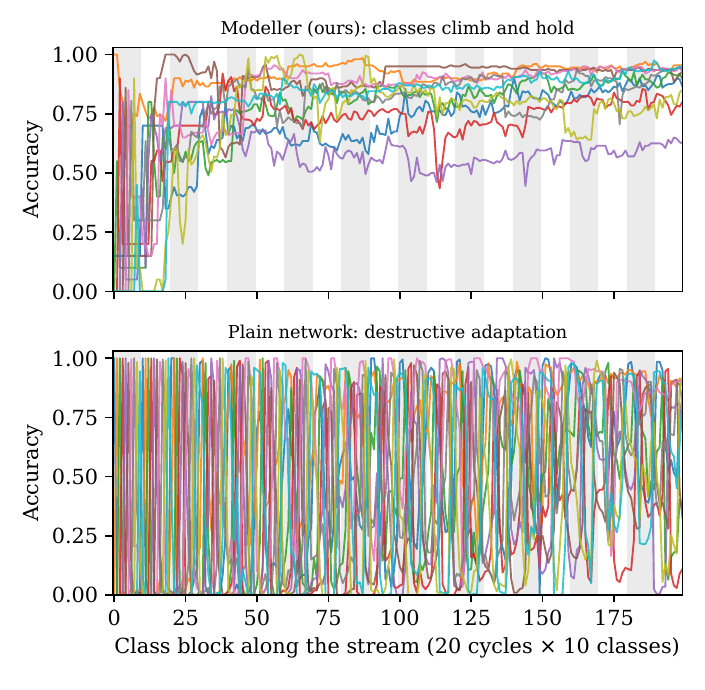}
\caption{Per-class held-out accuracy along the class-incremental MNIST stream (ten-seed
means). Each alternating white/grey band is one \emph{cycle}; within a band the ten tick
positions are that cycle's ten class blocks in order, so the $k$-th point is the model
just after the $k$-th class was trained. Each line follows one class throughout.
\textbf{Top:} the Modeller---classes climb and \emph{hold}, with transient, recovering
dips. \textbf{Bottom:} the plain network---each class collapses when the following
classes train, the destructive-adaptation signature.}
\label{fig:retention}
\end{figure}

\begin{figure}[t]
\centering
\includegraphics[width=0.62\columnwidth]{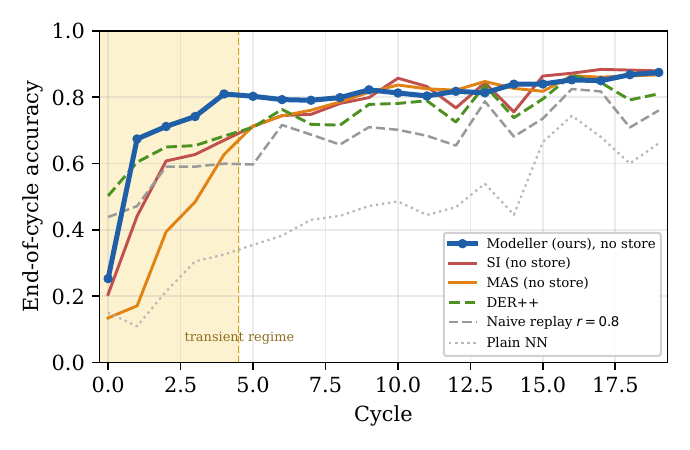}
\caption{End-of-cycle accuracy over the MNIST stream (ten-seed means, identical stream). The shaded
band marks cycles $0$--$4$, where continual learning is under test: there a method either holds
what it has learned or gives it up. Past it, repeated re-exposure turns the procedure into
stochastic gradient descent with a slow time constant, and the regularisers catch up. Per-class breakdowns and the remaining baselines are in
Secs.~\ref{app:clbaselines}--\ref{app:largebuffer}.}
\label{fig:replay}
\end{figure}

On the cyclic stream, Figure~\ref{fig:retention} tracks every class's held-out accuracy along the whole stream for the Modeller (top) against a plain network (bottom): each class's accuracy rises when it is introduced and then \emph{holds} through every later class. The remaining dips are transient and self-correcting---a class falls to around $84\%$ of its just-trained accuracy at its lowest point in its early episodes (Table~\ref{tab:wcr-main})---whereas the plain network drops to near zero the moment other classes train: our dips are residual interference around a retained representation, its collapses the representation itself being overwritten. End-of-cycle accuracy climbs steeply over the first cycles ($0.244, 0.641, 0.709$ for cycles $0$--$2$; Fig.~\ref{fig:replay}), plateaus by cycle $4$--$5$ and ends at \textbf{0.874} against the ${\approx}0.50$ previously reported \citep{erden2025foundations,erden2025thesis}, with mean forgetting $0.100$ (Table~\ref{tab:regimes}). Per-class accuracies lie between $0.80$ and $0.96$ for nine of ten classes (Table~\ref{tab:perclass-full}). Class~$4$ is the exception at $0.63$, confused with $9$---a representational residual discussed below.

The baselines' failure is mechanistic (Fig.~\ref{fig:retention}, bottom; per-class curves for all replay-free baselines in Fig.~\ref{fig:clforgetnostore}). A fully connected network loses nearly all information on a class as the following classes train; the regularisers reach high \emph{final} accuracy only by relearning fast enough to recover by the run's end, having given most of it up in between (SI and MAS reach $0.879$/$0.865$, indistinguishable from our $0.874$, while retaining far less within each cycle, Table~\ref{tab:wcr-main}). Replay attenuates but does not remove the oscillation---each stored example lifts the troughs, yet the per-class curves still collapse and recover cycle after cycle (Figs.~\ref{fig:replayforget} and~\ref{fig:clforgetreplay}), while ours shows none of it and stores no past data; enlarging the buffer improves replay in proportion to what it re-presents, but a buffer of $80$ makes each update dominated by stored rather than arriving data, ordinary training on a growing pool (Sec.~\ref{app:largebuffer}). The autoencoder-gated expert model is the opposite: isolating past experts brings within-cycle stability but caps accuracy by test-time routing ($0.607$), so isolation solves overwriting but breaks integration (Sec.~\ref{app:aecforget}).

\subsection{Testing on EMNIST} \label{sec:exp-emnist}
On EMNIST-Letters the same pattern sharpens (Table~\ref{tab:emnist-regimes}). The Modeller now leads in every regime, and its margin \emph{widens} as the class count more than doubles. Under the regular regimes the Modeller reaches $0.61$--$0.63$ final accuracy against the strongest baseline's (MAS) $0.48$--$0.49$---a gap of ${\approx}\,0.14$ where on MNIST the two were level---and as recurrence weakens the baselines fall away exactly as before: by \emph{blocked} every baseline collapses to chance ($0.04$--$0.05$, chance $=1/26=0.038$) while the Modeller still retains a considerable portion ($0.30$). Within-cycle retention tells the same story more sharply: at cycle~$1$ the Modeller holds WCR$_{\mathrm{end}}$ $0.72$ (cyclic) and $0.74$ (shuffled) against the best baseline's $0.14$/$0.15$, rising to $0.95$/$0.95$ by cycle~$10$ while the baselines reach only $0.60$/$0.64$. Relative to the accuracy a conventionally-trained network reaches on the same data under i.i.d.\ training---the representation's ceiling (Sec.~\ref{app:emnist})---the Modeller attains $0.81$ of it against the baselines' ${\approx}\,0.62$, so the harder benchmark lowers every absolute number, as the minimal feature vocabulary predicts, yet leaves the continual-learning advantage intact and, if anything, larger. We report WCR for the recurring regimes (cyclic, shuffled) only: under weak re-exposure final accuracy already \emph{is} the retention measure, so WCR adds nothing there (standard deviations and late-cycle WCR in Table~\ref{tab:emnist-regimes-full}, Sec.~\ref{app:emnist}).

\begin{table}[tbp]
\centering
\caption{EMNIST-Letters ($26$ classes), continual learning across the stream spectrum (means, ten seeds).
Final accuracy under all five regimes, and within-cycle retention (WCR$_{\mathrm{end}}$ at cycles $1$ and
$10$) for the two recurring regimes; the Modeller and the four replay-free baselines run on EMNIST.
The baselines collapse to chance ($1/26=0.038$) once recurrence is removed, while the Modeller
degrades gracefully. Bold: best in each column (all methods are replay-free). Standard deviations and
late-cycle WCR: Table~\ref{tab:emnist-regimes-full}. $^\star$This work.}
\label{tab:emnist-regimes}
\footnotesize
\setlength{\tabcolsep}{3pt}
\begin{tabular}{lccccccc}
\toprule
 & \multicolumn{5}{c}{\emph{Final accuracy}} & \multicolumn{2}{c}{\emph{WCR (cycle 1/10)}} \\
\cmidrule(lr){2-6}\cmidrule(lr){7-8}
Method & cyc & shuf & perm & long & blk & cyc & shuf \\
\midrule
Modeller$^\star$ & $\mathbf{.62}$ & $\mathbf{.63}$ & $\mathbf{.61}$ & $\mathbf{.63}$ & $\mathbf{.30}$ & $\mathbf{.72}/\mathbf{.95}$ & $\mathbf{.74}/\mathbf{.95}$ \\
MAS & $.48$ & $.49$ & $.48$ & $.22$ & $.05$ & $.14/.60$ & $.15/.64$ \\
SI & $.40$ & $.45$ & $.37$ & $.09$ & $.04$ & $.08/.36$ & $.06/.37$ \\
LwF & $.29$ & $.31$ & $.26$ & $.04$ & $.04$ & $.06/.22$ & $.06/.23$ \\
Plain NN & $.29$ & $.39$ & $.34$ & $.07$ & $.04$ & $.07/.19$ & $.08/.33$ \\
\bottomrule
\end{tabular}
\end{table}

\subsection{Sensitivity to Input Transformations} \label{sec:exp-transform} 
\begin{figure}[tbp]
\centering
\includegraphics[width=0.70\columnwidth]{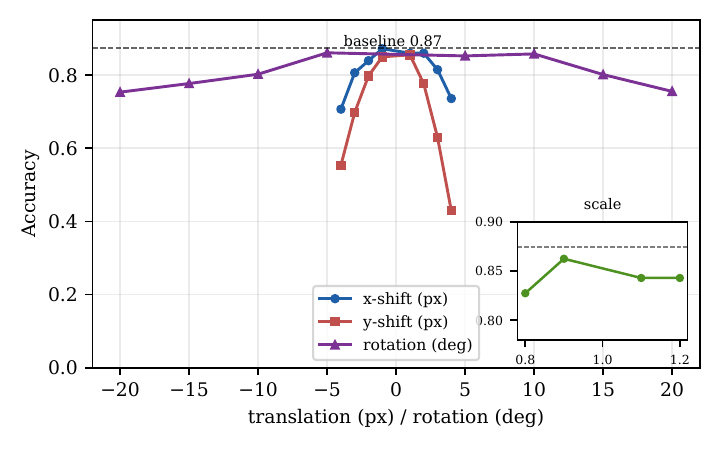}
\caption{Test-time transformation sensitivity (model trained on centred digits, not retrained;
ten-seed means, baseline $0.874$ dashed). Accuracy is robust to small translations and rotations
and to $20\%$ scaling (inset), and degrades smoothly beyond. Full grid: Sec.~\ref{app:transform}.}
\label{fig:transform}
\end{figure}

The geometric read-out uses absolute node position, which suits centred MNIST; we characterise what that costs under input transformations absent from training. We take the model trained on the standard centred protocol, leave it unchanged, and evaluate the held-out set under test-time translation, scaling and rotation (Fig.~\ref{fig:transform}; full grid in Sec.~\ref{app:transform}). From a baseline of $0.874$, accuracy is robust to small perturbations---a $\pm1$--$2$ pixel shift, a $\pm5^\circ$ rotation, or a $20\%$ rescaling each cost little (scale is the most robust axis)---and degrades smoothly rather than sharply beyond, with vertical translation costing more than horizontal (an asymmetry inherited from digit geometry). As the model is not retrained, this isolates the read-out's positional dependence: it is not invariant, but the degradation is graceful. Possible future modifications for learning invariances---by choosing a position-free (e.g.\ relative-position) read-out---would be a change to the read-out alone, leaving the learner untouched.

\subsection{Model Growth} \label{sec:exp-structure} Because the model \emph{grows} its own structure rather than filling a fixed architecture, its size is an observable of learning; and unbounded growth is the obvious failure mode for a learner that adds structure in response to what it sees. As intended by the Modeller's design, it does not occur. Under the cyclic MNIST stream (other regimes are treated below), growth is quick while classes are novel ($\sim\!21$ \csv{}s after cycle $0$, $\sim\!134$ after cycle $5$) then saturates near $\sim\!187$ from cycle ${\sim}11$, fluctuating by $7.7$ \csv{}s between checkpoints, even though novel samples keep arriving and none is re-presented. Generation does not stop; what it produces (which can be regarded as a transient pool of \textit{hypotheses}) is not \emph{retained} as long as existing structure can account for what arrives to a statistically significant degree, so the population settles into an \emph{equilibrium}, stable in size while continually regenerated---a \emph{design property}, since \textit{a system that stopped proposing structure would be unable to learn genuine novelty} should it ever arise---and its arrival is not signalled, as we assume no visible task boundaries. As expected from the single-target \csv{} design choice, negative structure comes to dominate ($105$ vs.\ $82$ positive at cycle $19$) as redundant representations from other classes serve as suppressors for multiple confusable categories (Secs.~\ref{app:morefigs},~\ref{app:variants},~\ref{app:tables}).

We confirm this directly: injecting a previously-unseen class after the model has saturated causes it to grow further and learn the new class, so a stable population size is not a depleted one---the mature model retains the capacity to adapt and add structure on demand (Modeller-only; the experiment and its details are in Sec.~\ref{app:novelty}).

The read-out makes the \textit{stable mature core} vs.\ \textit{transient hypothesis turnover} split precise. Ignoring every \csv{} with fewer than $M$ lifetime presence observations leaves accuracy flat ($0.874$ at $M{=}0$, $0.879$ at $M{=}10$, $0.874$ at $M{=}40$; ten seeds) while at $M{=}40$ only $92$ of $187$ \csv{}s remain: \textbf{half the learned population can be discarded at read-out time with no loss of accuracy.} The \csv{}s below the threshold are the recently formed and rarely satisfied: the model is a mature core carrying essentially all the discrimination, plus a provisional reserve of transient hypotheses that provides the capacity to adapt to genuine novelty if needed (Sec.~\ref{app:growth}).

Across the stream regimes of Sec.~\ref{sec:exp-regimes} the population size proves \emph{demand-driven} (Fig.~\ref{fig:sizeregimes}, Sec.~\ref{app:regimes}). The equilibrium property holds in every regime---final size stays within $0.90$--$0.95$ of peak throughout, never running away---but the level it settles at varies almost fivefold with how much discrimination the schedule forces: ${\sim}185$--$190$ \csv{}s under cyclic, shuffled and permuted, but only ${\sim}41$ under \emph{blocked}. The small blocked model is not economy but under-differentiation: without interleaving, a class is never presented alongside the classes it must be told apart from, so the cross-class conflict that drives refinement never arises---the same cause as its lower accuracy there. \emph{longgap} interpolates, growing in delayed steps as each round introduces its classes (Sec.~\ref{app:regimes}). The same demand-driven pattern holds on EMNIST: with twenty-six classes to tell apart, the population settles at $\sim\!484$ \csv{}s. Structure is thus elicited in proportion to the discrimination the stream demands, and always stabilises.
\subsection{Scaling} \label{sec:exp-scaling} \paragraph{With the number of classes.} We reran the identical learner and stream on MNIST with the number of classes varied from two to ten, drawn at random per seed (Table~\ref{tab:scaling}), measuring per-observation cost as calls to the matcher---the only expensive step, since a \csv{} whose downstream chain did not match is discharged by a constant-time test before any matching, and learning reaches \csv{}s by the same chain-aware pass, so a \csv{} skipped at read-out is skipped when learning too. From four to ten classes ($2.5\times$ the classes) the model grows $2.80\times$ but the work per observation grows only $2.09\times$: the fraction of the model actually matched (traversed) falls from $0.59$ to $0.43$ and is flat from six classes on, so roughly half the structure is skipped before any matching and that proportion holds as the model grows. Accuracy does not degrade as classes are added ($0.833$--$0.878$). The series continues within EMNIST: from ten to twenty-six EMNIST classes, a $2.6\times$ increase, the model grows from $\sim\!264$ to $\sim\!484$ \csv{}s, a factor of $1.84\times$---again sublinear in the class count, so the demand-driven growth characterised in Sec.~\ref{sec:exp-structure} holds on EMNIST as well. On MNIST, the per-sample \emph{worst} case, however, approaches full traversal ($199$ of $216$ \csv{}s in the largest run): the mean is the right summary, but the tail is near-full. Note that the current implementation does \textit{not} provide any cap, cutoff, or heuristic prioritization (e.g. by statistical significance or relevance) to the traversal; these alternatives can be considered for the mitigation of worst-case scenarios should specific applications require them. As noted before, \csv{} counts are \emph{additionally inflated} beyond what the discrimination requires by each \csv{} conditioning a \emph{single} target, since structure shared between classes is rebuilt separately under every class that uses it rather than referenced once. Allowing a \csv{} to condition several immediate targets (a directed acyclic organisation rather than a per-target tree) would let each shared sub-pattern exist once and be reused across classes, shrinking the model at equal competence; this is a change to the model's organisation, separate from the representation and learning this paper contributes, which we leave to future implementation.

\begin{table}[tb]
\centering
\caption{Scaling with the number of classes (four seeds each; classes drawn at random per seed).
Matched is the number of \csv{}s that reach the matcher for one observation, range
its per-sample minimum and maximum across all seeds, and frac the mean fraction of the
model matched. The two-class row (italic) measures which digit pair was drawn rather than genuine
complexity---its four seeds gave $5$, $16$, $31$ and $57$ \csv{}s---and should not be read as
representative of class-count scaling.}
\label{tab:scaling}
\small
\setlength{\tabcolsep}{4pt}
\begin{tabular}{lccccc}
\toprule
Classes & \csv{}s & Matched & Range & Frac & Accuracy \\
\midrule
$\mathit{2}$ & $\mathit{27}\!\pm\!\mathit{23}$ & $\mathit{23}\!\pm\!\mathit{18}$ & $\mathit{3}$--$\mathit{57}$ & $\mathit{0.88}$ & $\mathit{0.869}\!\pm\!\mathit{0.072}$ \\
$4$ & $68\!\pm\!15$ & $40\!\pm\!7$ & $10$--$74$ & $0.59$ & $0.878\!\pm\!0.057$ \\
$6$ & $124\!\pm\!2$ & $58\!\pm\!1$ & $16$--$123$ & $0.47$ & $0.833\!\pm\!0.065$ \\
$8$ & $145\!\pm\!12$ & $68\!\pm\!10$ & $20$--$150$ & $0.47$ & $0.842\!\pm\!0.025$ \\
$10$ & $191\!\pm\!18$ & $83\!\pm\!5$ & $24$--$199$ & $0.43$ & $0.874\!\pm\!0.014$ \\
\bottomrule
\end{tabular}
\end{table}

\paragraph{With input complexity.} A second axis is how structure grows with the \emph{complexity of each input}---the network or semantic complexity of an image (how much structure it takes to describe, not the absolute number of pixels since they are not visible to the learner). Within a dataset the two do \emph{not} track each other (on MNIST, Spearman $\rho\!\approx\!0$; Sec.~\ref{app:scaling}): the most structurally complex digit---an $8$ (two holes, $16.6$ nodes on average)---grows the \emph{fewest} \csv{}s (${\sim}4$), whereas the digits that grow the \emph{most} (${\sim}30$--$42$) are the confusable $6$, $2$ and $5$. What structure a class grows is set by how much it must be \emph{told apart} from the others---confusable classes accumulate large suppressor stacks---not by how much structure its images contain; the same holds within EMNIST ($\rho\!\approx\!0.15$; Sec.~\ref{app:scaling}). A fuller study across a wider range of input complexities is left for future work, alongside augmenting the method to process higher orders of change, 2D intensity surfaces, and 3D shapes.

\subsection{Ablating Multi-Scale Representation} \label{sec:exp-ablation} The central design claim is that carrying structure at multiple scales in one network is what enables high-performance matching. Retraining the identical system on the \emph{finest scale only}---no coarsening, no cross-scale relations, everything else fixed (MNIST, five seeds)---drops accuracy from $0.874$ to $0.730$ under the geometric read-out at a similar model size ($187$ vs.\ $172$ \csv{}s): a difference in what the structure can \emph{represent}, not in capacity, because when fine structure is refined away and no coarser relation spans the gap the structural continuity is never formed, leaving nothing to read. The full $2\times2$ ablation of representation and read-out, and the remaining ablations, are in Secs.~\ref{app:ablation},~\ref{app:supporting}, \ref{app:readout} and~\ref{app:variants}.

\subsection{Preservation of Past Responses} \label{sec:exp-preserve} The framework's defining property is that learning preserves the model's responses to observations it has already accounted for---a guarantee about the learning \emph{operation} (Theorem~1 of~\citep{erden2025foundations}; Sec.~\ref{sec:method}). The assembled system can depart from perfect preservation only through mechanisms that sit \emph{outside} that operation---matching (which source variables are found present), online re-estimation of the geometry the read-out consumes, and reintegration composing structure back in---a graceful deviation we now measure directly.

On MNIST (Sec.~\ref{app:preserve}), re-checking a fixed set of already-seen observations after every block shows that a \csv{} that fully matched an observation still fully matches it in ${\sim}97\%$ of checks over the whole run (Table~\ref{tab:preserve}, per seed). The ${\sim}3\%$ that lapse are predominantly near-misses---partial or majority matches that fall just short of a full match---rather than lost responses, and split roughly evenly between online spatial re-estimation (the running position estimates drift) and structural change (a unit's own source altered, or a downstream re-pinned), consistent with reintegration composing structure in. Predicted labels for already-seen observations are likewise ${\sim}97\%$ stable between consecutive blocks. In practice, then, the mechanisms deliver local preservation at or near the level the theory describes for the operation it governs.

\subsection{Comprehensibility of the Learned Structure} \label{sec:exp-interp} 
\begin{figure}[t]
\centering
\includegraphics[width=0.74\columnwidth]{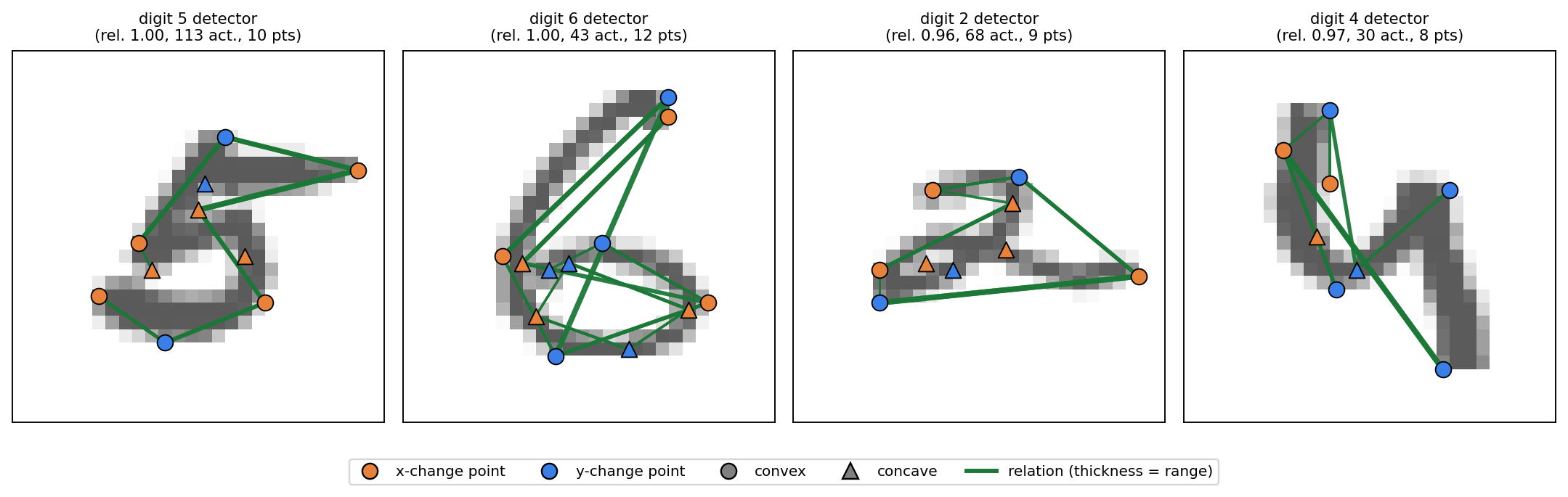}
\caption{The learned models are their own explanation. Mature \csv{}s for four MNIST
classes, each overlaid on a held-out image it matches. Nodes are typed
orientation-change points landing on meaningful loci; green edges are retained
cross-scale relations (thickness $\propto$ range), coarse ones carrying global
layout and fine ones local detail. No post-hoc attribution is involved: this
\emph{is} the learned representation.}
\label{fig:interp}
\end{figure}

Because a \csv{} is an explicit configuration of typed change points and relations it can simply be drawn, each mature one reading as a caricature of its class (Fig.~\ref{fig:interp}). What is drawn is not a post-hoc attribution but the learned representation itself: the nodes are typed orientation-change points that land on meaningful loci of the shape, and the retained cross-scale relations carry global layout at coarse range and local detail at fine range. Because the units are explicit, one can also read off which \csv{}s condition which---the chain structure---and so diagnose \emph{why} a class underperforms: which units it relies on, and which distinctions it lacks. The same transparency carries to EMNIST and there exposes finer structure: a letter's mature \csv{}s again read as caricatures, and because the units are explicit one can see \emph{where} case information is carried---some units fire on both forms of a letter, others on only one, at depths that differ by letter (Sec.~\ref{app:emnist-interp}, Fig.~\ref{fig:emnist-interp}).

\begin{figure}[tbp]
\centering
\includegraphics[width=0.68\columnwidth]{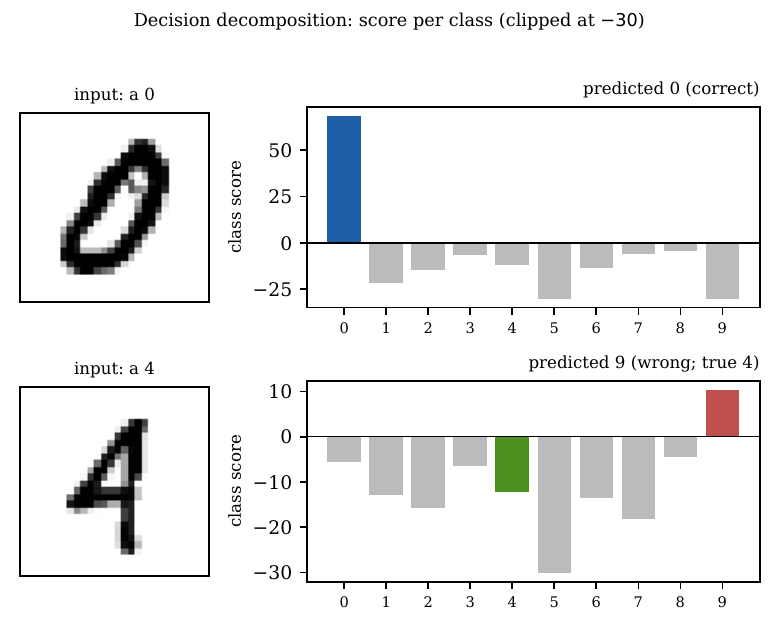}
\caption{A prediction decomposed into per-class scores (clipped at $-30$ for legibility; the
winning class is blue when correct, red when wrong, the true class green). \textbf{Top:} a
confident $0$---five class-$0$ \csv{}s along one chain fully match and sum to $+68$, every other
class sunk by absence. \textbf{Bottom:} a $4$ read as $9$---two class-$9$ \csv{}s fully match
(chains $272|21$ and $106|22$, $+11.4$ and $+9.6$) while \emph{no} class-$4$ \csv{} matches at
all ($0$ of $21$ fired), so the decision is $9$ by a narrow margin. The failure is legible at the
unit level, not a black-box error.}
\label{fig:worked}
\end{figure}

This is not only inspectability of the units but traceability of the \emph{decision}. Each prediction is a sum of signed contributions, one per \csv{}, read directly (Fig.~\ref{fig:worked}). For a clearly-written $0$ the support is a single chain of five class-$0$ \csv{}s ($+68$ across the firing, position and orientation terms) against only absence evidence elsewhere---clear-cut and nameable. The instructive case is a failure. A $4$ misread as $9$ is not diffuse: two class-$9$ \csv{}s fully match it (their upper structure, a loop over a stem, is present) while \emph{not one} of the twenty-one class-$4$ \csv{}s does, so the score is $+10$ for $9$ against $-12$ for $4$. The read-out exposes exactly this---which units fired, for which class, and by how much---so the $4/9$ confusion is diagnosable as the shared loop-and-stem being satisfied for $9$ and unmet for $4$, rather than an uninterpretable error. The confusion has a representational root: a closed $4$ without the extension of its horizontal line is identical to a $9$ in its \emph{first-order} change points, and what would separate them is the second-order boundary structure of the loops (smoother in $9$ vs.\ sharply cornered in $4$) that the current first-order vocabulary does not encode---the feature-vocabulary limit discussed as the system's main residual (Secs.~\ref{sec:changepoint} and~\ref{sec:conclusion}).

\section{Related Work} \label{sec:related} \subsubsection{Multi-scale and scale-space representations.} Describing visual structure across a hierarchy of scales---coarse structure stable and general, fine structure specific---is classical: scale-space filtering \citep{witkin1983scale}, automatic scale selection \citep{lindeberg1998feature}, image pyramids \citep{burt1983laplacian}, and wavelet multiresolution \citep{mallat1989theory}. Our representation inherits the principle but realizes it structurally rather than signal-analytically: we coarsen a \emph{relational graph} and, crucially, do not keep the scales as a stack of separate descriptions---all levels are accumulated into a \emph{single unified network} in which nodes are shared and edges from every level coexist, so relations are expressed both within and \emph{across} scales in one structure. Graph pyramids \citep{kropatsch1995building} are the closest graph-domain precedent, but build hierarchies of successively contracted graphs; we collapse the hierarchy into one augmented network serving a continual structural learner. Local descriptors such as SIFT \citep{lowe2004distinctive} and shape context \citep{belongie2002shape} reduce a shape to fixed-length vectors compared against stored exemplars, refining at most the alignment of a single pair rather than a model retained across samples. More fundamentally, across this body of work the multi-scale representation is a \emph{front end}: features are computed and handed to a \emph{separate} learner. We are not aware of a method in which a multi-scale representation is integral to a \emph{non-gradient, structural} learner that learns \emph{by refining that very representation}.

\subsubsection{Hierarchical visual architectures.} Layered abstraction of visual features runs from Marr's program \citep{marr1982vision} through the Neocognitron \citep{fukushima1980neocognitron} and HMAX \citep{riesenhuber1999hierarchical} to modern deep networks, which realize it as gradient-trained filter banks with a monolithic internal representation and the destructive-adaptation difficulties that motivate this work. Our hierarchy is instead explicit, gradient-free, and inspectable at every level.

\subsubsection{Structural shape representation and matching.} Representing shape as attributed relational structure and recognizing by graph matching is a long tradition: error-tolerant attributed-graph matching \citep{messmer1998new} and shock graphs \citep{siddiqi1999shock} do exactly this, and in cognitive science the object-centred structural descriptions of Marr and Nishihara \citep{marr1978representation} and recognition by components \citep{biederman1987recognition} give the same idea. Deformable part-based models \citep{felzenszwalb2010object} retain the relational arrangement but with learned appearance filters in place of symbolic parts, inferring the model directly against the image rather than matching two graphs; more recently, neural models learn representations over graph-structured data directly \citep{kipf2017semi}. Our state polynetworks sit in this tradition, but matching is reframed as a step \emph{inside online learning}: refinement incrementally computes, with statistical tolerance and a preservation guarantee, the shared substructure of a unit's recurrences, rather than a one-shot comparison or a gradient-trained embedding.

A standing limitation of the symbolic and structural tradition is its reliance on primitives fixed correctly in advance: where the parts or features are hand-engineered per domain, the representation is only as good as that engineering. Here, learning via local variation and selection lifts this dependence without giving up explicit structure---the primitives here are not task-tuned but a single over-complete vocabulary (typed points of change, applicable to any continuous signal; Sec.~\ref{sec:changepoint}), and experience determines online which relations and compositions over them persist. The representation is thus engineered once from a universal principle rather than crafted per task, with what survives fixed by the data, not the designer.

\subsubsection{Compositional and non-gradient concept learning.} Bayesian program learning \citep{lake2015human} shares our emphasis on structured, comprehensible concept models but is generative over a hand-designed compositional grammar, with its primitives learned from a background set of alphabets, and is not framed for lifelong integration; capsule networks \citep{sabour2017dynamic} pursue part--whole structure with gradient training. Outside the gradient paradigm, Adaptive Resonance Theory \citep{carpenter1987massively} treats the stability--plasticity dilemma via a vigilance test---analogous to our refinement tolerance $\Tref$---while incremental concept formation \citep{fisher1987knowledge} and growing self-organizing models \citep{fritzke1995growing}, which extend the fixed-lattice self-organizing map \citep{kohonen1982self}, insert units where the data demand them, as our variation process does. We differ in learning explicitly relational, multi-level structure with a preservation guarantee.

\subsubsection{Continual learning in machine learning and in vision.} Class-incremental learning is active: exemplar replay \citep{rebuffi2017icarl,buzzega2020derpp,caccia2022erace}, distillation \citep{li2018lwf}, regularization \citep{kirkpatrick2017overcoming,zenke2017si,aljundi2018mas,aljundi2019taskfree}, and expansion or expert-routing \citep{aljundi2017expertgate,erden2025aec}; see \citep{delange2021continual,vandeven2024continual,parisi2019continual} for surveys. Each family buys retention at the cost of an assumption online learning does not grant: replay stores and re-presents past data; expansion isolates per-task sub-models behind a routing signal. Distillation and regularization are the mildest case, and our measurements sharpen the picture (Sec.~\ref{app:clbaselines}): they require not \emph{task boundaries} as such but a \emph{consolidation schedule}---some designated moment at which the parameters currently held become the ones to protect---and they fail at both its extremes, are acutely sensitive to penalty strength (numerically inert at the values usually quoted for larger models), and retain poorly, decaying to chance once recurrence weakens (Sec.~\ref{sec:exp-regimes}). The dominant recent direction, prompt-based continual learning over a frozen pretrained backbone \citep{wang2022l2p,wang2022dualprompt}, leads current benchmarks but presupposes large-scale pretraining, supplying from the outset a representation of the kind whose \emph{formation} from a stream is what this work studies---a different question, not a limitation of that line. Two recent class-incremental methods are likewise gradient-free but rest on assumptions our setting removes: F-OAL \citep{zhuang2024foal} fits only an analytic classifier on top of a frozen pretrained backbone, while incremental rate reduction \citep{wu2021ratereduction} builds its network in closed form from per-class covariances, assuming disjoint tasks each presented once as a representative (i.i.d.) batch. Neither forms its representation online, one sample at a time and without task boundaries, as ours does.

\section{Conclusion and Limitations} \label{sec:discussion} \label{sec:conclusion}

\subsubsection{Limitations.} Evaluation is confined to a centred $2$D shape domain---a proof of concept for the underlying \textit{learning process} (Sec.~\ref{sec:exp-protocol}), not a general visual recognizer. What stands between it and a conventionally trained network is chiefly the feature vocabulary: first-order change points on a binarized contour discard most of the image---most sharply in the $4/9$ confusion on MNIST and the lower absolute ceiling on EMNIST, where letter classes that share first-order structure cannot be fully told apart. Two further limitations are structural: each \csv{} conditions a single target, so structure shared between classes is rebuilt per class and the model is larger than the discrimination requires (Sec.~\ref{sec:exp-scaling}); and the read-out uses absolute node position, which suits centred MNIST but degrades accuracy under large test-time translation (Sec.~\ref{sec:exp-transform}). Finally, wall-clock cost is dominated by graph matching, and the worst-case per-sample traversal approaches the full model even though the mean is about half of it (Sec.~\ref{sec:exp-scaling}). As discussed there, such worst cases can be mitigated if needed (e.g.\ by capping or prioritizing the currently uncapped traversal), as the mean traversal remains a decreasing fraction of the full model as the number of classes grows.

\subsubsection{Future work.} Several extensions follow the same principle without altering the learner. The feature vocabulary can be deepened by taking \emph{higher orders of change}; the construction can be applied to the un-thresholded image as an intensity \emph{surface} rather than a traced contour (Sec.~\ref{sec:changepoint}), which would also lift the topological filtering; and it extends to $3$D feature points, which the learner would process identically. Two further changes are read-out-level and leave the learner untouched: invariance to translation and scale, through a position-free read-out or stream augmentation (Sec.~\ref{sec:exp-transform}); and reducing the graph-matching cost that dominates wall-clock time. Restoring the multi-target, shared-subgroup conditioning of the base methodology \citep{erden2025foundations,erden2025thesis} would remove the per-class duplication and shrink the model at equal competence. Beyond the representation, the most consequential step is hierarchical, compositional structure: the construction here operates \emph{within} a single shape, and the same principle points one level up---to part--whole organization across entities, and ultimately to integration with deliberative behaviour for embodied settings such as robotics \citep{erden2025foundations,erden2025thesis}.

\subsubsection{A developmental-systems perspective.} More broadly, we read this work as an instance of \emph{developmental} learning---competence accumulated from an ongoing stream one observation at a time and preserved as it accumulates, rather than fitted to a fixed dataset and overwritten. Three of its properties are exactly those an autonomously developing agent needs: it learns online and open-endedly (no task boundaries, replay, or designated consolidation moment), so it can keep integrating experience as the world presents it, not as a curriculum arranges it; it retains earlier competence under a local guarantee as later competence is acquired (the stability--plasticity balance development requires); and its models are legible---each learned unit an explicit, drawable structure, so what the agent knows and why it errs can be inspected rather than inferred. This makes the approach a natural substrate for \emph{perceptual development} in embodied settings, where perception is acquired through interaction and must persist and stay interpretable over a lifetime \citep{erden2025foundations,erden2025thesis}; the present system demonstrates it for 2D shape, and the program is to carry the same process up through richer representations and outward into agents that perceive by developing.

\subsubsection{Summary.} We extended a developmental, gradient-free structural learner to visual shape recognition by designing the representation it needed: a multi-scale structural encoding that unifies all scales and their cross-scale relations in one network, with a local, absence-aware read-out. On two class-incremental shape benchmarks---MNIST and the harder EMNIST-Letters---it learns each class one sample at a time from a replay-free, boundary-free stream that no baseline learns in and under which they all lose accuracy, matches or exceeds every replay-free baseline while storing no past observations, and does not give up what it learns: continual learning in the sense meant, at an accuracy that makes it a genuine recognizer. Its central strength is preservation, seen through within-cycle retention---how much of a class survives the learning of others. Across the regime spectrum, on both datasets and with the margin wider on the harder one, the Modeller holds most of each class, whereas the regularisation baselines, whose apparent continual learning is really learning \emph{with} re-exposure, retain little once recurrence weakens and decay to chance under a single pass---where the Modeller still retains a considerable portion and degrades gracefully: not immunity to forgetting, but what the \emph{local} preservation guarantee (Sec.~\ref{sec:exp-preserve}) promises. The approach \emph{presupposes} its representation rather than learning it from data---a front-loaded cost, where a gradient-trained network instead pays with sample volume, repeated exposure and destructive adaptation---but incurs it once, as a universal principle (Sec.~\ref{sec:changepoint}) rather than per-task engineering, so experience, not feature design, fixes what persists. The structure it builds is an observable of learning rather than a fixed budget: it grows and then saturates, about half can be discarded at read-out with no loss of accuracy, and its size is demand-driven. And because each learned unit is an explicit configuration that can simply be drawn, the model is its own explanation---what it retains, and where it fails, can be read off directly.

\putbib
\end{bibunit}

\startsupplement{Multi-Scale Structural Features for Continual, Comprehensible Visual Recognition in a Developmental Learning Framework}

\begin{bibunit}
\nocite{IEEEexample:BSTcontrol}%

\section{The Coarsening Algorithm} \label{app:algs} The coarsening procedure described in Sec.~\ref{sec:representation} is given here in full; the matching procedure it and the learner rely on is Algorithm~\ref{alg:match} in the main text (Sec.~\ref{sec:matching}).

\begin{algorithm}[t]
\caption{Multi-scale structural representation of a shape}
\label{alg:coarsen}
\begin{algorithmic}[1]
\REQUIRE binary image $I$
\STATE extract the contours of $I$, distinguishing outer boundaries from holes
\STATE $N \gets$ directional change points of every contour, each typed by change
axis, direction, and local convexity
\STATE $E \gets$ contour edges between successive change points, plus spatial edges,
in horizontal and vertical layers
\STATE $\mathcal{A} \gets E$ \COMMENT{accumulator: the edges of every level}
\WHILE{some pair of the kind below remains}
  \STATE among the nodes of each type, taken in contour order, collect every
  consecutive pair $(u,v)$ whose subtypes are one maximum and one minimum
  \STATE \textbf{if} there is no such pair \textbf{then} exit
  \STATE $(u,v) \gets$ the pair of least extent---measured along $x$ for $x$-typed
  nodes, along $y$ for $y$-typed ones, and as the distance between them otherwise
  \COMMENT{an adjacent maximum and minimum lying close together is the smallest
  remaining excursion of the contour}
  \STATE remove \emph{both} $u$ and $v$, reconnecting the surviving contour
  neighbours of each
  \STATE recompute the spatial layers at this level, giving edge set $E$
  \STATE $\mathcal{A} \gets \mathcal{A} \cup E$
\ENDWHILE
\STATE \textbf{return} augmented network $(N, \mathcal{A})$: all nodes, and the
edges of every level in parallel
\end{algorithmic}
\end{algorithm}

\section{Differences from the Prior Instantiation} \label{app:differences} This section records, for readers familiar with the prior work \citep{erden2025foundations,erden2025thesis}, the points at which this system differs from the mechanism as described there; it can be read in place of Sec.~\ref{sec:method}.

\subsubsection{Rerelation removed.} The prior algorithm (\emph{network refinement with rerelation}) created bridging edges between predecessors and successors of removed elements. Here the algorithm is plain \emph{network refinement}: removal of non-recurring nodes and edges only, identical to the prior algorithm but operating on node/edge lists rather than lists of state variables. Relational continuity across removals is guaranteed by the representation instead---the multi-scale augmented network already contains the coarser-scale edges spanning any removed fine structure (Sec.~\ref{sec:representation}). The prior statistical absence-ratio criterion for noise tolerance is likewise replaced by coverage-thresholded matching (a poorly-matching source is treated as absent, so it is not stripped). The preservation guarantee is unaffected: removal-only refinement is strictly more conservative.

\subsubsection{Composed sources and anchors.} In the prior vision instantiation an upstream \csv{} is a \emph{subvariant} of its downstream: its source network \emph{encompasses} the downstream's, so a chain is a sequence of ever-larger patterns, each containing the one below it. Here a chain is instead a \emph{composition} of distinct sources. An upstream's source holds only the structure it itself contributes, plus \emph{anchors}---nodes that are not owned by it but are references to nodes owned by its downstream---and the boundary edges that join the two. Ownership is explicit and single-valued: each node is owned by exactly one \csv{}, and being an anchor is a role a node plays \emph{in another source}, not a property of the node, so the same node is ordinary content for its owner and an anchor for the upstream that references it. The same construction is used whether the new structure comes from refining a partial match (the residual that was missing) or from generating a conditioner for the part of an observation the chain did not cover (the remainder); in both cases only the new structure is stored, anchored to what is already represented.

Three consequences follow. First, structure within a chain is stored once rather than restated at every level, so the \csv{} counts reported in Sec.~\ref{sec:exp-structure} measure disjoint contributions rather than repeated copies. Second, matching is cheaper: matching a chain downstream-first binds an upstream's anchors as a side effect of its downstream's placement---no separate search, and no walk back down the chain, since an anchor is literally the downstream's node---so the work at each level is proportional to the new structure at that level. Anchors are consequently excluded from a \csv{}'s coverage, which therefore measures only its own content. Third, each \csv{} is readable on its own as what that level adds, which is what Fig.~\ref{fig:interp} displays.

The arrangement does impose an invariant: every anchor of a \csv{} must remain owned by some \csv{} in its own downstream chain, since an anchor pointing outside that chain will never be bound and the upstream can then no longer match at all. Refinement, removal, absorption and reintegration all move or delete nodes, so each has to preserve it.

\subsubsection{Feature extraction.} The prior description extracted polygonal approximations of contours; here no polygonal approximation or smoothing is applied---contours are traversed directly and orientation-change points read off the raw contour, typed by axis, direction, and convexity. The multi-scale coarsening and the accumulation of all levels into one augmented network are new in this work.

\subsubsection{Single-target \csv{}s and externalized negatives.} In the base methodology---the non-visual formulation---units condition multiple targets, sharing source subgroups across them, and negative evidence is carried as negative sources inside a unit. This implementation instead assigns each \csv{} one target (duplicating sources across classes as needed) and represents negative evidence as separate negative \csv{}s. It should be noted that this is not a departure introduced here: the visual formulation of the prior work already made both of these choices, so on this point the present system follows it rather than differing from it. Both remain implementation economies; the multi-target, shared-subgroup mechanism would require a substantial refactor beyond the scope of this paper's contributions. The visible cost is redundancy---structure shared between classes is stored per class, inflating model size (Sec.~\ref{sec:exp-structure}) and forgoing cross-class reuse; we expect the multi-target mechanism to reduce the \csv{} population substantially at equal competence.

\subsubsection{Statistics and maintenance.} Presence statistics, reliability and $P(s\mid t)$ follow the prior design, as does removal. \emph{Reintegration} has no counterpart there and is introduced here; without it, upstreams that have ceased to carve any distinction accumulate as permanent structure. Both operations read the same underlying statistic, $\Pst$, and are stated exactly here since the main text gives only their sense. In this expression $\mathrm{SS}(C)$ is the event that conditioner $C$'s sources are \emph{satisfied}---its source configuration is matched in full by the observation---and $I(T)$ the event that its target $T$ is in the state $C$ conditions toward, that is, active for a positive conditioner and inactive for a negative one. So $\Pst$ reads: given that the target came out the way this conditioner speaks for, how often was the conditioner in fact present. Reintegration applies on the high band of $p=\Pst$, that is $p>\Treint$; removal applies on the low band of the conditioner's \emph{chain-effective} significance $\hat p$, that is $\hat p<\Trem$. By default (depth-scaled significance; Sec.~\ref{app:repro}) $\hat p$ is $\Pst$ diluted by the downstream chain---the product of $\Pst$ along the links from the conditioner to its external target---so it equals $\Pst$ for a top-level conditioner and is smaller for a deeper one, which makes deep upstreams that no longer carve a distinction easier to prune. Each fires on a given step with probability
\begin{equation*}
\rho_{\mathrm{rem}}\frac{\Trem-\hat p}{\Trem}
\qquad\text{and}\qquad
\rho_{\mathrm{int}}\frac{p-\Treint}{1-\Treint}
\end{equation*}
respectively, and zero outside its band. Both probabilities rise linearly from zero at the band edge to their peak at the extreme value, so removal and reintegration have the same shape, mirrored around the middle of the range. A single width parameter $\esign$ places the two thresholds symmetrically about $\tfrac12$: removal fires in the \emph{low} band $\hat p<\Trem$ with $\Trem=\esign$, and reintegration in the \emph{high} band $p>\Treint$ with $\Treint=1-\esign$. The two significance measures coincide for a top-level conditioner ($\hat p=p$); for a mid-chain conditioner, removal uses the chain-diluted $\hat p$ while reintegration uses the plain $p$. Two further conditions apply to reintegration only: the conditioner must be \emph{positive}, since folding suppressive structure into what it suppresses would combine opposing evidence, and it must condition another \csv{} rather than a class node, since a top-level \csv{} has no pattern to be folded into. Neither operation acts on a \csv{} on its birth step---a freshly spawned upstream inherits $p\approx1$ and carries no independent evidence yet---and one whose relation gained no evidence on the current step has its probability down-weighted (Sec.~\ref{app:repro} gives the values). Chain-aware anchored matching (Sec.~\ref{sec:matching}) is used at both learning and prediction time. The strictly local geometric read-out (Sec.~\ref{sec:readout}) is likewise new; a simpler presence-only read-out is reported alongside it in Sec.~\ref{app:readout}.

\subsubsection{Refinement and upstream spawning.} Algorithm~\ref{alg:refine} expands the two steps at the centre of Algorithm~\ref{alg:learn}'s traversal loop---the refinement of a \csv{}'s source and the spawning of the upstream that carries what refinement removed---since this is where anchors originate and where the composed-source arrangement above is actually established.

\begin{algorithm}[t]
\caption{Refinement and upstream spawning for one \csv{}}
\label{alg:refine}
\begin{algorithmic}[1]
\REQUIRE \csv{} $C$ with source $S$; observation $O$; the best match $\phi$ of $S$ into
$O$; coverage threshold $\tau$
\IF{$\mathrm{degree}(\phi) < \tau$}
  \STATE treat $C$ as absent and leave $S$ unchanged; \textbf{return}
  \COMMENT{a poor match must not strip structure}
\ENDIF
\STATE $(R,A) \gets$ the part of $S$ that $\phi$ left unplaced, together with the placed
nodes whose edges reach into it \COMMENT{$R$: residual nodes and edges, identities
preserved; $A$: the boundary nodes, to become anchors}
\STATE $S \gets \phi(S)$ \COMMENT{refinement: $S$ shrinks to the placed part, which
$\phi$ now matches fully}
\IF{$R$ is non-empty}
  \STATE create $U$, a positive \csv{} conditioning $C$, with source $R\cup A$ and $A$
  flagged as anchors
  \STATE $U$ inherits $C$'s presence counts, so its statistics begin from $C$'s history
  rather than from nothing
  \STATE re-parent onto $U$ every conditioner that previously conditioned $C$
  \COMMENT{they conditioned the larger configuration $U$ now represents}
  \STATE transfer ownership of $R$'s nodes to $U$; the anchors $A$ stay owned by $C$
\ENDIF
\end{algorithmic}
\end{algorithm}

\section{Read-Out Details} \label{app:readout-details} The read-out is strictly read-only: it inspects the learned structure, statistics, and matcher and computes a prediction, mutating nothing.

\subsubsection{Presence.} Presence is computed once per observation, by a single target-free pass over all \csv{}s using the matching procedure of Sec.~\ref{sec:matching}. Two quantities are recorded per \csv{}: a coverage \emph{degree} in $[0,1]$---the fraction of its own nodes and edges placed, with edges counted so that the same nodes under a different connectivity are not treated as the same match---and whether the match is \emph{full}. A \csv{} whose source does not match, or whose downstream is not present, counts as absent.

The pass is ordered \emph{downstream-first}, and this is what keeps its cost from scaling with the size of the model, so it is worth being explicit that no \csv{} is matched independently of its chain. Because an upstream meets its downstream only at anchors (Sec.~\ref{app:differences}), a \csv{} whose downstream failed to place has nowhere for its anchors to be bound, and is settled as absent \emph{without the matcher being invoked on it at all}. A chain is therefore descended only as far as the observation supports it, and one failure discharges the whole branch above: the matching work is proportional to the structure the image actually evokes, not to the number of \csv{}s the model holds. What does range over every \csv{} is the scoring sum, since an absent \csv{} still contributes its absence term---but that is a few arithmetic operations each, not a match. Discarding absent \csv{}s rather than scoring them would therefore save nothing that matters and would yield a different, weaker read-out (Sec.~\ref{app:readout}).

\subsubsection{Polarity.} The read-out never inspects a \csv{}'s polarity. A suppressor nonetheless enters with the correct sign automatically: because it is a unit that fires when its own class is \emph{absent}, its own-class firing rate falls below its foreign rate, $P^{o}_c<P^{f}_c$, and its present-term $\log(P^{o}_c/P^{f}_c)$ is therefore negative---a veto---while its absent-term is positive. No separate handling of negative structure is required.

\subsubsection{Matching policy.} Anchored (pinned) matching can be replaced by \emph{unified} matching, in which each \csv{}'s whole chain is matched fresh as one pattern; on the final models it scores slightly below anchored. Anchored matching is used throughout.

\section{Ablating the Geometry: a Presence-Only Read-Out} \label{app:readout} This section reports a second, deliberately impoverished read-out. It is best read as an \emph{ablation of the geometric information} rather than as a competing method: it is given the same trained models and differs from the read-out of Sec.~\ref{sec:readout} in exactly two respects---it consults only \emph{which} \csv{}s matched, using none of the learned positions or orientations, and it treats a non-matching \csv{} as no evidence rather than as evidence against its class. Its lower accuracy is therefore the point of reporting it: the gap measures what those two sources of information contribute, on identical structure. Each class is scored through its positive top-level \csv{}s; a present \csv{} contributes its reliability $\rho$ and present upstream evidence is fused recursively (positive as support $p_i$, negative as suppression $q_j$) by summation in log-odds, with the reliability term pivoted at the uniform prior $1/K$ over $K$ classes so that a \csv{} aligned with its class only at the base rate contributes nothing:
\begin{equation}
\ell = \operatorname{logit}(\rho) - \operatorname{logit}(1/K)
+ \textstyle\sum_i \operatorname{logit}(p_i) - \sum_j \operatorname{logit}(q_j).
\label{eq:readout}
\end{equation}
Both read-outs are read-only and differ only in how they score a fixed trained model, so neither requires retraining. On the same final models this scheme reaches $0.697$, against $\mathbf{0.874}$ for the geometric read-out. Variations within it change little: ignoring suppressive structure gives $0.635$, a maturity filter that discards \csv{}s with fewer than $M$ lifetime observations moves it within one standard deviation ($0.697/0.700/0.704$ at $M{=}0/5/10$), and two further variants---a continuous \emph{graded} coverage scheme and a \emph{precedence} scheme---are dominated by the log-odds form. The maturity filter behaves the same way under the geometric read-out, and we report that measurement, which is the one the main text draws on, in Table~\ref{tab:maturity}. The spread among all of these is small compared with the difference to the geometric read-out, which indicates that what matters is not the particular decision rule but whether the learned spatial statistics and the informativeness of absence are consulted at all.

\section{Learning Variants} \label{app:variants} We report learning-side variants around the baseline; the baseline (with suppressors, geometric read-out) remains the main version. \textbf{Suppressors: dispensable here, but formed by design.} Preventing suppressor formation does \emph{not} hurt on MNIST---it is marginally higher, $0.8805$ vs.\ $0.8735$, a difference that is not significant over the ten paired seeds ($p=0.64$), at roughly a third of the model size ($58$ vs.\ $187$ \csv{}s)---because the discrimination MNIST needs is carried by the positive, geometry-bearing structure. This does not mean suppressors should be removed. The learner does not optimize our accuracy metric: it has no test set, receives no error signal, and never stops changing on the basis of accuracy, by deliberate design. It forms a complete-but-minimal model of what it has seen, and suppressors are part of that completeness; it keeps forming them because they may be discriminative for other data, even where---as on MNIST---they are not needed. \textbf{Position-gated matching.} Two variants gate matching or suppressor formation on the match centroid position; neither improves on the baseline ($0.8645$ and $0.8650$; $p=0.39$ and $p=0.34$, both non-significant), and a per-node position variant did not either---consistent with Sec.~\ref{sec:representation}: local position is not the lever for discrimination; global structure is.

\begin{table}[t]
\centering
\caption{Learning variants (ten-seed mean). Baseline is the main version; forgetting here uses end-of-cycle evaluations only ($0.097$, against $0.100$ with the per-block evaluation of Table~\ref{tab:regimes}).}
\label{tab:variants}
\small
\begin{tabular}{lccc}
\toprule
Variant & Accuracy & \csv{}s & Retention loss \\
\midrule
\textbf{Baseline} & \textbf{0.8735} & 187 & 0.097 \\
No suppressors & 0.8805 & 58 & 0.095 \\
Pos-gate & 0.8645 & 186 & 0.108 \\
Pos-skip & 0.8650 & 178 & 0.107 \\
\bottomrule
\end{tabular}
\end{table}

\section{Ablations of Supporting Mechanisms} \label{app:supporting} \subsubsection{Suppressive structure.} The role of suppressors depends on the read-out. Under the \emph{presence-only} read-out, re-scoring the final models with all negative \csv{}s ignored lowers accuracy ($0.697\!\to\!0.635$), and the effect is strongly class-specific: some classes depend on suppression to hold their boundary (most sharply $d8$, $0.77\!\to\!0.23$ when suppressors are removed), while others are over-suppressed and \emph{improve} without them (most sharply $d2$, $0.57\!\to\!0.74$). Under the \emph{geometric} read-out the picture inverts: preventing suppressor formation entirely (Sec.~\ref{app:variants}) is net neutral-to-positive ($0.874\!\to\!0.881$) at a third of the model size, because the geometry-bearing positive structure carries the discrimination the suppressors provided under the weaker read-out. Suppressors thus remain a designed, load-bearing mechanism in general, even where---as on MNIST with this read-out---they prove unnecessary. \subsubsection{Read-out policy.} Varying the maturity threshold changes final accuracy by less than one standard deviation, and unified matching scores marginally below anchored (Secs.~\ref{app:readout} and~\ref{app:readout-details}); read-out policy is a stable, secondary factor relative to the information source.

\section{The Multi-Scale Ablation in Full} \label{app:ablation} 
\begin{figure}[t]
\centering
\includegraphics[width=\columnwidth]{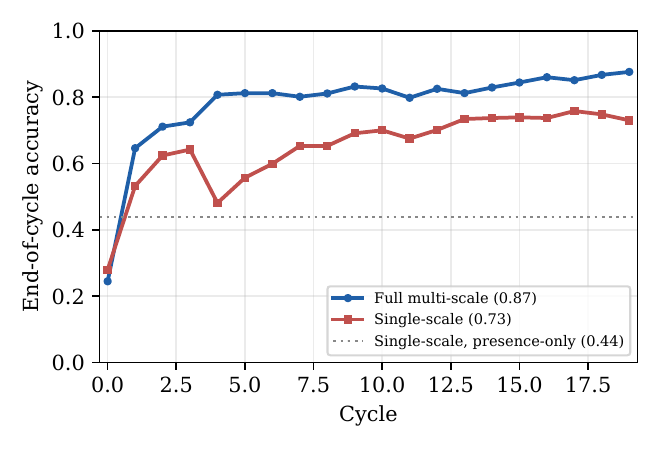}
\caption{Removing multi-scale structure. Retrained on the \emph{finest scale only}:
the full system (blue) reaches $0.87$; the single-scale system under the geometric
read-out (red) plateaus near $0.73$; under the simpler alternative read-out (dotted)
it reaches only $0.44$.}
\label{fig:ablation}
\end{figure}

Section~\ref{sec:exp-ablation} reports this ablation in brief; here we give the full $2\times2$ (Table~\ref{tab:ablation}), the per-cycle trajectories (Fig.~\ref{fig:ablation}) and one further quantity.

\begin{table}[tbp]
\centering
\caption{Multi-scale ablation: representation $\times$ read-out in one pipeline (five
seeds; the full multi-scale, geometric cell reproduces the ten-seed result of
Table~\ref{tab:regimes}). Final end-of-cycle accuracy; model size is comparable across
the representation rows ($187$ vs.\ $172$ \csv{}s), so the gap is not a capacity effect.}
\label{tab:ablation}
\small
\begin{tabular}{lcc}
\toprule
 & Presence-only read-out & Geometric read-out \\
\midrule
Full multi-scale & $0.700$ & $\mathbf{0.874}$ \\
Single-scale     & $0.439$ & $0.730$ \\
\bottomrule
\end{tabular}
\end{table}
Across the $2\times2$ the multi-scale gap under the geometric read-out ($0.730\rightarrow0.874$, i.e.\ $0.14$) is smaller than under the presence-only read-out ($0.439\rightarrow0.700$, i.e.\ $0.26$): the geometric read-out recovers part of the single-scale deficit but not the part that comes from structure the single-scale learner never formed, so the gap narrows without closing. That residual---structural continuity lost when refinement erodes fine detail with no coarser relation spanning the gap---is exactly what carrying multiple scales in one network is designed to prevent, and it is the component that cannot be repaired at read-out time.

\section{Statistical Tests} \label{app:stats} Seeds are paired across methods---a given seed fixes the sample draw and the order of presentation, so every method sees the identical stream---which makes a paired test the appropriate family. With ten seeds we do not assume normality, so we use the two-sided Wilcoxon signed-rank test on final end-of-cycle accuracy, one comparison per baseline (Table~\ref{tab:stats}).

\begin{table}[tb]
\centering
\caption{Two-sided Wilcoxon signed-rank tests on final accuracy (cyclic), our system against
each baseline, over the ten paired seeds of the regimes runs (Tables~\ref{tab:regimes}
and~\ref{tab:baselines-regimes}). Difference is ours minus theirs, so a
negative value means the baseline is ahead. The first block is the three comparisons
that do not favour us on this metric; seeds ahead counts the paired seeds on
which our accuracy is higher. Replay comparisons in italics. $^\dagger$CNN is from the separate
baseline sweep (Table~\ref{tab:full-baselines}).}
\label{tab:stats}
\footnotesize
\setlength{\tabcolsep}{4pt}
\begin{tabular}{lcccc}
\toprule
Comparison & difference & $W$ & $p$ & seeds ahead \\
\midrule
vs.\ SI & $-0.005$ & $20$ & $0.8203$ & $5/10$ \\
vs.\ MAS & $+0.009$ & $15$ & $0.2207$ & $7/10$ \\
\textit{vs.\ DER++ ($80$ img)} & $\mathit{-0.037}$ & $\mathit{1}$ & $\mathit{0.0039}$ & $\mathit{1/10}$ \\
\midrule
\textit{vs.\ DER++ ($8$ img)} & $\mathit{+0.077}$ & $\mathit{1}$ & $\mathit{0.0039}$ & $\mathit{9/10}$ \\
vs.\ LwF & $+0.065$ & $0$ & $0.0020$ & $10/10$ \\
vs.\ Plain NN & $+0.213$ & $0$ & $0.0020$ & $10/10$ \\
vs.\ Expert-AEC ($k{=}2$) & $+0.267$ & $0$ & $0.0020$ & $10/10$ \\
\textit{vs.\ ER ($8$ img)} & $\mathit{+0.148}$ & $\mathit{0}$ & $\mathit{0.0020}$ & $\mathit{10/10}$ \\
\textit{vs.\ ER-ACE ($8$ img)} & $\mathit{+0.496}$ & $\mathit{0}$ & $\mathit{0.0020}$ & $\mathit{10/10}$ \\
vs.\ CNN$^\dagger$ & $+0.547$ & $0$ & $0.0020$ & $10/10$ \\
\bottomrule
\end{tabular}
\end{table}

Three features of the outcome are worth stating plainly. First, for most baselines the test statistic is $W=0$: our system is more accurate on \emph{every one} of the ten paired seeds, which is arguably more informative than the $p$-value itself. Second, with $n=10$ the smallest attainable two-sided $p$ is $2/2^{10}=0.00195$, so several comparisons sit at the resolution limit of the test---they cannot be more significant than they are, and the test cannot distinguish ``clearly better'' from ``overwhelmingly better'' at this sample size. Third, and against us, two comparisons are \emph{not} significant and one is significant in the opposite direction. On final accuracy SI is indistinguishable from ours ($p=0.82$; it is ahead on four of ten seeds) and so is MAS ($p=0.22$), while DER++ with $80$ stored images has $W=1$ in its own favour, exceeding us on nine of ten seeds. Retention separates SI and MAS from ours clearly where final accuracy does not (Table~\ref{tab:wcr-main}); DER++ with $80$ images retains comparably, by replaying stored data.

We deliberately do not report significance for the multi-scale ablation (Sec.~\ref{app:ablation}) or the threshold sweep (Sec.~\ref{app:sensitivity}), which use five and two seeds; a signed-rank test bottoms out at $p=0.0625$ for $n=5$ and is meaningless for $n=2$, so we report those as means with their spread and make no significance claim. For the learning variants (Sec.~\ref{app:variants}) the same test over ten paired seeds returns $p=0.64$, $p=0.39$ and $p=0.34$ for the no-suppressor, pos-gate and pos-skip variants respectively---that is, none of them differs significantly from the baseline, which is what the text there claims.

\section{Within-Cycle Retention: Definition and Full Values} \label{app:wcr}

\subsubsection{Definition.} A class's training along the stream is grouped into \emph{episodes}: an episode is a maximal run of consecutive blocks in which that class is trained---one block under cyclic, shuffled and permuted, five under \emph{longgap}, twenty under \emph{blocked}. For an episode $e$, let $r_e$ be the class's held-out accuracy immediately after the episode's last block (its just-trained reference), and consider the interval from there until the class's \emph{next} episode begins---or, if it never returns, until the end of the stream. Writing $\mathcal{P}=\{e:\ r_e>\theta\}$ for the episodes entering the average,
\begin{align}
\mathrm{WCR}_{\mathrm{end}} &= \frac{1}{|\mathcal{P}|}\sum_{e\in\mathcal{P}}
  \min\!\Big(\frac{a^{\mathrm{ret}}_e}{r_e},\,1\Big), \\
\mathrm{WCR}_{\mathrm{wst}} &= \frac{1}{|\mathcal{P}|}\sum_{e\in\mathcal{P}}
  \min\!\Big(\frac{a^{\min}_e}{r_e},\,1\Big),
\end{align}
where $a^{\mathrm{ret}}_e$ is the accuracy at the point the class next returns (the last evaluation before its next episode) and $a^{\min}_e$ its lowest accuracy anywhere in the interval; both are averaged additionally over seeds, with $\theta=0.3$ and each ratio capped at $1$. On the cyclic stream an episode is a single block and the next episode is the same class one cycle later, so WCR is the fraction of a class held from just after its own block to just before its next---a full cycle of other classes later. We report WCR over each class's \emph{early episodes} (its first three) and over \emph{all} its episodes.

\subsubsection{Justification of choices.} \emph{The threshold $\theta$.} A class that is still at chance right after its own block has nothing to lose, and its ratio would be $\approx\!1$ for that reason alone. This is not hypothetical: in cycle $0$ the model has ${\sim}20$ \csv{}s and eight of the ten classes sit near chance, so without a threshold the measure is dominated by episodes that cannot move. We therefore require $r>\theta=0.3$. \emph{The cap at $1$.} Early in the stream a class may keep \emph{improving} after its own block, as later classes sharpen the decision among them; such pairs give ratios above $1$ that would offset genuine losses elsewhere in the average. Capping each ratio at $1$ makes the measure record only what was given up. \emph{The worst-case variant.} Comparing only the return point is blind to what happens during the absence: a class can lose half its accuracy in the interval and recover by the time it returns, and $\mathrm{WCR}_{\mathrm{end}}$ would report almost no loss. One class in our own runs does exactly this, falling from $0.80$ to $0.40$ before returning to $0.74$. $\mathrm{WCR}_{\mathrm{wst}}$ is reported alongside for that reason.

\subsubsection{Early episodes versus all.} We report both windows because they answer different questions. A class's \emph{early} episodes are where continual learning is under genuine stress: the class has been seen only a few times, and the methods that depend on re-exposure have not yet had it. As episodes accumulate the stream becomes effectively i.i.d.---every class returns often---and every method improves: over all cyclic episodes the Modeller rises from $0.89/0.84$ (early) to $0.95/0.92$, and the regularisers from ${\approx}0.2$--$0.3$ to ${\approx}0.75$, so the all-episode average narrows the gap by folding the easy later episodes in. The early window is therefore the sterner test of continual learning, and the all-episode figures understate rather than overstate the replay-free difference. Table~\ref{tab:wcr-main} reports both windows for the main methods and Table~\ref{tab:wcr-full} for every method, with standard deviations; Fig.~\ref{fig:wcr} shows the all-episode values by regime.

\begin{table*}[t]
\centering
\caption{Within-cycle retention WCR$_{\mathrm{end}}$/WCR$_{\mathrm{wst}}$ for every method, by regime,
mean $\pm$ standard deviation over ten seeds (definition above; the main-text subset is
Table~\ref{tab:wcr-main}). Bold: best replay-free method in each column; replay methods in italics. $^\star$This work.}
\label{tab:wcr-full}
\scriptsize
\setlength{\tabcolsep}{3pt}
\begin{tabular}{lccccc}
\toprule
Method & cyclic & shuffled & permuted & longgap & blocked \\
\midrule
\multicolumn{6}{l}{\emph{Early episodes (each class's first three)}} \\
Modeller$^\star$ & $\mathbf{.89}\!\pm\!.01$/$\mathbf{.84}\!\pm\!.01$ & $\mathbf{.91}\!\pm\!.01$/$\mathbf{.84}\!\pm\!.01$ & $\mathbf{.88}\!\pm\!.03$/$\mathbf{.81}\!\pm\!.03$ & $\mathbf{.88}\!\pm\!.02$/$\mathbf{.76}\!\pm\!.02$ & $\mathbf{.72}\!\pm\!.06$/$\mathbf{.52}\!\pm\!.06$ \\
SI & $.29\!\pm\!.02$/$.13\!\pm\!.03$ & $.24\!\pm\!.02$/$.16\!\pm\!.02$ & $.28\!\pm\!.04$/$.22\!\pm\!.03$ & $.17\!\pm\!.02$/$.09\!\pm\!.02$ & $.10\!\pm\!.01$/$.10\!\pm\!.01$ \\
MAS & $.18\!\pm\!.02$/$.02\!\pm\!.01$ & $.13\!\pm\!.03$/$.06\!\pm\!.02$ & $.19\!\pm\!.03$/$.14\!\pm\!.04$ & $.14\!\pm\!.04$/$.06\!\pm\!.03$ & $.11\!\pm\!.01$/$.10\!\pm\!.00$ \\
LwF & $.05\!\pm\!.02$/$.00\!\pm\!.00$ & $.07\!\pm\!.03$/$.04\!\pm\!.03$ & $.15\!\pm\!.05$/$.13\!\pm\!.04$ & $.00\!\pm\!.00$/$.00\!\pm\!.00$ & $.10\!\pm\!.00$/$.10\!\pm\!.00$ \\
Plain NN & $.10\!\pm\!.02$/$.00\!\pm\!.00$ & $.06\!\pm\!.03$/$.02\!\pm\!.02$ & $.11\!\pm\!.04$/$.07\!\pm\!.03$ & $.01\!\pm\!.00$/$.00\!\pm\!.00$ & $.10\!\pm\!.00$/$.10\!\pm\!.00$ \\
Expert-AEC & $.63\!\pm\!.17$/$.60\!\pm\!.21$ & $.75\!\pm\!.15$/$.72\!\pm\!.15$ & $.68\!\pm\!.16$/$.67\!\pm\!.16$ & $.60\!\pm\!.13$/$.56\!\pm\!.12$ & $.53\!\pm\!.10$/$.47\!\pm\!.10$ \\
\textit{ER (8)} & $\mathit{.33}\!\pm\!\mathit{.04}$/$\mathit{.12}\!\pm\!\mathit{.03}$ & $\mathit{.30}\!\pm\!\mathit{.05}$/$\mathit{.15}\!\pm\!\mathit{.03}$ & $\mathit{.29}\!\pm\!\mathit{.06}$/$\mathit{.20}\!\pm\!\mathit{.06}$ & $\mathit{.19}\!\pm\!\mathit{.04}$/$\mathit{.08}\!\pm\!\mathit{.03}$ & $\mathit{.22}\!\pm\!\mathit{.05}$/$\mathit{.16}\!\pm\!\mathit{.04}$ \\
\textit{ER-ACE (8)} & $\mathit{.77}\!\pm\!\mathit{.14}$/$\mathit{.67}\!\pm\!\mathit{.16}$ & $\mathit{.71}\!\pm\!\mathit{.16}$/$\mathit{.63}\!\pm\!\mathit{.18}$ & $\mathit{.66}\!\pm\!\mathit{.18}$/$\mathit{.59}\!\pm\!\mathit{.19}$ & $\mathit{.55}\!\pm\!\mathit{.11}$/$\mathit{.44}\!\pm\!\mathit{.10}$ & $\mathit{.48}\!\pm\!\mathit{.14}$/$\mathit{.33}\!\pm\!\mathit{.13}$ \\
\textit{DER++ (8)} & $\mathit{.50}\!\pm\!\mathit{.06}$/$\mathit{.40}\!\pm\!\mathit{.08}$ & $\mathit{.45}\!\pm\!\mathit{.06}$/$\mathit{.39}\!\pm\!\mathit{.06}$ & $\mathit{.51}\!\pm\!\mathit{.08}$/$\mathit{.46}\!\pm\!\mathit{.07}$ & $\mathit{.30}\!\pm\!\mathit{.03}$/$\mathit{.21}\!\pm\!\mathit{.03}$ & $\mathit{.27}\!\pm\!\mathit{.04}$/$\mathit{.23}\!\pm\!\mathit{.03}$ \\
\textit{DER++ (80)} & $\mathit{.88}\!\pm\!\mathit{.01}$/$\mathit{.80}\!\pm\!\mathit{.02}$ & $\mathit{.85}\!\pm\!\mathit{.02}$/$\mathit{.77}\!\pm\!\mathit{.03}$ & $\mathit{.86}\!\pm\!\mathit{.02}$/$\mathit{.80}\!\pm\!\mathit{.02}$ & $\mathit{.82}\!\pm\!\mathit{.02}$/$\mathit{.76}\!\pm\!\mathit{.03}$ & $\mathit{.76}\!\pm\!\mathit{.06}$/$\mathit{.71}\!\pm\!\mathit{.06}$ \\
\midrule
\multicolumn{6}{l}{\emph{All episodes}} \\
Modeller$^\star$ & $\mathbf{.95}\!\pm\!.00$/$\mathbf{.92}\!\pm\!.01$ & $\mathbf{.96}\!\pm\!.01$/$\mathbf{.92}\!\pm\!.01$ & $\mathbf{.95}\!\pm\!.01$/$\mathbf{.92}\!\pm\!.01$ & $\mathbf{.90}\!\pm\!.02$/$\mathbf{.79}\!\pm\!.02$ & $\mathbf{.72}\!\pm\!.06$/$\mathbf{.52}\!\pm\!.06$ \\
SI & $.75\!\pm\!.01$/$.67\!\pm\!.01$ & $.74\!\pm\!.01$/$.68\!\pm\!.01$ & $.72\!\pm\!.01$/$.67\!\pm\!.02$ & $.29\!\pm\!.02$/$.20\!\pm\!.02$ & $.10\!\pm\!.01$/$.10\!\pm\!.01$ \\
MAS & $.75\!\pm\!.01$/$.68\!\pm\!.02$ & $.76\!\pm\!.01$/$.73\!\pm\!.01$ & $.73\!\pm\!.02$/$.70\!\pm\!.02$ & $.24\!\pm\!.05$/$.16\!\pm\!.03$ & $.11\!\pm\!.01$/$.10\!\pm\!.00$ \\
LwF & $.53\!\pm\!.02$/$.45\!\pm\!.03$ & $.59\!\pm\!.01$/$.54\!\pm\!.02$ & $.57\!\pm\!.01$/$.52\!\pm\!.01$ & $.04\!\pm\!.02$/$.04\!\pm\!.01$ & $.10\!\pm\!.00$/$.10\!\pm\!.00$ \\
Plain NN & $.41\!\pm\!.01$/$.17\!\pm\!.01$ & $.48\!\pm\!.02$/$.37\!\pm\!.02$ & $.46\!\pm\!.01$/$.37\!\pm\!.01$ & $.05\!\pm\!.01$/$.04\!\pm\!.01$ & $.10\!\pm\!.00$/$.10\!\pm\!.00$ \\
Expert-AEC & $.92\!\pm\!.02$/$\mathbf{.92}\!\pm\!.02$ & $.92\!\pm\!.01$/$.91\!\pm\!.01$ & $.90\!\pm\!.02$/$.90\!\pm\!.03$ & $.70\!\pm\!.09$/$.67\!\pm\!.09$ & $.53\!\pm\!.10$/$.47\!\pm\!.10$ \\
\textit{ER (8)} & $\mathit{.58}\!\pm\!\mathit{.02}$/$\mathit{.36}\!\pm\!\mathit{.02}$ & $\mathit{.59}\!\pm\!\mathit{.02}$/$\mathit{.47}\!\pm\!\mathit{.02}$ & $\mathit{.58}\!\pm\!\mathit{.02}$/$\mathit{.48}\!\pm\!\mathit{.02}$ & $\mathit{.24}\!\pm\!\mathit{.05}$/$\mathit{.13}\!\pm\!\mathit{.03}$ & $\mathit{.22}\!\pm\!\mathit{.05}$/$\mathit{.16}\!\pm\!\mathit{.04}$ \\
\textit{ER-ACE (8)} & $\mathit{.88}\!\pm\!\mathit{.03}$/$\mathit{.85}\!\pm\!\mathit{.03}$ & $\mathit{.89}\!\pm\!\mathit{.02}$/$\mathit{.85}\!\pm\!\mathit{.03}$ & $\mathit{.86}\!\pm\!\mathit{.06}$/$\mathit{.83}\!\pm\!\mathit{.07}$ & $\mathit{.62}\!\pm\!\mathit{.10}$/$\mathit{.53}\!\pm\!\mathit{.09}$ & $\mathit{.48}\!\pm\!\mathit{.14}$/$\mathit{.33}\!\pm\!\mathit{.13}$ \\
\textit{DER++ (8)} & $\mathit{.72}\!\pm\!\mathit{.03}$/$\mathit{.64}\!\pm\!\mathit{.04}$ & $\mathit{.70}\!\pm\!\mathit{.03}$/$\mathit{.64}\!\pm\!\mathit{.03}$ & $\mathit{.72}\!\pm\!\mathit{.02}$/$\mathit{.67}\!\pm\!\mathit{.02}$ & $\mathit{.37}\!\pm\!\mathit{.03}$/$\mathit{.27}\!\pm\!\mathit{.03}$ & $\mathit{.27}\!\pm\!\mathit{.04}$/$\mathit{.23}\!\pm\!\mathit{.03}$ \\
\textit{DER++ (80)} & $\mathit{.94}\!\pm\!\mathit{.00}$/$\mathit{.92}\!\pm\!\mathit{.01}$ & $\mathit{.94}\!\pm\!\mathit{.01}$/$\mathit{.91}\!\pm\!\mathit{.01}$ & $\mathit{.93}\!\pm\!\mathit{.00}$/$\mathit{.91}\!\pm\!\mathit{.01}$ & $\mathit{.84}\!\pm\!\mathit{.02}$/$\mathit{.79}\!\pm\!\mathit{.02}$ & $\mathit{.76}\!\pm\!\mathit{.06}$/$\mathit{.71}\!\pm\!\mathit{.06}$ \\
\bottomrule
\end{tabular}
\end{table*}

\begin{figure}[t]
\centering
\includegraphics[width=\columnwidth]{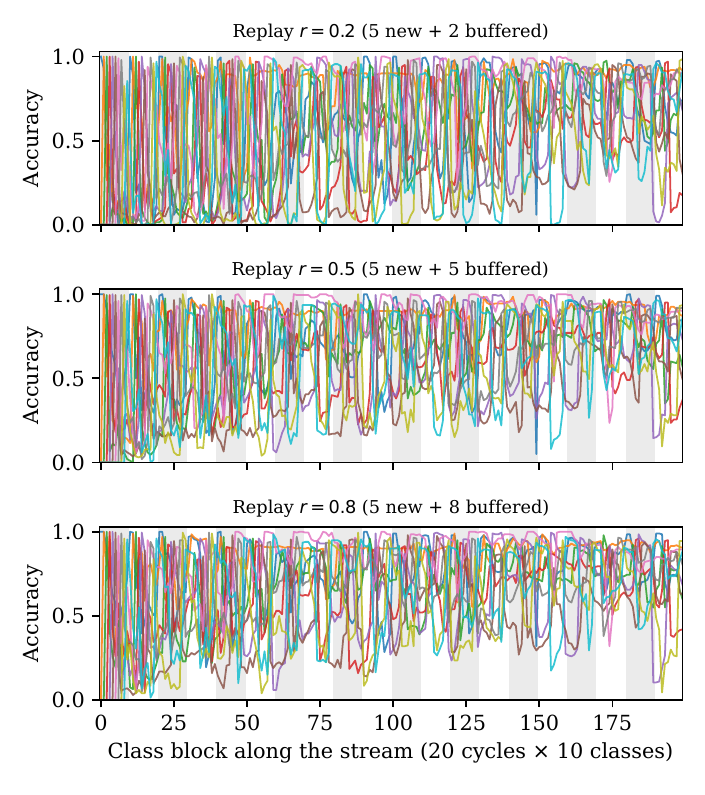}
\caption{Replay \emph{damps} destructive adaptation but does not remove it.
Per-class accuracy along the stream (ten-seed means; axis convention as in
Fig.~\ref{fig:retention}) for the three replay buffers. Compare the plain network
(Fig.~\ref{fig:retention}, bottom, $0$ buffered): each added stored example lifts
the troughs, yet at every setting the per-class curves still collapse and recover
repeatedly, cycle after cycle---the oscillation is attenuated, not eliminated. The
Modeller (Fig.~\ref{fig:retention}, top) shows no such pattern while storing
nothing.}
\label{fig:replayforget}
\end{figure}

\begin{figure}[t]
\centering
\includegraphics[width=0.8\columnwidth]{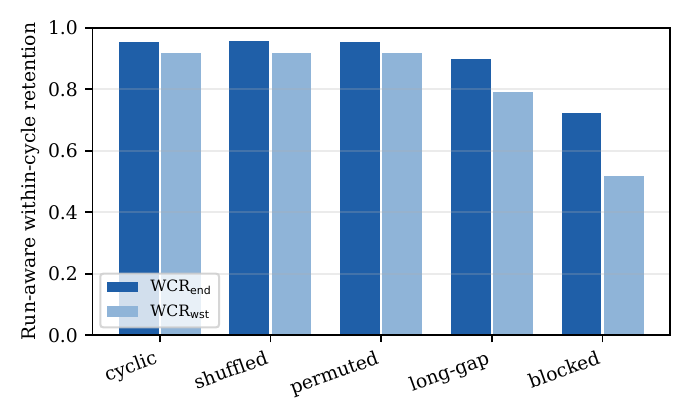}
\caption{Within-cycle retention by regime (all episodes, Modeller, ten-seed means). Retention is
indistinguishable across cyclic, shuffled and permuted, eases under \emph{longgap}'s long
absences, and drops under never-revisited \emph{blocked}.}
\label{fig:wcr}
\end{figure}

\section{Retention Across Stream Regimes} \label{app:regimes} \textbf{Regimes.} All five present the identical data---$200$ class-blocks, twenty per class, five samples per block---and differ only in order. \emph{cyclic}: $0,1,\dots,9$ repeated twenty times (a class recurs every nine blocks). \emph{shuffled}: each ten-block cycle is an independent random permutation of the classes (each class still once per cycle). \emph{permuted}: one global shuffle of all $200$ blocks, no cycle structure. \emph{longgap}: four rounds; within each round every class is trained as a consecutive run of five blocks, in a random order re-drawn per round, so a class is absent for ${\sim}45$ blocks (range $5$--$90$) between runs. \emph{blocked}: each class trained once as twenty consecutive blocks, then never revisited. Held-out evaluation ($20$ images/class) after every block. All regimes use ten seeds, for the Modeller and the neural baselines alike; the Modeller's \emph{cyclic} column in the main tables (Tables~\ref{tab:regimes},~\ref{tab:wcr-main}) is the ten-seed main-experiment configuration evaluated after every block (Sec.~\ref{sec:exp-Modeller}, $0.874$ final / $0.100$ forgetting).

\textbf{Accuracy and forgetting} (Table~\ref{tab:regimes-full}). \emph{forget} is the mean over classes of (peak $-$ final); \emph{trough} the mean lowest accuracy after each class's peak.

\begin{table}[t]
\centering
\caption{Modeller by regime: final accuracy, forgetting (peak$-$final) and trough, mean $\pm$
standard deviation over ten seeds.}
\label{tab:regimes-full}
\small
\begin{tabular}{lccc}
\toprule
Regime & Final & Forget & Trough \\
\midrule
cyclic   & $.874\!\pm\!.020$ & $.100\!\pm\!.021$ & $.594\!\pm\!.041$ \\
shuffled & $.859\!\pm\!.018$ & $.112\!\pm\!.014$ & $.501\!\pm\!.055$ \\
permuted & $.845\!\pm\!.026$ & $.118\!\pm\!.024$ & $.505\!\pm\!.073$ \\
longgap  & $.786\!\pm\!.030$ & $.171\!\pm\!.028$ & $.446\!\pm\!.065$ \\
blocked  & $.661\!\pm\!.043$ & $.296\!\pm\!.049$ & $.414\!\pm\!.072$ \\
\bottomrule
\end{tabular}
\end{table}

\begin{figure}[t]
\centering
\includegraphics[width=0.82\columnwidth]{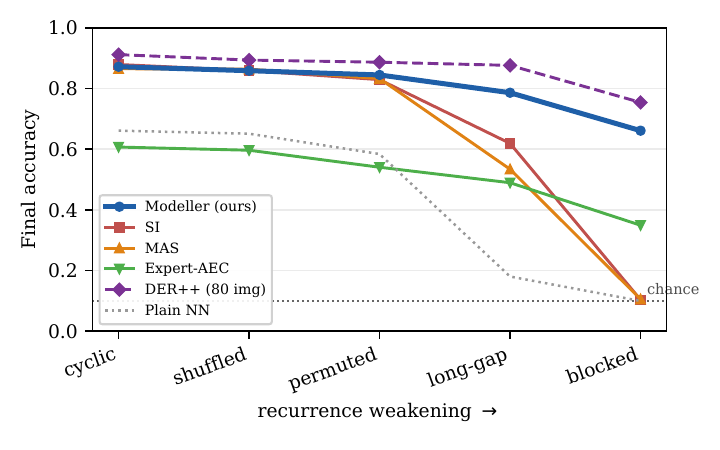}
\caption{Final accuracy as the stream's recurrence is weakened on MNIST (same data reordered). SI and MAS,
level with the Modeller under cyclic recurrence, fall to chance once classes stop recurring,
exposing their apparent continual learning as relearning; the Modeller degrades gracefully. Of the
baselines only the $80$-image replay buffer survives \emph{blocked}; the expert model degrades
heavily.}
\label{fig:regimes}
\end{figure}

\textbf{Baselines by regime} (Table~\ref{tab:baselines-regimes}). Same tuned configurations as Table~\ref{tab:full-baselines}; ten seeds each.

\begin{table*}[t]
\centering
\caption{Final accuracy (top) and forgetting (bottom) by method and regime, mean $\pm$ standard
deviation over ten seeds. The regularisation methods (SI, MAS, LwF) fall to chance under blocked;
only the $80$-image replay buffer (DER++ (80)) retains substantially. Bold: best replay-free method in
each column; replay methods in italics. $^\star$This work.}
\label{tab:baselines-regimes}
\footnotesize
\setlength{\tabcolsep}{4pt}
\begin{tabular}{lccccc}
\toprule
Method & cyclic & shuffled & permuted & longgap & blocked \\
\midrule
\multicolumn{6}{l}{\emph{Final accuracy (higher is better)}} \\
Modeller$^\star$ & $.874\!\pm\!.020$ & $.859\!\pm\!.018$ & $\mathbf{.845}\!\pm\!.026$ & $\mathbf{.786}\!\pm\!.030$ & $\mathbf{.661}\!\pm\!.043$ \\
SI & $\mathbf{.879}\!\pm\!.015$ & $.859\!\pm\!.022$ & $.829\!\pm\!.026$ & $.619\!\pm\!.055$ & $.104\!\pm\!.010$ \\
MAS & $.865\!\pm\!.016$ & $\mathbf{.863}\!\pm\!.040$ & $.832\!\pm\!.024$ & $.534\!\pm\!.095$ & $.106\!\pm\!.008$ \\
LwF & $.810\!\pm\!.027$ & $.807\!\pm\!.034$ & $.722\!\pm\!.043$ & $.172\!\pm\!.065$ & $.100\!\pm\!.000$ \\
Plain NN & $.661\!\pm\!.045$ & $.651\!\pm\!.048$ & $.584\!\pm\!.087$ & $.180\!\pm\!.051$ & $.100\!\pm\!.000$ \\
Expert-AEC & $.607\!\pm\!.058$ & $.597\!\pm\!.063$ & $.540\!\pm\!.056$ & $.489\!\pm\!.089$ & $.350\!\pm\!.065$ \\
\textit{ER (8)} & $\mathit{.727}\!\pm\!\mathit{.052}$ & $\mathit{.688}\!\pm\!\mathit{.070}$ & $\mathit{.646}\!\pm\!\mathit{.086}$ & $\mathit{.374}\!\pm\!\mathit{.088}$ & $\mathit{.217}\!\pm\!\mathit{.051}$ \\
\textit{ER-ACE (8)} & $\mathit{.378}\!\pm\!\mathit{.078}$ & $\mathit{.399}\!\pm\!\mathit{.063}$ & $\mathit{.372}\!\pm\!\mathit{.097}$ & $\mathit{.401}\!\pm\!\mathit{.059}$ & $\mathit{.350}\!\pm\!\mathit{.090}$ \\
\textit{DER++ (8)} & $\mathit{.797}\!\pm\!\mathit{.038}$ & $\mathit{.766}\!\pm\!\mathit{.035}$ & $\mathit{.749}\!\pm\!\mathit{.049}$ & $\mathit{.546}\!\pm\!\mathit{.066}$ & $\mathit{.269}\!\pm\!\mathit{.043}$ \\
\textit{DER++ (80)} & $\mathit{.911}\!\pm\!\mathit{.013}$ & $\mathit{.894}\!\pm\!\mathit{.017}$ & $\mathit{.886}\!\pm\!\mathit{.020}$ & $\mathit{.876}\!\pm\!\mathit{.036}$ & $\mathit{.754}\!\pm\!\mathit{.060}$ \\
\midrule
\multicolumn{6}{l}{\emph{Forgetting, peak$-$final (lower is better)}} \\
Modeller$^\star$ & $\mathbf{.100}\!\pm\!.021$ & $\mathbf{.112}\!\pm\!.014$ & $\mathbf{.118}\!\pm\!.024$ & $\mathbf{.171}\!\pm\!.028$ & $\mathbf{.296}\!\pm\!.049$ \\
SI & $.120\!\pm\!.014$ & $.141\!\pm\!.022$ & $.170\!\pm\!.026$ & $.381\!\pm\!.055$ & $.896\!\pm\!.010$ \\
MAS & $.135\!\pm\!.016$ & $.137\!\pm\!.040$ & $.167\!\pm\!.024$ & $.466\!\pm\!.095$ & $.894\!\pm\!.008$ \\
LwF & $.191\!\pm\!.027$ & $.193\!\pm\!.034$ & $.278\!\pm\!.043$ & $.828\!\pm\!.065$ & $.900\!\pm\!.000$ \\
Plain NN & $.339\!\pm\!.045$ & $.349\!\pm\!.048$ & $.416\!\pm\!.087$ & $.820\!\pm\!.051$ & $.900\!\pm\!.000$ \\
Expert-AEC & $.245\!\pm\!.080$ & $.232\!\pm\!.041$ & $.303\!\pm\!.072$ & $.321\!\pm\!.053$ & $.426\!\pm\!.110$ \\
\textit{ER (8)} & $\mathit{.273}\!\pm\!\mathit{.052}$ & $\mathit{.311}\!\pm\!\mathit{.070}$ & $\mathit{.354}\!\pm\!\mathit{.086}$ & $\mathit{.625}\!\pm\!\mathit{.088}$ & $\mathit{.781}\!\pm\!\mathit{.052}$ \\
\textit{ER-ACE (8)} & $\mathit{.525}\!\pm\!\mathit{.092}$ & $\mathit{.561}\!\pm\!\mathit{.059}$ & $\mathit{.511}\!\pm\!\mathit{.118}$ & $\mathit{.458}\!\pm\!\mathit{.052}$ & $\mathit{.417}\!\pm\!\mathit{.099}$ \\
\textit{DER++ (8)} & $\mathit{.196}\!\pm\!\mathit{.036}$ & $\mathit{.227}\!\pm\!\mathit{.037}$ & $\mathit{.246}\!\pm\!\mathit{.049}$ & $\mathit{.454}\!\pm\!\mathit{.066}$ & $\mathit{.730}\!\pm\!\mathit{.042}$ \\
\textit{DER++ (80)} & $\mathit{.071}\!\pm\!\mathit{.013}$ & $\mathit{.090}\!\pm\!\mathit{.016}$ & $\mathit{.100}\!\pm\!\mathit{.019}$ & $\mathit{.112}\!\pm\!\mathit{.035}$ & $\mathit{.235}\!\pm\!\mathit{.062}$ \\
\bottomrule
\end{tabular}
\end{table*}

\textbf{Model size by regime} (Table~\ref{tab:size-regimes}). \csv{} count at the end of selected cycles (block $(c{+}1)\cdot10-1$), mean over seeds, with peak and the final/peak ratio.

\begin{table}[t]
\centering
\caption{\csv{}-count evolution by regime (end of the given cycle), peak, and final/peak. The
equilibrium property (final/peak near $1$) holds everywhere; the level is demand-driven, from
${\sim}41$ (blocked) to ${\sim}190$ (shuffled).}
\label{tab:size-regimes}
\footnotesize
\setlength{\tabcolsep}{3pt}
\begin{tabular}{lccccccc}
\toprule
Regime & cyc0 & cyc2 & cyc5 & cyc10 & cyc15 & cyc19 & peak (f/pk) \\
\midrule
cyclic   & $21$ & $68$ & $134$ & $168$ & $177$ & $187$ & $197$ ($.95$) \\
shuffled & $22$ & $76$ & $142$ & $180$ & $198$ & $192$ & $210$ ($.91$) \\
permuted & $18$ & $63$ & $118$ & $162$ & $183$ & $187$ & $207$ ($.90$) \\
longgap  & $5$  & $23$ & $58$  & $115$ & $150$ & $176$ & $193$ ($.91$) \\
blocked  & $2$  & $5$  & $10$  & $24$  & $34$  & $41$  & $44$ ($.91$) \\
\bottomrule
\end{tabular}
\end{table}

\begin{figure}[tbp]
\centering
\includegraphics[width=0.70\columnwidth]{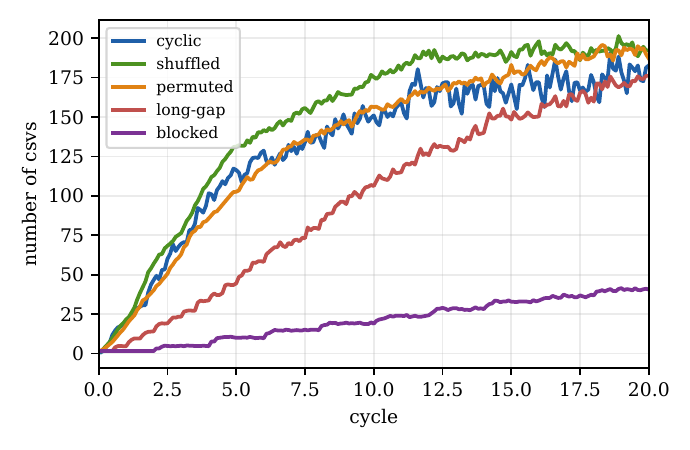}
\caption{Model size (\csv{} count) over the stream, by regime (ten-seed means). Every regime settles
into equilibrium, but the level is demand-driven---${\sim}190$ where interleaving forces
discrimination, ${\sim}41$ under \emph{blocked} where it never does; \emph{longgap} grows in
delayed steps as its rounds introduce classes.}
\label{fig:sizeregimes}
\end{figure}

\section{Transformation Sensitivity: Full Grid} \label{app:transform} The model is trained on the standard centred protocol and left unchanged; the held-out set is transformed at evaluation only. Ten seeds; baseline (identity) accuracy $0.874$. Translations are in pixels along the column ($x$) and row ($y$) axes; scale is about the image centre; rotation in degrees. Values are mean accuracy (SEM $\le 0.01$ throughout); see Table~\ref{tab:transform}.

\begin{table}[t]
\centering
\caption{Test-time transformation sensitivity, mean accuracy over ten seeds (baseline $0.874$).}
\label{tab:transform}
\small
\setlength{\tabcolsep}{5pt}
\resizebox{\columnwidth}{!}{%
\begin{tabular}{lcccccccc}
\toprule
$x$-shift (px)  & $-4$ & $-3$ & $-2$ & $-1$ & $1$ & $2$ & $3$ & $4$ \\
accuracy        & $.707$ & $.807$ & $.839$ & $.872$ & $.860$ & $.860$ & $.815$ & $.736$ \\
\midrule
$y$-shift (px)  & $-4$ & $-3$ & $-2$ & $-1$ & $1$ & $2$ & $3$ & $4$ \\
accuracy        & $.553$ & $.698$ & $.798$ & $.850$ & $.855$ & $.776$ & $.630$ & $.429$ \\
\midrule
scale           & \multicolumn{2}{c}{$0.8$} & \multicolumn{2}{c}{$0.9$} & \multicolumn{2}{c}{$1.1$} & \multicolumn{2}{c}{$1.2$} \\
accuracy        & \multicolumn{2}{c}{$.827$} & \multicolumn{2}{c}{$.863$} & \multicolumn{2}{c}{$.843$} & \multicolumn{2}{c}{$.843$} \\
\midrule
rotation (deg)  & $-20$ & $-15$ & $-10$ & $-5$ & $5$ & $10$ & $15$ & $20$ \\
accuracy        & $.754$ & $.777$ & $.802$ & $.861$ & $.853$ & $.858$ & $.802$ & $.755$ \\
\bottomrule
\end{tabular}}
\end{table}

\section{Preservation Audit} \label{app:preserve} After every block we re-check a fixed probe set of already-seen observations (ten per class). For each, we record whether a \csv{} that \emph{fully matched} it before still fully matches it (a \emph{satisfaction} check---the direct test of the preservation property) and whether its predicted label is unchanged; every satisfaction lapse is attributed to a cause by comparing the \csv{}'s state between the two checks. Table~\ref{tab:preserve} gives the totals over the run; lapses run at ${\sim}2.7\%$ of checks (so ${\sim}97\%$ exact preservation), and label predictions are ${\sim}97\%$ stable block-to-block.

\begin{table}[tb]
\centering
\caption{Preservation audit over the full run, per seed. ``Satisfaction lapses'' are
previously-matching (\csv{}, observation) pairs that no longer fully match; ``label changes''
are predicted-label flips on already-seen observations between consecutive blocks.}
\label{tab:preserve}
\small
\setlength{\tabcolsep}{5pt}
\resizebox{\columnwidth}{!}{%
\begin{tabular}{lccccc}
\toprule
Seed & pairs checked & satisfaction lapses & lapse rate & label changes & reintegrations \\
\midrule
$0$ & $315{,}580$ & $8{,}240$ & $2.6\%$ & $505$ & $382$ \\
$1$ & $346{,}074$ & $9{,}755$ & $2.8\%$ & $580$ & $340$ \\
$2$ & $322{,}366$ & $8{,}440$ & $2.6\%$ & $621$ & $383$ \\
\bottomrule
\end{tabular}}
\end{table}

Attributing the lapses (all seeds pooled): $14{,}811$ to online spatial re-estimation (the running node-position estimates drift, and matching is spatial), $5{,}865$ to the \csv{}'s own source changing, $5{,}755$ to a downstream unit re-pinning in the chain, and $4$ unexplained. So roughly $56\%$ are spatial and $44\%$ structural---the structural share consistent with reintegration composing requirements back in, as discussed in Sec.~\ref{sec:exp-preserve}. None of these mechanisms is the removal-only refinement the preservation theorem governs; they are the additional mechanisms of the full instantiation, and their aggregate cost is the ${\sim}3\%$ departure from exact preservation.

\section{Standard Continual-Learning Baselines} \label{app:clbaselines}

This section documents the regularisation and replay baselines of Table~\ref{tab:full-baselines}: what was implemented, how each was tuned, and the sweeps behind the single configuration reported for each, which was selected by final held-out accuracy. All run on the identical stream, backbone and gradient budget as the other baselines, and were implemented directly rather than taken from a library, so their specification is given here in full.

\paragraph{Methods.} \textbf{SI} \citep{zenke2017si} accumulates a per-parameter path integral $\omega_k \mathrel{+}= -g_k\,\Delta\theta_k$ during training and, at each consolidation, folds $\mathrm{relu}\!\left(\omega_k/(\Delta_k^2+\xi)\right)$ into a cumulative importance $\Omega_k$, penalising $\sum_k \Omega_k(\theta_k-\tilde\theta_k)^2$; $g_k$ is the gradient of the \emph{task} loss, and $\xi{=}0.1$. \textbf{MAS} \citep{aljundi2018mas} instead accumulates $\left|\partial\lVert f(x)\rVert^2/\partial\theta_k\right|$ over the block's samples, computed on the logits and requiring no labels, with the same quadratic penalty. \textbf{LwF} \citep{li2018lwf} keeps a snapshot of the model and adds a distillation term between snapshot and current outputs on the \emph{current} block's inputs, at temperature $T{=}2$ with the standard $T^2$ scaling. \textbf{DER++} \citep{buzzega2020derpp} stores $(x,y)$ together with the logits emitted when the example was inserted, drawing two independent buffer batches for $\mathcal{L}=\mathrm{CE}(\text{new}) + \alpha\lVert f(x_1)-z_1\rVert^2 + \beta\,\mathrm{CE}(x_2,y_2)$. \textbf{ER-ACE} \citep{caccia2022erace} restricts the incoming batch's cross-entropy to the classes present in it plus those not yet seen, leaving the buffer term over the full label space. Buffers are filled by reservoir sampling.

\paragraph{Penalty strength dominates, and the usual values are inert.} The regularisation methods are far more sensitive to their penalty coefficient $C$ than to anything else. With Adam at $10^{-3}$ the parameters move little within a block, so $(\theta-\tilde\theta)^2$ is of order $10^{-6}$ and at $C\!\sim\!1$---the scale commonly quoted for these methods on larger models---the penalty gradient is under $1\%$ of the task gradient and the regulariser is numerically switched off: SI, MAS and the plain network then return \emph{identical} results. Sweeping $C$ over five orders of magnitude, each method has a clear and quite different optimum (Table~\ref{tab:csweep} brackets each). Reporting a regularisation baseline without this sweep measures nothing. The values tried were $C\in\{0.1,1,10,100,300,10^3,3\!\times\!10^3,10^4\}$ for SI, $\{10^{-3},10^{-2},10^{-1},1,10,10^2\}$ for MAS and $\{0.1,0.3,1,3,10,30\}$ for LwF; six $(\alpha,\beta)$ pairs in $[0.1,1]$ for DER++; the consolidation cadence over $\{1,2,3,7,10,13,20,200\}$ gradient steps; and the buffer over $\{2,5,8,20,50,80\}$ examples.

\begin{table}[h]
\centering
\caption{Penalty-strength sweep, bracketing each method's optimum (ten seeds, consolidation
per block). WCR is WCR$_{\mathrm{end}}$ over each class's first three episodes (the early window of Table~\ref{tab:wcr-main}); Early is the mean end-of-cycle accuracy over cycles $0$--$2$.}
\label{tab:csweep}
\small
\begin{tabular}{llccc}
\toprule
Method & $C$ & Final & Early & WCR \\
\midrule
SI & $300$ & $0.806\!\pm\!0.039$ & $0.469$ & $0.30$ \\
SI & $1000$ & $0.803\!\pm\!0.063$ & $0.474$ & $0.34$ \\
SI & $3000$ & $0.807\!\pm\!0.048$ & $0.468$ & $0.37$ \\
MAS & $0.003$ & $0.856\!\pm\!0.018$ & $0.174$ & $0.11$ \\
MAS & $0.01$ & $0.836\!\pm\!0.020$ & $0.185$ & $0.13$ \\
MAS & $0.03$ & $0.867\!\pm\!0.012$ & $0.233$ & $0.18$ \\
LwF & $1.0$ & $0.712\!\pm\!0.041$ & $0.136$ & $0.02$ \\
LwF & $3.0$ & $0.730\!\pm\!0.038$ & $0.178$ & $0.03$ \\
LwF & $10$ & $0.695\!\pm\!0.061$ & $0.228$ & $0.03$ \\
\bottomrule
\end{tabular}
\end{table}

\paragraph{The consolidation schedule, and asynchrony.} Each regularisation method must be told \emph{when} to consolidate---when to declare the current parameters the ones to protect. We varied that cadence from every gradient step to once per cycle ($20$ steps is one class block, $200$ one cycle). Cadences that divide $20$ are \emph{aligned}: a consolidation interval never spans a class change. Cadences of $3$, $7$ and $13$ do not divide $20$, so consolidation drifts through the block and straddles class changes constantly.

Two extremes fail, for different reasons. At every step the anchor $\tilde\theta$ is reset to the current parameters, so at the next step $\theta=\tilde\theta$ and both the penalty and its gradient are \emph{identically} zero: all three methods reduce exactly to the plain network, and SI, MAS and plain agree to three decimals ($0.663$). At once per cycle the anchor goes stale across ten class blocks and performance collapses. Between those extremes the choice matters little (Table~\ref{tab:cadence})---and, contrary to what the mechanisms suggest, \emph{alignment does not matter for accuracy}: the asynchronous cadences are as accurate as the aligned ones, or more, for every method. Retention is low at every cadence; SI retains most at an aligned cadence ($10$ steps), MAS is flat, and LwF retains slightly more when asynchronous. Our system has no cadence to choose.

\begin{table}[h]
\centering
\caption{Consolidation cadence, in gradient steps, at each method's best $C$ on the initial grid (ten
seeds); for MAS this is $C{=}0.01$, while bracketing later found $C{=}0.03$ slightly better
(Table~\ref{tab:csweep}), which the main tables use. Each pair is Final and WCR$_{\mathrm{end}}$,
the latter over each class's first three episodes (the early window of Table~\ref{tab:wcr-main}). \textsc{async} cadences do not divide the $20$-step
block, so they straddle class changes.}
\label{tab:cadence}
\small
\setlength{\tabcolsep}{3pt}
\begin{tabular}{llcccccc}
\toprule
& & \multicolumn{2}{c}{SI} & \multicolumn{2}{c}{MAS} & \multicolumn{2}{c}{LwF} \\
\cmidrule(lr){3-4}\cmidrule(lr){5-6}\cmidrule(lr){7-8}
Cadence & Phase & Final & WCR & Final & WCR & Final & WCR \\
\midrule
$1$ & aligned & $0.663$ & $0.10$ & $0.663$ & $0.10$ & $0.644$ & $0.11$ \\
$2$ & aligned & $0.850$ & $0.22$ & $0.818$ & $0.10$ & $0.675$ & $0.05$ \\
$3$ & \textsc{async} & $0.864$ & $0.25$ & $0.830$ & $0.12$ & $0.726$ & $0.04$ \\
$7$ & \textsc{async} & $0.879$ & $0.29$ & $0.830$ & $0.12$ & $0.798$ & $0.05$ \\
$10$ & aligned & $0.836$ & $0.43$ & $0.840$ & $0.12$ & $0.742$ & $0.04$ \\
$13$ & \textsc{async} & $0.836$ & $0.32$ & $0.855$ & $0.12$ & $0.725$ & $0.08$ \\
$20$ (block) & aligned & $0.803$ & $0.34$ & $0.836$ & $0.13$ & $0.730$ & $0.03$ \\
$200$ (cycle) & aligned & $0.329$ & $0.11$ & $0.624$ & $0.13$ & $0.569$ & $0.08$ \\
\bottomrule
\end{tabular}
\end{table}

\paragraph{One sample at a time.} Our system integrates one sample at a time; the baselines see each block as a batch of five, revisited. Because that difference is a property of the protocol rather than of the methods, we reran the replay-free baselines in our regime---batch of one, four gradient steps per sample, so that the total gradient budget per block is unchanged and only the \emph{grouping} of the data differs. Every one of them loses accuracy, and their retention stays far below ours (Table~\ref{tab:sbs}). We restrict this comparison to replay-free methods: for a replay method the constraint does not bind, since the buffer supplies a multi-sample batch at every step regardless. The parity is therefore on presentation order, not on work done per observation, for which our system has no per-sample analogue.

\begin{table}[h]
\centering
\caption{Batch of five versus one sample at a time, at an identical total gradient budget (ten
seeds). Replay-free methods only; MAS at $C{=}0.01$, its optimum on the initial grid (the main tables use
the slightly better $C{=}0.03$; Table~\ref{tab:csweep}). WCR is WCR$_{\mathrm{end}}$ over each class's first
three episodes (the early window of Table~\ref{tab:wcr-main}); Early is the mean end-of-cycle accuracy over
cycles $0$--$2$. Bold: best in each column (all methods are replay-free). $^\star$This work.}
\label{tab:sbs}
\small
\setlength{\tabcolsep}{3pt}
\begin{tabular}{lcccccc}
\toprule
& \multicolumn{3}{c}{Batch of five} & \multicolumn{3}{c}{One at a time} \\
\cmidrule(lr){2-4}\cmidrule(lr){5-7}
Method & Final & Early & WCR & Final & Early & WCR \\
\midrule
Plain & $0.663$ & $0.158$ & $0.10$ & $0.577$ & $0.155$ & $0.08$ \\
SI & $0.803$ & $\mathbf{0.474}$ & $\mathbf{0.34}$ & $0.776$ & $0.406$ & $0.26$ \\
MAS & $\mathbf{0.836}$ & $0.185$ & $0.13$ & $0.659$ & $0.150$ & $0.09$ \\
LwF & $0.730$ & $0.178$ & $0.03$ & $0.622$ & $0.206$ & $0.03$ \\
\midrule
Modeller$^\star$ & \multicolumn{3}{c}{---} & $\mathbf{.874}$ & $\mathbf{.546}$ & $\mathbf{.89}$ \\
\bottomrule
\end{tabular}
\end{table}

\begin{figure}[h]
\centering
\includegraphics[width=\columnwidth]{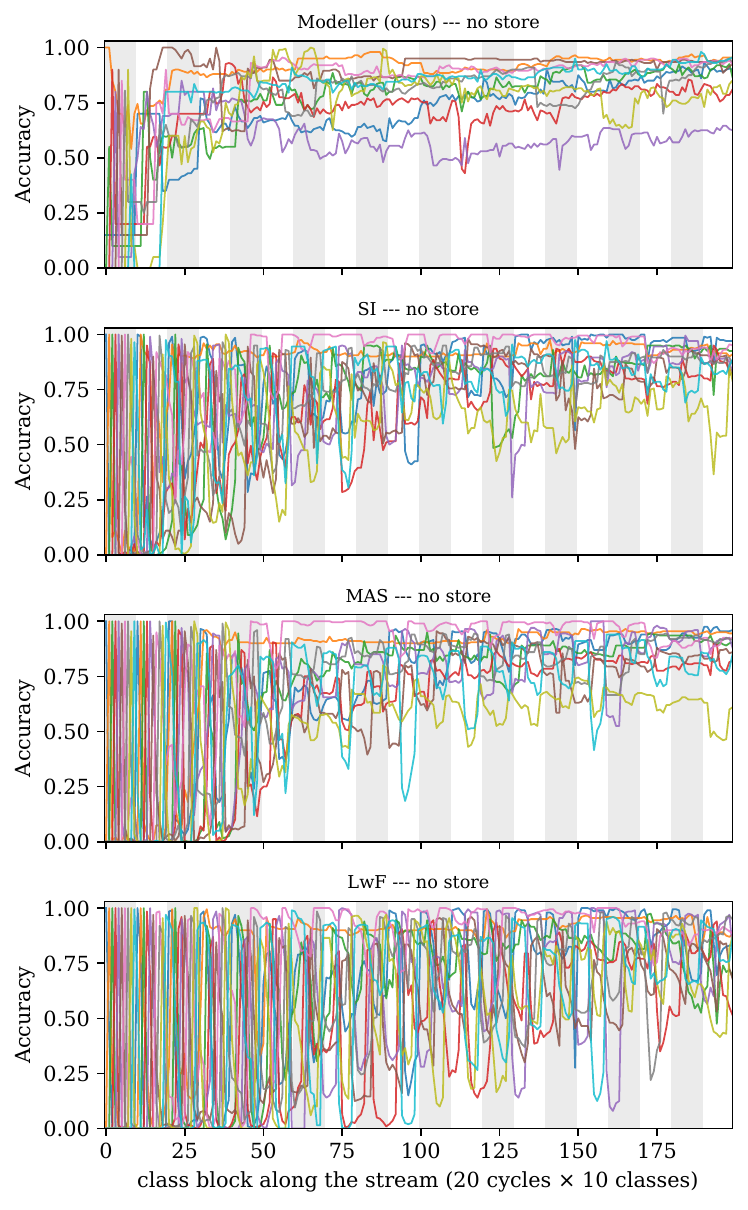}
\caption{Equal final accuracy, opposite behaviour. Per-class accuracy along the stream
(ten-seed means; axis convention as in Fig.~\ref{fig:retention}) for the methods that store no
past data. SI and MAS end the run at accuracies indistinguishable from ours
(Table~\ref{tab:regimes}), but their per-class curves collapse and recover repeatedly---each
class is given up and relearned every cycle---whereas ours are held. This is also why final
accuracy converges: over twenty cycles of re-exposure the procedure becomes stochastic gradient
descent with a slow time constant, under which any method that learns at all approaches the
same ceiling. Retention is what still distinguishes them at that horizon.}
\label{fig:clforgetnostore}
\end{figure}

\begin{figure}[h]
\centering
\includegraphics[width=\columnwidth]{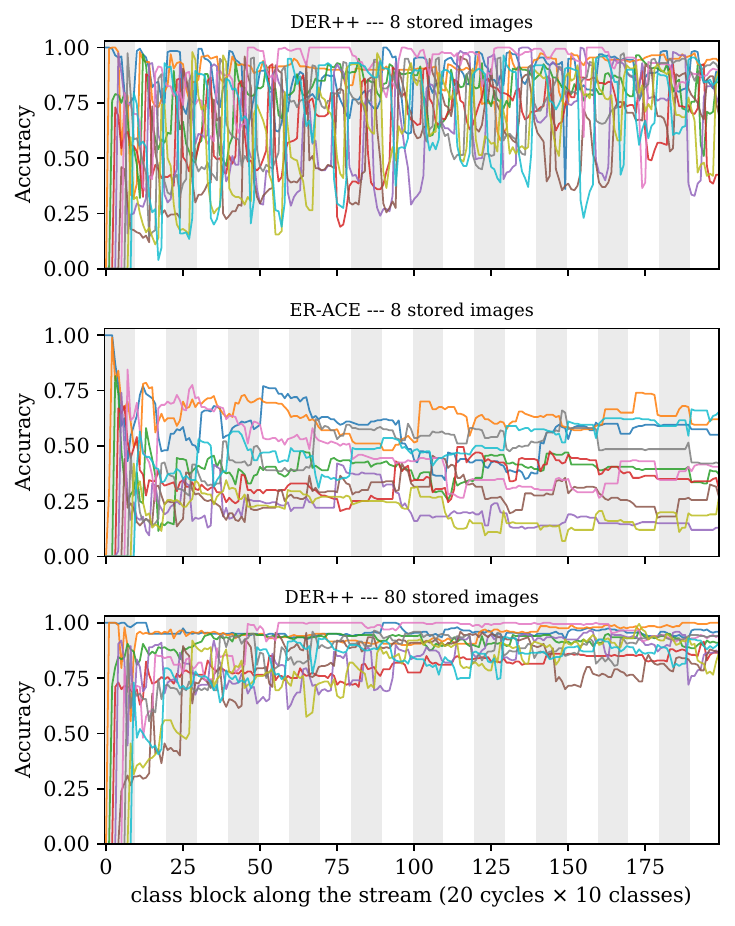}
\caption{Modern replay at the two buffer capacities. At $8$ stored images DER++ still
oscillates. At $80$ the oscillation is largely gone and it exceeds us in accuracy, with
retention comparable to ours---but at that capacity each update is dominated by stored data rather than by the
arriving sample, so the procedure is close to ordinary training on a growing pooled batch and
is no longer a continual-learning result in the sense at issue here. ER-ACE holds what it has
at either capacity but learns little at the small one, for the reason given in the text.}
\label{fig:clforgetreplay}
\end{figure}

\paragraph{Replay methods and the size of the store.} DER++ was swept over $\alpha$ and $\beta$ at both buffer capacities, receiving the same tuning as the regularisation methods. At the capacity the main comparison holds fixed ($8$ images) it reaches $0.810$; at $80$ images---$8\%$ of the entire stream, with the incoming stream unchanged---it reaches $0.912$, with retention comparable to ours (Table~\ref{tab:wcr-main}). ER-ACE behaves differently: its retention is high at every capacity, because its whole mechanism is to prevent arriving data from suppressing established classes, but its accuracy is low at small buffers. The reason is specific to a cyclic stream: once every class has been seen, the incoming batch of a single-class block admits only its own label, its cross-entropy is exactly zero, and the method learns from its buffer alone. Its numbers here therefore reflect a protocol mismatch rather than a defect of the method.

\begin{table}[h]
\centering
\caption{Replay methods at both buffer capacities (ten seeds). Each pair is Final and
WCR$_{\mathrm{end}}$, the latter over each class's first three episodes (the early window of Table~\ref{tab:wcr-main}).}
\label{tab:replaymodern}
\small
\setlength{\tabcolsep}{3pt}
\begin{tabular}{llcccc}
\toprule
\multicolumn{2}{l}{DER++} & \multicolumn{2}{c}{$8$ images} & \multicolumn{2}{c}{$80$ images} \\
\cmidrule(lr){3-4}\cmidrule(lr){5-6}
$\alpha$ & $\beta$ & Final & WCR & Final & WCR \\
\midrule
$0.1$ & $0.5$ & $0.799$ & $0.41$ & $0.883$ & $0.78$ \\
$0.5$ & $0.5$ & $0.779$ & $0.46$ & $0.905$ & $0.87$ \\
$1.0$ & $0.5$ & $0.793$ & $0.50$ & $0.910$ & $0.91$ \\
$0.5$ & $1.0$ & $0.790$ & $0.47$ & $0.912$ & $0.88$ \\
$1.0$ & $1.0$ & $0.810$ & $0.50$ & $0.906$ & $0.91$ \\
$0.2$ & $1.0$ & $0.790$ & $0.45$ & $0.898$ & $0.84$ \\
\midrule
\multicolumn{2}{l}{ER-ACE} & Final & WCR & Final & WCR \\
\midrule
\multicolumn{2}{l}{$r{=}0.2$} & $0.150$ & $0.61$ & $0.631$ & $0.87$ \\
\multicolumn{2}{l}{$r{=}0.5$} & $0.337$ & $0.69$ & $0.772$ & $0.95$ \\
\multicolumn{2}{l}{$r{=}0.8$} & $0.399$ & $0.77$ & $0.825$ & $0.97$ \\
\bottomrule
\end{tabular}
\end{table}

\section{Enlarging the Replay Buffer} \label{app:largebuffer}

Two questions about the baselines of Sec.~\ref{sec:exp-regimes} are worth settling here. The first is whether they are adequately configured at all: the same fully-connected network trained in the ordinary i.i.d.\ manner reaches $0.979$ test accuracy on full MNIST, and $0.90$ when restricted to the stream's total budget of $100$ samples per class presented i.i.d.\ (the corresponding $26$-class EMNIST ceiling is $0.763$, Sec.~\ref{app:emnist}). Their low continual numbers therefore reflect the class-incremental stream, not the architecture or the optimizer.

The second is how far \emph{naive experience replay} (plain reservoir sampling, the simplest replay baseline) improves when its buffer is no longer kept comparable in scale to the five new samples per block. With its buffer enlarged by an order of magnitude---$20$, $50$, $80$ stored examples at $r{=}0.2,0.5,0.8$, against a per-class training budget of only $100$ samples, so the buffer approaches a class's entire data---final accuracy rises to $0.807$, $0.833$, and $0.882$ respectively (three seeds), up from $0.717$, $0.750$, $0.760$; the largest setting slightly exceeds the Modeller ($0.874$). The crossover therefore lies between $50$ and $80$ retained examples---between roughly $5\%$ and $8\%$ of the entire stream. We lead at every capacity below it, so the comparison is not decided at the $8$-example setting the main text holds fixed. Relaxing the stream further, so that each block supplies a class's full data rather than five samples, raises batch-composition replay to ${\sim}0.93$.

The same holds for the tuned modern methods: DER++ with an $80$-image buffer (Table~\ref{tab:replaymodern}) is the only setting in which any baseline is more accurate than the Modeller while matching its retention, and two things locate it. The first is its composition: $80$ retained examples replayed alongside five new ones is sixteen stored for every one arriving, so roughly $94\%$ of each update is data the learner has already seen---much closer to ordinary training on a growing pooled set than to learning from a stream. The second is its size relative to the stream, which is the only way to compare buffers across datasets of different scale. Our stream is $1{,}000$ samples in total, so $80$ retained is $8\%$ of everything the learner ever sees; the buffers used for Seq-CIFAR-10 by \citet{buzzega2020derpp}---$200$, $500$ and $5120$ against $50{,}000$ training images---are $0.4\%$, $1\%$ and $10\%$ of theirs. Measured that way our largest setting sits near the top of the range the field itself uses, not below it. We report it because it is the honest upper end of what replay buys, not because it is a comparable setting.

\begin{figure}[t]
\centering
\includegraphics[width=\columnwidth]{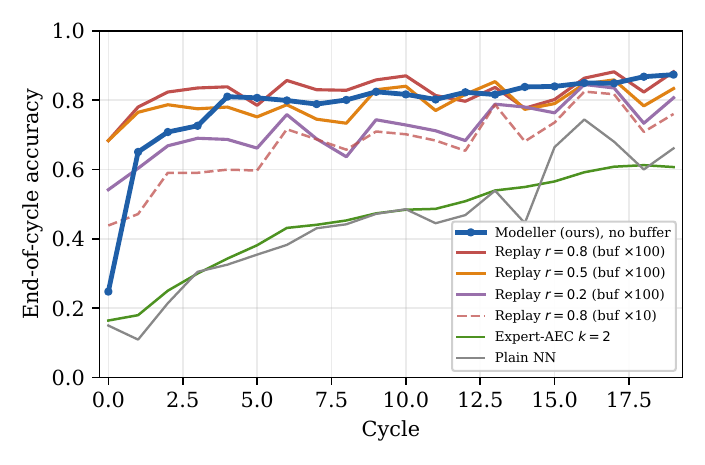}
\caption{End-of-cycle accuracy with enlarged \emph{naive-replay} buffers ($\times$100 scale),
against the Modeller, the original $\times$10 setting (dashed), the expert model,
and the plain network. Larger buffers lift replay to the Modeller's band, but only
late; the Modeller leads the early cycles while storing no past data.}
\label{fig:bigbufspectrum}
\end{figure}

\begin{figure}[t]
\centering
\includegraphics[width=\columnwidth]{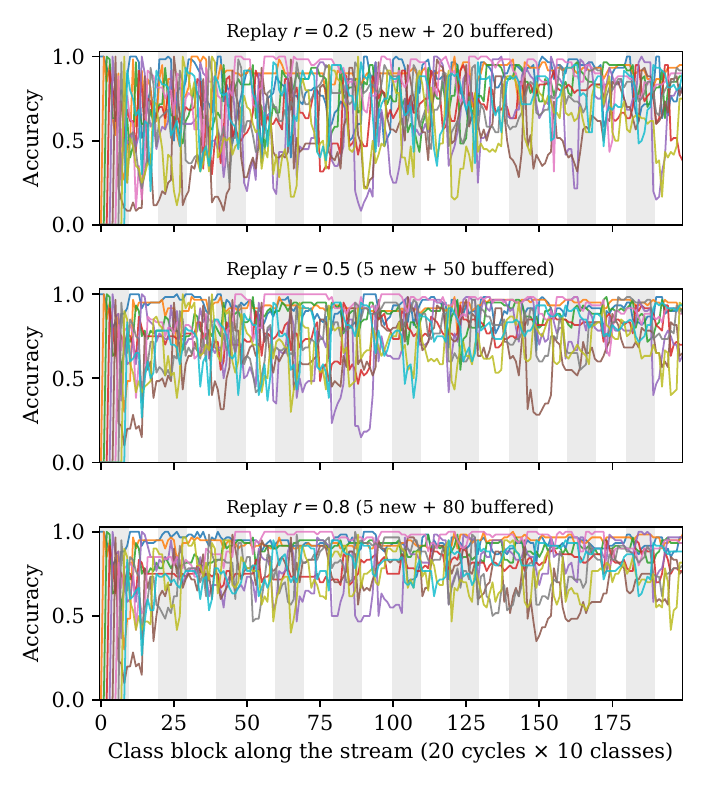}
\caption{Per-class accuracy along the stream with enlarged \emph{naive-replay} buffers (three-seed
means; axis convention as in Fig.~\ref{fig:retention}). With $20$--$80$ stored examples
the oscillation of the small-buffer settings is largely suppressed---destructive
adaptation is bought off with stored data, with up to $80$ stored examples, as many as $80\%$ of a single class's
training budget.}
\label{fig:x100forget}
\end{figure}

The naive-replay spectrum shows the same shape. Figure~\ref{fig:bigbufspectrum} shows the trajectories: the Modeller still leads through the early cycles, and the large-buffer runs overtake it only late, once enough of the past has been stored and re-presented. Figure~\ref{fig:x100forget} shows why the accuracy improves---with $20$--$80$ examples in the buffer the per-class oscillation of Fig.~\ref{fig:replayforget} is strongly suppressed, since most of each class is re-presented at every step. Retention here is therefore purchased directly with stored data: at $r{=}0.8$ the buffer holds $80$ examples---as many as $80\%$ of a single class's training budget---which is closer to interleaved i.i.d.\ training than to a boundary-free stream. For reference, the same network trained i.i.d.\ on this data budget reaches ${\sim}0.90$, and relaxing the stream so each block supplies a class's full data raises batch-replay to ${\sim}0.93$. The Modeller attains $0.874$ while storing no past data and re-presenting nothing.

\section{Expert-Routing Baseline: All Configurations} \label{app:aecforget}

\begin{figure}[t]
\centering
\includegraphics[width=\columnwidth]{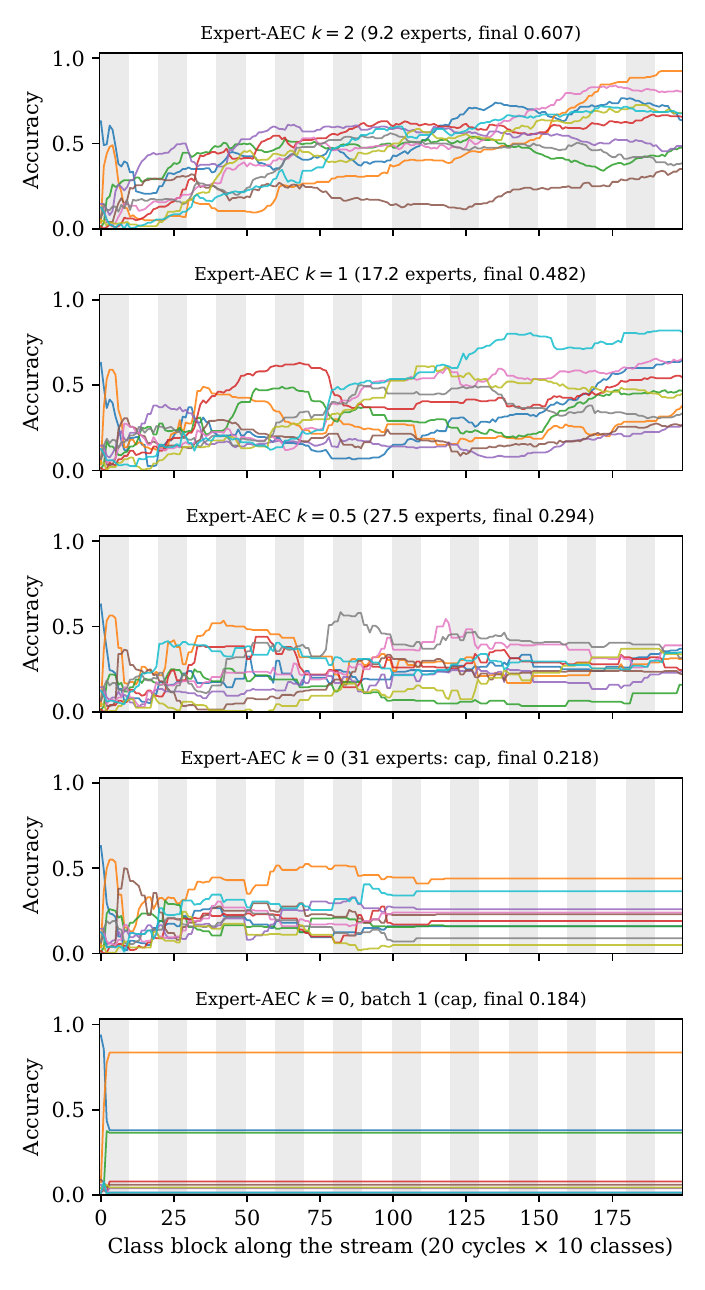}
\caption{Per-class accuracy along the stream for every expert-routing
configuration (ten-seed means; axis convention as in Fig.~\ref{fig:retention}). Curves
are smooth rather than oscillating---an expert is reached almost only by its own
class---but
plateau far below, capped by test-time routing; the two capped settings (bottom)
stop improving entirely.}
\label{fig:aecforget}
\end{figure}

\begin{figure}[t]
\centering
\includegraphics[width=\columnwidth]{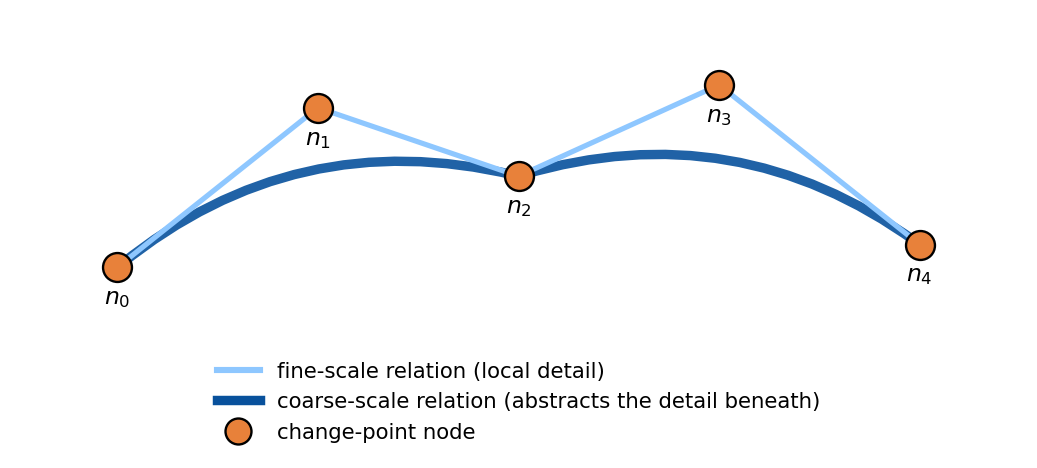}
\caption{The core idea in miniature. Short-range relations link neighbouring change
points (thin); long-range relations span several of them (thick). Both coexist over
the same nodes in one network, so relations of different extent are simultaneously
present and interconnected, with no ordering or priority between them.}
\label{fig:schematic}
\end{figure}

The expansion baseline spawns a new expert whenever the reconstruction error of every existing expert exceeds a $k$-scaled threshold, so $k$ controls how readily capacity is added; prediction routes each test sample to a single expert. Figure~\ref{fig:aecforget} shows the per-class trajectories for every configuration we ran.

The failure mode is qualitatively different from replay's. Because an expert is reached almost exclusively by observations of one class, the curves show none of the replay oscillation: they are smooth and drift slowly upward, consistent with its high within-cycle retention (WCR$_{\mathrm{end}}$ ${\approx}0.92$ over all episodes, Table~\ref{tab:wcr-main}). What limits the method is \emph{routing}: many classes sit at $0.1$--$0.5$ for the whole run despite being retained somewhere in the model, because test-time expert selection sends their samples to the wrong expert. Lowering $k$ spawns more experts, makes routing harder, and lowers the ceiling monotonically ($0.607\rightarrow0.482\rightarrow 0.294\rightarrow0.218$). In the two configurations that reach the $30$-expert cap the curves flatten into horizontal lines---capacity is exhausted and the system stops improving at all, most starkly in the batch-$1$ setting. Isolation, in short, removes overwriting at the cost of integration: the knowledge is retained but cannot be brought to bear on a query, which is precisely the failure our single-model, no-routing read-out avoids.

\section{Parameter Sensitivity} \label{app:sensitivity} Four thresholds govern how readily structure is created and retired, and it is worth saying what each controls before reporting the sweep. The \emph{significance threshold} $\esign$ is the level of $\Pst$ below which a conditioner is discarded as not accounting for its target; lowering it retains more weakly-supported structure. The \emph{reintegration threshold} is how nearly coextensive an upstream's presence must be with its downstream's before the two are merged; \emph{lowering} it merges more readily, and so collapses distinctions that refinement had drawn, while raising it merges less and leaves upstream levels standing after they have stopped carving any distinction. The \emph{removal} and \emph{reintegration rates} set how often each maintenance operation is considered at all, and thus how quickly the population responds to its own statistics. All four were held \emph{fixed} at the values of Sec.~\ref{app:repro} for every run reported in this paper; the sweep below was carried out afterwards, to characterise sensitivity rather than to choose them.

We vary these thresholds one axis at a time around the shipped configuration (two seeds each, geometric read-out; Table~\ref{tab:sensitivity}). The shipped defaults sit at or near the best value on every axis, and no off-default setting exceeds the default meaningfully. Accuracy is fairly stable across the removal rate ($0.848$--$0.873$) and the reintegration rate ($0.855$--$0.863$), and more sensitive to the reintegration threshold ($0.815$--$0.873$) and the significance threshold ($0.795$--$0.873$). The reintegration threshold is the one axis that degrades in \emph{both} directions, and for opposite reasons: at $0.80$ ($0.843$) too much is folded back and distinctions refinement had drawn are collapsed, while at $0.95$ ($0.815$) too little is, and upstreams that have ceased to carve a distinction are left standing as separate levels. Loosening the significance threshold to $0.05$ ($0.795$) instead retains too much spurious structure. The result thus does not hinge on fine-grained tuning, but the two threshold defaults are well-chosen rather than arbitrary. We deliberately do not sweep the reliability gates, which prior analysis shows to be load-bearing: relaxing them causes unbounded structure growth rather than an informative accuracy change.

\begin{table}[t]
\centering
\caption{One-axis-at-a-time threshold sweep (two seeds; final accuracy). Default
column is the shipped configuration (ten-seed baseline $0.8735$).}
\label{tab:sensitivity}
\small
\setlength{\tabcolsep}{4pt}
\begin{tabular}{lccc}
\toprule
Axis (low/def/high) & low & def & high \\
\midrule
Reintegration thresh.\ ($.80/.90/.95$) & 0.843 & \textbf{0.873} & 0.815 \\
Removal rate ($.3/.5/.7$)              & 0.873 & \textbf{0.873} & 0.848 \\
Reintegration rate ($.3/.5/.7$)        & 0.863 & \textbf{0.873} & 0.855 \\
Significance thresh.\ ($.05/.1/.2$)    & 0.795 & \textbf{0.873} & 0.878 \\
\bottomrule
\end{tabular}
\end{table}

\section{Baseline Assumptions in Full} \label{app:assumptions} The main text summarizes this; the fuller statement follows.

The comparison is not between methods on equal footing, and it is more informative to say what each one requires than to normalize them. Every method receives the identical stream of \emph{new} observations---the same five previously-unseen samples of the current class per block, $100$ unique samples per class over the run---and they differ in what they additionally need. The regularisation baselines (SI, MAS, LwF) store no past data, but each requires a \emph{consolidation schedule}---a designated moment at which the parameters currently held are declared the ones to protect---and LwF additionally keeps a full copy of the previous model. Our system requires neither (Sec.~\ref{app:clbaselines}). The replay baselines need past data to be \emph{available, stored, and re-presented} to the network; the replay ratio $r$ sets the buffer size relative to the training batch ($10$ here), so the buffer holds $2$, $5$, or $8$ examples at $r{=}0.2,0.5,0.8$, a re-exposure budget of the same order as the five new samples. Beyond storage, and in common with statistical learning generally, none of these baselines learns on a sample-by-sample basis: each block is a small batch revisited for $20$ gradient updates ($4{,}000$ over the run, against the Modeller's $1{,}000$---one per sample, never revisited). The Modeller stores no past data, re-presents nothing, and needs no buffer.

We deliberately do not compare against methods that require \emph{visible task boundaries}---a signal announcing that one task has ended and another begun. On a genuine stream of experience no such signal exists, and assuming it is difficult to defend: it presupposes exactly the segmentation of experience that a continual learner ought to discover. Instead we include the expert-routing model \citep{erden2025aec} as the representative of the expansion/isolation family, run \emph{without} any task-boundary signal: it must decide for itself, from reconstruction error alone, when a new expert is warranted. That is the same information our system has, and it makes the comparison meaningful rather than merely favorable. It is worth noting that this method still carries a weaker assumption of the same family: although no boundary is announced, a task is expected to persist unchanged long enough for the current expert to stabilize before the input distribution moves on. Our setting grants no such interval---the class in front of the learner changes every five samples---so the assumption is considerably weaker than an explicit boundary but not absent.

For completeness we also examined what happens when the replay assumption is relaxed further---an order-of-magnitude larger buffer, and a variant in which each block supplies a class's full data. Replay improves as expected in proportion to how much past data it stores and re-presents, up to ${\sim}0.93$, but those settings are no longer comparable to the Modeller's; we report them, with the corresponding verification that the baselines here are well-configured, in Sec.~\ref{app:largebuffer}.

\section{Model Growth and Turnover in Full} \label{app:growth} This section gives the full form of the argument summarized in Sec.~\ref{sec:exp-structure}.

\begin{figure}[t]
\centering
\includegraphics[width=\columnwidth]{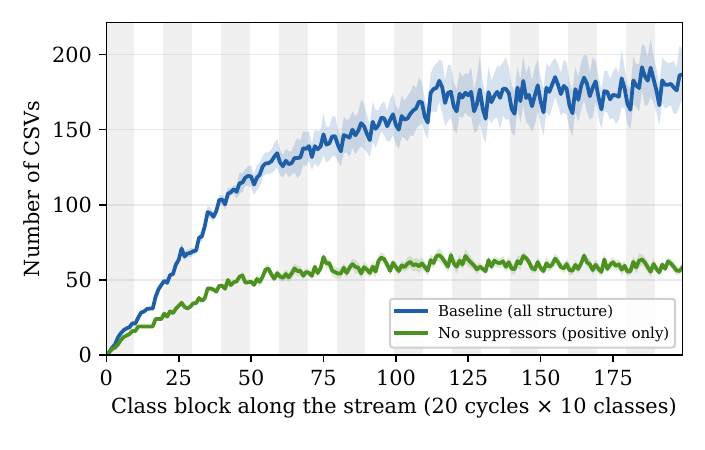}
\caption{The model grows and then saturates. \csv{} count along the
stream (ten seeds; line $=$ mean, band $=\pm1$ s.d.; alternating shading marks
cycles). Growth is steep while classes are novel, decelerates from cycle
${\sim}5$, and levels off around ${\sim}185$ \csv{}s---bounded, not unbounded,
under a stationary class population. The no-suppressor variant (green) isolates
how much of the population is suppressive: it saturates near $60$, so roughly
more than half of the structure the baseline forms is negative, boundary-carving
structure.}
\label{fig:ncsv}
\end{figure}

Because the model \emph{grows} its own structure rather than filling a fixed architecture, its size is an observable of learning, not a hyperparameter (Fig.~\ref{fig:ncsv}). The learned structure grows fast while classes are novel ($\sim\!21$ \csv{}s after cycle $0$, $\sim\!134$ after cycle $5$) and then saturates, plateauing around $\sim\!187$ \csv{}s from cycle ${\sim}11$ onward---the model does not grow without bound, even though novel samples keep arriving at every step and none is ever re-presented (Table~\ref{tab:size} gives the counts by cycle, split by polarity). This is worth checking rather than assuming, since a learner that adds structure in response to what it observes has no built-in ceiling and unbounded growth is the obvious failure mode. What the plateau shows is that the system does not \emph{retain} the structure it keeps generating when what it already has accounts for what it is seeing: it continues to propose---which is what leaves it able to capture a genuine novelty---while retiring proposals that earn nothing, so the population settles instead of accumulating.

The plateau should not be read as a finished, static model. It is an \emph{equilibrium}: the number stops changing while its \emph{membership} does not. Variation continues to propose structure for whatever each new sample raises, and refinement, removal, and reintegration continue to retire whatever fails to reach significance, so a portion of the population at any moment consists of \csv{}s that are formed and later discarded rather than kept. Some of this turnover is directly measurable: between consecutive checkpoints in the plateau the count changes by $7.7$ \csv{}s on average, and by as much as $29$, while the level itself stays near $187$. That figure is only a \emph{lower bound} on the activity, however, and understates it substantially, because it records the \emph{net} change between two checkpoints. Formation and retirement proceed concurrently, so a \csv{} created and another removed in the same interval cancel in the count and leave no trace in the fluctuation at all---a population could be turning over completely while the number never moved. What is needed alongside it is a measure of how much of the population is \emph{long-lived}, one that does not depend on net change; the maturity filter below supplies exactly that, by counting how many \csv{}s have accumulated a given number of lifetime observations and asking what is lost by consulting only those.

This dynamism, and the transient redundancy that comes with it, is a design property rather than an inefficiency to be tuned away. The learner has no way of knowing that the stream will keep presenting the same ten classes; that is our knowledge of the experiment, not information available to it. A system that stopped proposing new structure once its size stabilized would be committing to the assumption that nothing further will ever need to be represented, and would be unable to react if the stream did in fact turn to something new. The continued churn is precisely the capacity to notice novelty, held in reserve; on a stationary class population it shows up as turnover around a fixed level, and on a non-stationary one it is what would allow the model to grow again.

Negative (suppressive) structure comes to dominate ($105$ negative vs.\ $82$ positive at cycle $19$), reflecting that class boundaries, not class prototypes, are where structure keeps being demanded; the no-suppressor variant makes the split visible directly, saturating near a third of the baseline's population (Fig.~\ref{fig:ncsv}). That variant also shows that this particular task can be solved with a much smaller model: suppressing their formation altogether costs no measurable accuracy here (Sec.~\ref{app:variants}, Table~\ref{tab:variants}). We do not adopt it as our main configuration, because ``suppressive structure is dispensable'' is a property of \emph{this} data rather than an assumption the learner could safely make in general---it forms such structure because it is accounting for what it has seen, not because it has been told what will turn out to matter. Notably, accuracy plateaus several cycles \emph{before} size does (cycle ${\sim}5$ vs.\ ${\sim}11$): the late-added structure does not carry the discrimination (Sec.~\ref{app:tables}). This is expected rather than surprising, since the learner has no access to accuracy and is not optimizing it. What it does is account for the statistically significant relationships it encounters, as completely as it can and with as little structure as it can---so it continues to add structure for regularities that are \emph{real} but redundant for telling the ten classes apart.

\subsubsection{The load-bearing fraction of the population.} The read-out lets us measure this split directly, and the answer sharpens the picture above. We score the same trained models while ignoring every \csv{} with fewer than $M$ lifetime presence observations, varying $M$ and changing nothing else (ten seeds; Table~\ref{tab:maturity}). Accuracy is flat throughout: $0.874$ at $M{=}0$, $0.879$ at $M{=}10$, and $0.874$ again at $M{=}40$, every deviation well inside the $\pm0.018$ spread across seeds. What changes is how much of the model is consulted---at $M{=}40$ only $92$ of $187$ \csv{}s remain. \textbf{Half the learned population can be discarded at read-out time with no loss of accuracy whatsoever.}

Taken together with the turnover measured above, this identifies what that half \emph{is}. The \csv{}s that fall below the maturity threshold are those with few lifetime observations---the recently formed and the rarely satisfied---which is the transient population that the equilibrium consists of. So the $\sim\!187$ \csv{}s are not $187$ pieces of predictive structure: they are a persistent, mature core that carries essentially all of the discrimination, plus a comparable body of provisional structure that is being formed and retired around it and contributes nothing to recognizing these ten classes. The two measurements are the same fact seen from two directions---one in time, as churn at a stable level; one in a single frozen model, as a large sub-population whose removal changes nothing. Read together, the two measurements also close the argument of the preceding paragraph. The net fluctuation could not see structure that is created and retired in equal measure; the maturity count sees it directly, as the gap between the full population and the long-lived one. And the provisional body it exposes is not waste to be trimmed. Those \csv{}s are indeed unnecessary for these ten classes---the filter identifies them as such, and discarding them costs nothing---but they are not \emph{errors}: each was a hypothesis about a regularity the data had just presented, and the mechanism that produced it is the same one that would capture a regularity that did matter. They keep being recreated because the system has no way to know, and no basis for assuming, that nothing further will need representing. Pruning the mechanism away permanently would buy a smaller model at the cost of the system's ability to respond to anything new; filtering its products at read-out time, as here, costs nothing and leaves the capacity intact.

\section{Additional Figures and Tables} \label{app:morefigs}

These support claims made in Sec.~\ref{sec:experiments}; most are referenced from the main text at the point they bear on.

\section{Full Experimental Tables} \label{app:tables} 
\begin{figure}[t]
\centering
\includegraphics[width=\columnwidth]{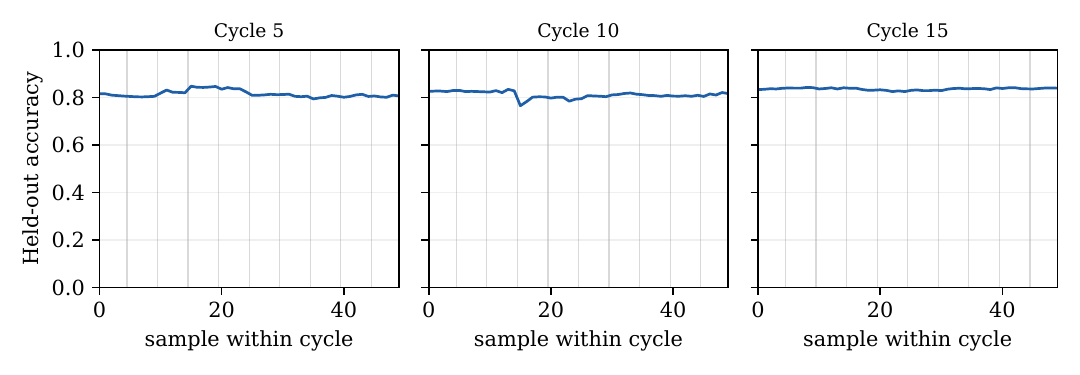}
\caption{Per-sample learning progression within a cycle, for the baseline
configuration at cycles $5$, $10$ and $15$. Those three were fixed in advance rather
than selected afterwards; evaluating after every sample of every cycle would multiply
an already evaluation-dominated cost (Sec.~\ref{app:repro}). Accuracy is recorded
after \emph{each individual sample} rather than only at block ends---fifty checkpoints
per cycle instead of ten---and each point is the accuracy over the entire held-out set,
\emph{averaged across all ten classes}; ten-seed means. The $x$-axis is the sample's position within the cycle, with thin verticals at
the ten class-block boundaries. Accuracy rises as each block's five samples are
integrated and does not fall back as the later blocks train, so the per-block sampling
used elsewhere in the paper conceals nothing between its checkpoints---the
sample-level counterpart of Sec.~\ref{sec:exp-Modeller}.}
\label{fig:persample}
\end{figure}

This section collects the fuller tables behind Sec.~\ref{sec:experiments}; Fig.~\ref{fig:persample} shows, in addition, how accuracy progresses sample by sample within a cycle at three points of the stream.

\begin{table}[t]
\centering
\caption{Model size (\csv{} count) by cycle: total, positive/negative split, and
the count surviving read-out maturity gates $M{\ge}5$ and $M{\ge}10$.}
\label{tab:size}
\small
\begin{tabular}{lccccc}
\toprule
cycle & total & pos & neg & $M{\ge}5$ & $M{\ge}10$ \\
\midrule
$0$  & $21.2$  & $16.0$ & $5.2$   & $8.5$   & $3.0$   \\
$5$  & $134.3$ & $62.8$ & $71.5$  & $95.8$  & $71.5$  \\
$10$ & $168.1$ & $72.1$ & $96.0$  & $130.1$ & $113.4$ \\
$15$ & $177.1$ & $78.7$ & $98.4$  & $149.8$ & $138.5$ \\
$19$ & $186.7$ & $81.7$ & $105.0$ & $154.4$ & $143.8$ \\
\bottomrule
\end{tabular}
\end{table}

Accuracy plateaus by cycle ${\sim}5$ while size saturates only by cycle ${\sim}11$: the discriminative content is concentrated in the mature core (Sec.~\ref{sec:exp-structure}), as the maturity measurement below makes precise.

\begin{table}[t]
\centering
\caption{Read-out maturity filter on the geometric read-out (ten seeds). M: a
\csv{} is consulted only if it has accumulated at least $M$ lifetime presence
observations; nothing else changes, and no model is retrained. Accuracy is flat
throughout while the consulted population halves.}
\label{tab:maturity}
\small
\begin{tabular}{lcccc}
\toprule
$M$ & \csv{}s kept & kept & accuracy & vs.\ $M{=}0$ \\
\midrule
$0$  & $186.7$ & $1.00$ & $0.8735 \pm 0.018$ & --- \\
$5$  & $154.4$ & $0.83$ & $0.8735 \pm 0.019$ & $+0.000$ \\
$10$ & $143.8$ & $0.77$ & $0.8790 \pm 0.021$ & $+0.006$ \\
$20$ & $128.8$ & $0.69$ & $0.8770 \pm 0.023$ & $+0.004$ \\
$40$ & $\mathbf{91.9}$ & $\mathbf{0.49}$ & $\mathbf{0.8735} \pm 0.021$ & $+0.000$ \\
\bottomrule
\end{tabular}
\end{table}

Table~\ref{tab:maturity} gives the measurement discussed in Sec.~\ref{sec:exp-structure}. Every value lies within one standard deviation of $M{=}0$, so the filter costs nothing across the whole range; at $M{=}40$ fewer than half the \csv{}s are consulted and accuracy is exactly what it was with all of them. Since low-maturity \csv{}s are the recently formed and the rarely satisfied, this is the static counterpart of the turnover reported in Sec.~\ref{sec:exp-structure}: the population divides into a mature core that carries the discrimination and a comparable transient body that does not.

The filter is also usable rather than merely diagnostic. Because it is a read-out policy and touches nothing the learner does, a maturity threshold can be fixed in advance and applied whenever the model is consulted, which yields a model that is smaller by the transient fraction at no measured cost in accuracy---roughly half the \csv{}s at $M{=}40$. What one gives up is not accuracy on the classes already learned but the \emph{readiness} the transient population represents: those \csv{}s are the ones a genuinely new regularity would be picked up by, and a threshold set high enough excludes them from being consulted while they remain immature.

\begin{table}[t]
\centering
\caption{Full baseline comparison (ten seeds), grouped by what each method stores.
Final: end-of-cycle accuracy after cycle $19$. Early: mean end-of-cycle accuracy over cycles
$0$--$2$. Rows marked $^\dagger$ come from the separate baseline sweep (Sec.~\ref{app:clbaselines});
all others from the ten-seed regimes runs (Table~\ref{tab:baselines-regimes}). Within-cycle retention
for every method is in Table~\ref{tab:wcr-full}. Regularisation and replay methods are at their best
configuration over the sweeps in Sec.~\ref{app:clbaselines}, selected by final held-out accuracy;
the expansion (Expert-AEC) is swept over its spawn threshold $k$, higher $k$ spawning fewer experts
(count in Notes; ``failed'' = exceeded the $30$-expert cap). Two things stand out:
SI and MAS reach our accuracy while retaining far less within each cycle (Table~\ref{tab:wcr-main}),
and storage buys retention---the same method at $8$ and at $80$ images differs more than
most methods differ from each other. Bold: best replay-free method in each column; replay methods
in italics. $^\star$This work.}
\label{tab:full-baselines}
\small
\setlength{\tabcolsep}{4pt}
\begin{tabular}{lccl}
\toprule
Method & Final & Early & Notes \\
\midrule
SI & \textbf{.879} & .418 & no store \\
Modeller$^\star$ & .874 & \textbf{.546} & no store \\
MAS & .865 & .233 & no store \\
LwF & .810 & .187 & no store \\
Plain NN ($r{=}0$) & .661 & .158 & no store \\
Expert-AEC $k{=}2$ & .607 & .198 & 9.2 exp.\  \\
Expert-AEC $k{=}1^\dagger$ & .482 & .176 & 17.2 exp.\ \\
CNN$^\dagger$ & .328 & .105 & unstable \\
Expert-AEC $k{=}0.5^\dagger$ & .294 & .163 & 27.5 exp.\ \\
Expert-AEC $k{=}0^\dagger$ & .218 & .170 & 31 exp., failed \\
Expert-AEC (bs $1$)$^\dagger$ & .184 & .184 & 31 exp., failed \\
\midrule
\textit{ER} & \textit{.727} & \textit{.428} & \textit{$8$ img} \\
\textit{DER++} & \textit{.797} & \textit{.587} & \textit{$8$ img} \\
\textit{ER-ACE} & \textit{.378} & \textit{.382} & \textit{$8$ img} \\
\midrule
\textit{DER++} & \textit{.911} & \textit{.747} & \textit{$80$ img} \\
\textit{ER-ACE$^\dagger$} & \textit{.825} & \textit{.714} & \textit{$80$ img} \\
\bottomrule
\end{tabular}
\end{table}

\begin{table}[t]
\centering
\caption{Final per-class held-out accuracy under the cyclic reference regime (mean over ten
seeds): our system vs.\ representative baselines. Class $4$ (the $4/9$ confusion) is the Modeller's
weakest; the per-class pattern under the other stream regimes is in
Table~\ref{tab:perclass-regimes}. Bold: best replay-free method in each column; the replay method is in italics. $^\star$This work.}
\label{tab:perclass-full}
\scriptsize
\setlength{\tabcolsep}{2.5pt}
\begin{tabular}{lccccccccccc}
\toprule
 & $0$ & $1$ & $2$ & $3$ & $4$ & $5$ & $6$ & $7$ & $8$ & $9$ & avg \\
\midrule
Modeller$^\star$ & .90 & \textbf{.96} & .87 & .81 & .63 & \textbf{.93} & .93 & \textbf{.94} & \textbf{.85} & \textbf{.96} & \textbf{.88} \\
SI & \textbf{.97} & .91 & \textbf{.93} & \textbf{.84} & .83 & .83 & \textbf{.95} & .86 & .80 & .89 & \textbf{.88} \\
MAS & \textbf{.97} & .95 & .90 & .82 & \textbf{.88} & .86 & .93 & .91 & .61 & .85 & .87 \\
\textit{DER++ (80 img)} & \textit{.96} & \textit{.99} & \textit{.94} & \textit{.89} & \textit{.86} & \textit{.87} & \textit{.93} & \textit{.96} & \textit{.86} & \textit{.89} & \textit{.91} \\
\bottomrule
\end{tabular}
\end{table}

\begin{table}[t]
\centering
\caption{Final per-class held-out accuracy by stream regime (mean over ten seeds): our system vs.\
representative baselines; the cyclic regime is Table~\ref{tab:perclass-full}. As recurrence
weakens (top to bottom) the replay-free regularisers lose whole classes---by blocked only
the last few classes trained survive---while the Modeller and the $80$-image buffer keep every class.
Bold: best replay-free method in each column; the replay method is in italics. $^\star$This work.}
\label{tab:perclass-regimes}
\scriptsize
\setlength{\tabcolsep}{2.5pt}
\begin{tabular}{lccccccccccc}
\toprule
 & $0$ & $1$ & $2$ & $3$ & $4$ & $5$ & $6$ & $7$ & $8$ & $9$ & avg \\
\midrule
\multicolumn{12}{l}{\emph{shuffled}} \\
Modeller$^\star$ & .88 & \textbf{.97} & .90 & .78 & .64 & \textbf{.88} & .94 & .86 & \textbf{.83} & \textbf{.93} & \textbf{.86} \\
SI & .95 & .95 & .89 & \textbf{.80} & .85 & .84 & .96 & .89 & .62 & .86 & \textbf{.86} \\
MAS & \textbf{.96} & .95 & \textbf{.92} & .77 & \textbf{.90} & .85 & \textbf{.97} & \textbf{.92} & .55 & .86 & \textbf{.86} \\
\textit{DER++ (80)} & \textit{.96} & \textit{.98} & \textit{.91} & \textit{.86} & \textit{.90} & \textit{.89} & \textit{.93} & \textit{.92} & \textit{.75} & \textit{.86} & \textit{.89} \\
\midrule
\multicolumn{12}{l}{\emph{permuted}} \\
Modeller$^\star$ & .90 & .94 & .86 & .73 & .64 & \textbf{.86} & .88 & \textbf{.86} & \textbf{.84} & \textbf{.95} & \textbf{.85} \\
SI & .90 & .94 & .87 & .77 & \textbf{.84} & .81 & \textbf{.94} & .84 & .59 & .80 & .83 \\
MAS & \textbf{.96} & \textbf{.97} & \textbf{.92} & \textbf{.80} & .83 & .83 & .93 & .80 & .54 & .77 & .83 \\
\textit{DER++ (80)} & \textit{.98} & \textit{.98} & \textit{.94} & \textit{.87} & \textit{.88} & \textit{.82} & \textit{.93} & \textit{.90} & \textit{.72} & \textit{.86} & \textit{.89} \\
\midrule
\multicolumn{12}{l}{\emph{longgap}} \\
Modeller$^\star$ & .78 & .91 & \textbf{.83} & \textbf{.67} & .59 & \textbf{.89} & \textbf{.90} & \textbf{.75} & \textbf{.78} & \textbf{.80} & \textbf{.79} \\
SI & \textbf{.81} & \textbf{.98} & .50 & .33 & \textbf{.67} & .61 & .89 & .66 & .29 & .47 & .62 \\
MAS & .62 & .90 & .42 & .33 & .62 & .55 & .77 & .67 & .14 & .35 & .53 \\
\textit{DER++ (80)} & \textit{.95} & \textit{.97} & \textit{.94} & \textit{.83} & \textit{.87} & \textit{.83} & \textit{.93} & \textit{.91} & \textit{.71} & \textit{.84} & \textit{.88} \\
\midrule
\multicolumn{12}{l}{\emph{blocked}} \\
Modeller$^\star$ & \textbf{.65} & \textbf{.85} & \textbf{.46} & \textbf{.43} & \textbf{.53} & \textbf{.90} & \textbf{.87} & \textbf{.71} & \textbf{.51} & \textbf{.71} & \textbf{.66} \\
SI & .00 & .10 & .20 & .23 & .00 & .10 & .00 & .21 & .10 & .10 & .10 \\
MAS & .00 & .10 & .20 & .20 & .01 & .10 & .01 & .25 & .10 & .10 & .11 \\
\textit{DER++ (80)} & \textit{.81} & \textit{.97} & \textit{.81} & \textit{.84} & \textit{.49} & \textit{.57} & \textit{.88} & \textit{.71} & \textit{.76} & \textit{.73} & \textit{.75} \\
\bottomrule
\end{tabular}
\end{table}

\section{Scaling} \label{app:scaling}

Three scaling questions are separable; the paper answers the first two and examines the third. The first is whether structure grows without bound as \emph{novel samples} keep arriving: it does not, saturating near $187$ \csv{}s while samples continue to arrive and none is re-presented (Sec.~\ref{app:growth}). The second, addressed here, is how size and per-sample cost grow with the \emph{number of classes}. The third---growth with the \emph{complexity of each input}, meaning the network or semantic complexity of an image (how much structure it takes to describe) rather than the number of pixels or information rank---is examined below on MNIST, with a single-model check on EMNIST (Table~\ref{tab:inputcomplexity}); a systematic study, relating \csv{} count to a defined per-class complexity measure on a suitably graded set of inputs, is left for future work.

\paragraph{Measured quantities.} We reran the identical learner and stream with the class count varied, the classes drawn at random per seed (four seeds each). For each finished model we then replayed the held-out set and counted the calls to the single matcher invocation in the read-out. This is the whole of the expensive work: the read-out loop walks every \csv{}, but for one whose anchors are unbound---whose downstream chain did not match the observation---it performs only a constant-time test and skips before any matching. The integration of the surviving detectors' statistics is arithmetic on scalars and is negligible beside it. \textbf{Learning needs no separate measurement}: it reaches \csv{}s by the same chain-aware pass and is cut by the same condition, so a \csv{} skipped at read-out is skipped when learning from that sample too.

We measure traversal rather than size because that is where the cost is. A \csv{} that is never reached costs storage and nothing further; the computation is entirely in those that are matched, and the model may in principle be arbitrarily large without that changing. We do not treat the storage itself as the pressing question, and not because it is small. Neural networks are known to be substantially overparameterised for the tasks they solve, and that surplus is not incidental but appears to be necessary for them to be trainable at all; a learner that instead retains only what the evidence requires, and retires what it does not (Sec.~\ref{app:growth}), is attempting the opposite discipline by construction. We would therefore not expect size to be more of a problem here than it already is for the models this work is compared against. Establishing that would require tasks far more complex than this design can currently address, so we offer it as an argument about where the burden lies rather than as an experimental claim.

\paragraph{Cost grows more slowly than the model.} From four to ten classes---$2.5\times$ the classes---the model grows $2.80\times$ while the work per observation grows $2.09\times$. The fraction of the model matched falls from $0.59$ to $0.43$ and is flat from six classes on ($0.47$, $0.47$, $0.43$, with standard deviations of $0.02$--$0.03$ across seeds). Roughly half the model is skipped before any matching, and that proportion holds as the model grows. Accuracy does not degrade as classes are added ($0.869$, $0.878$, $0.833$, $0.842$, $0.874$), so the flat cost is not bought by the model quietly getting worse. At ten classes these runs give $0.874\!\pm\!0.014$ over four seeds against the $0.874\!\pm\!0.018$ of the ten-seed main result (Sec.~\ref{sec:exp-Modeller})---an incidental but useful reproduction through a separately written harness. The per-sample maximum, however, approaches the model size ($199$ of $216$ in the largest run): the mean is the right summary, but the worst case is near-full traversal.

\paragraph{The falling matched fraction, and the excess model size.} Both follow from one property of the current design. Each \csv{} conditions a \emph{single} target, so structure shared between classes cannot be referenced from more than one place and is instead rebuilt separately under each class that uses it. For any one observation the copies belonging to other classes fail the anchor test and are cut, so as classes are added a growing share of the model is about classes the observation is not---which is exactly why the matched fraction falls. The same duplication inflates the absolute count.

This identifies the lever. Allowing a \csv{} to condition several immediate targets---a directed acyclic organisation rather than a per-target tree---lets each distinct sub-pattern exist once and be referenced by many parents, which is the minimal shared representation, and it is consistent with the locality the learner already respects (connections to immediate targets, not to a whole chain). The expected effect is not a faster procedure but a smaller model: fewer \csv{}s in total and fewer reached per observation. Note that the fraction would likely \emph{rise} rather than fall, since what survives is shared structure that is relevant more often---which is a further reason to treat the absolute count, not the ratio, as the quantity of interest. We flag one condition: the representational change is necessary but not sufficient on its own, since refinement must also recognise an existing sub-pattern and connect to it rather than spawn a fresh copy. We have not implemented this, and report it as the identified direction rather than a measured improvement.

\paragraph{The traversal is unprioritised.} A second lever is untouched. Within what the cut leaves, the traversal is exhaustive and unordered: wherever a \csv{}'s anchors are bound it is matched, in no particular sequence, with no attempt to visit first those most likely to match or those carrying the most weight in the decision. Nothing in the design requires this. Each \csv{} already maintains the statistics such an ordering would need---how often it has been satisfied, and how specific it is to its own target---so the traversal could be ordered by expected contribution and stopped once the candidates that remain cannot change the prediction, in the manner of a best-first search with a bound. We have implemented no such scheme, and the consequence is worth stating precisely: the $83$ of $191$ in Table~\ref{tab:scaling} is what an \emph{exhaustive} traversal costs, and is therefore the pessimistic case rather than the operating cost of a read-out that ordered its work. Together with the sharing above, the two levers act on different terms---one on how many \csv{}s exist, the other on how many of them need to be visited.

\paragraph{An independent bound at read-out.} Two further results already bound the cost of consulting the model. Ignoring every \csv{} with fewer than $M$ lifetime presence observations leaves accuracy flat while removing half the population at $M{=}40$ (Sec.~\ref{sec:exp-structure}), and because that is a read-out policy touching nothing the learner does, it can be fixed in advance. And the profile in Sec.~\ref{app:repro} shows wall-clock dominated by evaluation---matching \csv{}s against held-out images---rather than by learning, so the quantity measured here is the one that governs runtime.

\paragraph{Model complexity versus input complexity.} The class-count axis above varies \emph{how many} classes must be separated; a distinct question is whether the structure grown for a single class tracks the structural complexity of that class's own images. We examine this with the existing system (Table~\ref{tab:inputcomplexity}). Per-class input complexity is the mean size of the multi-scale network the front end builds for a class's images ($150$ filtered images per class); per-class model complexity is the number of \csv{}s a trained model holds whose target resolves to that class, split into positive detectors and negative suppressors (ten seeds; grand total $186.7\pm17.0$ \csv{}s, range $161$--$216$, with seed~$0$ giving $174$). The two are essentially uncorrelated---Spearman $\rho=-0.07$ ($p=0.86$) between mean \csv{} count and mean node count, and $+0.03$ ($p=0.93$) against mean edge count (both far from significance): the structurally richest digit $8$ ($16.6$ nodes) grows the \emph{fewest} \csv{}s ($3.6$), with $3$ next ($5.0$), while the digits that grow the most---$6$ ($41.9$), $2$ ($33.6$) and $5$ ($30.7$)---are the confusable ones, not the structurally rich ones. Node count is by contrast strongly anti-correlated with accuracy ($\rho=-0.76$, $p=0.01$)---structurally richer digits are harder---while the two input measures agree with one another ($\rho=0.88$). The reading is that \csv{} growth is driven by \emph{discrimination demand}: a class accumulates structure, and especially suppressors, in proportion to how confusable it is with the other nine (the suppressor-heavy classes are $5$, $6$ and $2$), not in proportion to the intrinsic structure of its images. This has limits---ten seeds, on ten MNIST classes, so ``confusability'' is defined only against these nine alternatives, and class~$6$ in particular is unstable across seeds (total $20$--$90$, its suppressors proliferating), so the top of the ranking is fragile---but the contrast is far larger than the seed-to-seed noise. A single-model check on $26$-class EMNIST reproduces the pattern: per-class \csv{} count and mean node count are again uncorrelated (Spearman $\rho=0.15$, $p=0.46$), with the structurally richest letter ($q$, $14.1$ nodes) growing only $8$ \csv{}s while the most \csv{}-heavy letter ($t$, $59$) is of middling structural complexity.

\begin{table}[t]
\centering
\caption{Per-class input complexity versus model complexity (ten seeds). nodes,
edges: mean ($\pm$sd) size of the multi-scale network over $150$ filtered images per class
(seed-independent). pos, neg, \csv{}: positive detectors, negative suppressors
and their total the model grows for the class, ten-seed means (grand total $186.7\pm17.0$; \csv{}
column shown as mean$\pm$sd). acc: ten-seed held-out accuracy. \csv{} count does not track
input size (Spearman $\rho\!\approx\!0$).}
\label{tab:inputcomplexity}
\small
\setlength{\tabcolsep}{4pt}
\begin{tabular}{ccccccc}
\toprule
Class & nodes & edges & pos & neg & \csv{} & acc \\
\midrule
$0$ & $9.3\!\pm\!1.9$  & $67\!\pm\!32$  & $2.2$  & $4.2$  & $6.4\!\pm\!1.4$  & $.90$ \\
$1$ & $5.0\!\pm\!1.9$  & $19\!\pm\!16$  & $1.0$  & $4.9$  & $5.9\!\pm\!0.7$  & $.96$ \\
$2$ & $13.8\!\pm\!3.1$ & $104\!\pm\!48$ & $17.6$ & $16.0$ & $33.6\!\pm\!5.6$ & $.87$ \\
$3$ & $14.3\!\pm\!2.3$ & $102\!\pm\!28$ & $2.2$  & $2.8$  & $5.0\!\pm\!4.6$  & $.81$ \\
$4$ & $12.4\!\pm\!2.1$ & $90\!\pm\!28$  & $7.9$  & $12.9$ & $20.8\!\pm\!4.1$ & $.63$ \\
$5$ & $12.3\!\pm\!2.6$ & $84\!\pm\!31$  & $14.9$ & $15.8$ & $30.7\!\pm\!7.5$ & $.93$ \\
$6$ & $12.3\!\pm\!1.4$ & $98\!\pm\!24$  & $17.1$ & $24.8$ & $41.9\!\pm\!21.7$& $.93$ \\
$7$ & $9.4\!\pm\!2.2$  & $56\!\pm\!24$  & $5.7$  & $12.8$ & $18.5\!\pm\!2.3$ & $.94$ \\
$8$ & $16.6\!\pm\!1.9$ & $210\!\pm\!53$ & $2.2$  & $1.4$  & $3.6\!\pm\!1.1$  & $.85$ \\
$9$ & $11.6\!\pm\!1.8$ & $94\!\pm\!26$  & $10.9$ & $9.4$  & $20.3\!\pm\!2.9$ & $.96$ \\
\bottomrule
\end{tabular}
\end{table}

\section{EMNIST-Letters: Full Results} \label{app:emnist}
This section gives the fuller form of the EMNIST-Letters results summarised in Sec.~\ref{sec:exp-emnist}. The benchmark is the $26$-class EMNIST-Letters set, each class merging the upper- and lower-case forms of a letter (so one class spans two distinct shapes), filtered to single-connected-component glyphs by the component criterion of Sec.~\ref{app:repro}, without the digit-specific loop requirement ($600$ per class, $15{,}600$ in all) and run through the same pipeline as MNIST. Held-out evaluation uses ten images per class, over ten seeds.

\paragraph{Representation ceiling and the lower scaling anchor.} The same fully-connected network trained in the ordinary i.i.d.\ manner on this data budget reaches $0.763$ on $26$-class EMNIST (against $0.90$ on $10$-class MNIST, Sec.~\ref{app:largebuffer})---so the harder benchmark lowers the \emph{representation's} achievable ceiling, as the deliberately minimal feature vocabulary predicts, and both ceilings sit far above the continual baselines' numbers: the gap under the stream is the schedule, not the architecture. Relative to this ceiling the Modeller attains $0.81$ (cyclic $0.62/0.763$) against the best baseline's ${\approx}\,0.62$. As a lower scaling anchor, a $10$-class EMNIST-Letters run (cyclic) reaches $0.660\pm0.021$, below the $10$-class MNIST $0.874$: the letters are harder than digits at equal class count, consistent with the lower ceiling.

\paragraph{Read-out robustness.} The local, absence-aware geometric read-out (Sec.~\ref{sec:readout}) was carried over to EMNIST \emph{without any per-dataset retuning}---the same terms and the same weights used on MNIST---so the results below are not the product of tuning the read-out to the new benchmark.

\begin{table*}[tbp]
\centering
\caption{EMNIST-Letters, fuller per-regime results (mean$\pm$sd). Final accuracy under all five
regimes, and within-cycle retention WCR$_{\mathrm{end}}$ at cycles $1$, $10$ and $19$
(c1, c10, c19) for the two recurring regimes; the Modeller and the four replay-free baselines run on EMNIST. The late-cycle column
shows the baselines recovering by maximal re-exposure, flattering them relative to the early
column where continual learning is under genuine stress. Bold: best in each column (all methods are replay-free). $^\star$This work.}
\label{tab:emnist-regimes-full}
\scriptsize
\setlength{\tabcolsep}{3pt}
\begin{tabular}{lccccc|ccc|ccc}
\toprule
 & \multicolumn{5}{c}{\emph{Final accuracy}} & \multicolumn{3}{c}{\emph{WCR cyclic}} & \multicolumn{3}{c}{\emph{WCR shuffled}} \\
\cmidrule(lr){2-6}\cmidrule(lr){7-9}\cmidrule(lr){10-12}
Method & cyc & shuf & perm & long & blk & c1 & c10 & c19 & c1 & c10 & c19 \\
\midrule
Modeller$^\star$ & $\mathbf{.62}\!\pm\!.02$ & $\mathbf{.63}\!\pm\!.03$ & $\mathbf{.61}\!\pm\!.03$ & $\mathbf{.63}\!\pm\!.03$ & $\mathbf{.30}\!\pm\!.06$ & $\mathbf{.72}\!\pm\!.02$ & $\mathbf{.95}\!\pm\!.03$ & $\mathbf{.97}\!\pm\!.02$ & $\mathbf{.74}\!\pm\!.04$ & $\mathbf{.95}\!\pm\!.02$ & $\mathbf{.98}\!\pm\!.01$ \\
MAS & $.48\!\pm\!.06$ & $.49\!\pm\!.08$ & $.48\!\pm\!.03$ & $.22\!\pm\!.09$ & $.05\!\pm\!.02$ & $.14\!\pm\!.04$ & $.60\!\pm\!.13$ & $.74\!\pm\!.08$ & $.15\!\pm\!.03$ & $.64\!\pm\!.07$ & $.74\!\pm\!.11$ \\
SI & $.40\!\pm\!.06$ & $.45\!\pm\!.05$ & $.37\!\pm\!.04$ & $.09\!\pm\!.03$ & $.04\!\pm\!.00$ & $.08\!\pm\!.02$ & $.36\!\pm\!.05$ & $.51\!\pm\!.09$ & $.06\!\pm\!.02$ & $.37\!\pm\!.07$ & $.56\!\pm\!.05$ \\
LwF & $.29\!\pm\!.05$ & $.31\!\pm\!.06$ & $.26\!\pm\!.03$ & $.04\!\pm\!.01$ & $.04\!\pm\!.00$ & $.06\!\pm\!.02$ & $.22\!\pm\!.05$ & $.36\!\pm\!.06$ & $.06\!\pm\!.03$ & $.23\!\pm\!.05$ & $.39\!\pm\!.07$ \\
Plain NN & $.29\!\pm\!.04$ & $.39\!\pm\!.06$ & $.34\!\pm\!.06$ & $.07\!\pm\!.03$ & $.04\!\pm\!.00$ & $.07\!\pm\!.02$ & $.19\!\pm\!.03$ & $.33\!\pm\!.05$ & $.08\!\pm\!.02$ & $.33\!\pm\!.05$ & $.47\!\pm\!.08$ \\
\bottomrule
\end{tabular}
\end{table*}

\subsection{Comprehensibility on Letters} \label{app:emnist-interp}
\begin{figure*}[tbp]
\centering
\includegraphics[width=0.8\textwidth]{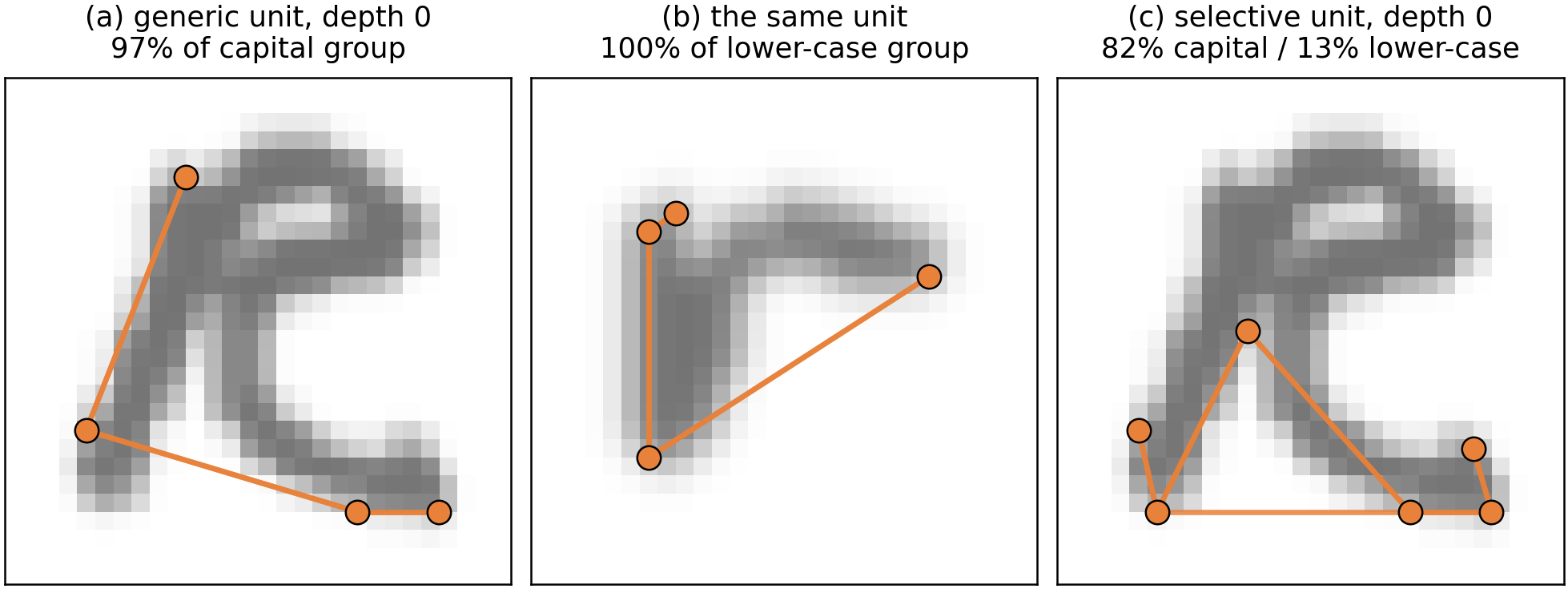}
\caption{Where case information is carried, for the letter class \emph{r} (one trained model), drawn over
held-out glyphs (nodes grounded on the matched observation points). Clustering the class's firing
patterns splits its glyphs into a capital-dominated group and a lower-case group. (a, b) A depth-$0$
unit fires on both groups ($0.97$ and $1.00$)---it is case-generic. (c) Another depth-$0$ unit is
case-selective, firing on the capital group ($0.82$) but rarely on lower-case \emph{r} ($0.13$):
for this letter, whose capital adds structure the lower-case form lacks, the split is carved already at
depth $0$. The distinction is readable directly off the explicit units.}
\label{fig:emnist-interp}
\end{figure*}

The same drawability carries to EMNIST (Fig.~\ref{fig:emnist-interp}). A letter's mature \csv{}s read as caricatures of its shape exactly as the digit models do, and because each class merges two case variants the explicit chain exposes \emph{where} case information is carried: some units of a class are case-generic---present in both the upper- and lower-case forms---while others are case-selective, carving the variant-specific structure, and the depth at which the split appears varies by letter. For \emph{h}, whose two forms share their coarse shape, the shallowest units are case-generic and the split arises one to two levels deeper; for \emph{r}, whose capital adds structure the lower-case form lacks, capital-selective units appear already at depth~$0$ (Fig.~\ref{fig:emnist-interp}). This, too, is readable off the chain. This is the same transparency reported for MNIST (Sec.~\ref{sec:exp-interp}), now exposing finer, case-level structure.

\section{Post-Saturation Novelty} \label{app:novelty}
This section details the novelty probe of Sec.~\ref{sec:exp-structure}, run on MNIST. Per seed we hold out one random (seed-dependent) class, saturate on the other nine under the cyclic schedule, then inject the held-out class and continue cyclically over all ten (ten seeds). It tests the Modeller's \emph{preserved-capacity} property---that structure generation stays available for genuine novelty after the model has saturated.

\begin{figure}[tbp]
\centering
\includegraphics[width=0.82\columnwidth]{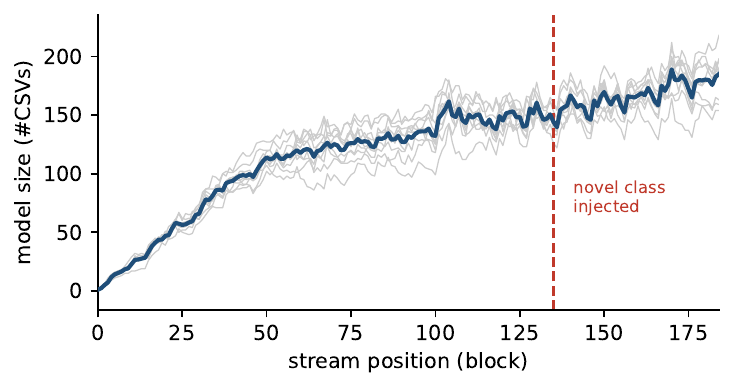}
\caption{Post-saturation novelty on MNIST (Modeller-only; mean over ten seeds).
Model size (\csv{} count) along the stream: flat at the saturated plateau through the pre-phase, a
step up at the injection of the previously-unseen class (dashed line) to a new, higher equilibrium
where it re-stabilises---structure generation reopens for genuine novelty and re-equilibrates rather
than running away.}
\label{fig:novelty}
\end{figure}

The signature is the model-size trace (Fig.~\ref{fig:novelty}). The population is flat at its saturated plateau ($\sim\!150$ \csv{}s) through the pre-phase; at injection it grows to account for the unseen class---rising with a transient peak near $\sim\!189$ and re-stabilising at a new, higher equilibrium ($\sim\!178$, net $\sim\!{+}28$ over the plateau)---so generation reopens precisely when existing structure cannot account for what arrives and then re-equilibrates rather than running away. The injected class is acquired (held-out accuracy $0.825\pm0.072$) and the previously-learned classes are retained ($0.849\pm0.046$ at the end, against $0.814\pm0.059$ at the moment of injection); once the class is present the stream is the cyclic regime already reported (Table~\ref{tab:novelty}).

\begin{table}[tbp]
\centering
\caption{Post-saturation novelty on MNIST (Modeller-only, ten seeds; mean$\pm$sd where applicable). A previously-unseen class is injected after saturation; the model grows to a new, higher equilibrium and learns the class while the pre-trained classes are retained.}
\label{tab:novelty}
\small
\begin{tabular}{lc}
\toprule
Quantity & Value \\
\midrule
Plateau size (pre-injection) & $\sim\!150$ \csv{}s \\
Transient peak after injection & $\sim\!189$ \csv{}s \\
New equilibrium size (settled) & $\sim\!178$ \csv{}s ($\sim\!{+}28$) \\
Novel class, after learning & $0.825\pm0.072$ \\
Old classes, at injection & $0.814\pm0.059$ \\
Old classes, at end (retained) & $0.849\pm0.046$ \\
\bottomrule
\end{tabular}
\end{table}

\section{Reproducibility} \label{app:repro} We record the full configuration used for every reported Modeller result.

\subsubsection{Data.} We use MNIST \citep{lecun1998gradient}, filtered for topological consistency, at native $28\times28$ resolution and binarized at intensity threshold $0.5$; $59{,}595$ of the $70{,}000$ images are kept.

\emph{Rationale for filtering.} The filtering is a scope simplification, not a requirement of the method. The present feature construction reads structure off clean contours and does not model topological \emph{gaps} (breaks or disconnections within a stroke), just as it does not model smooth intensity transitions; both are handleable in principle by the same change-point construction---gaps as additional structure, transitions as higher orders of change (Sec.~\ref{sec:changepoint})---but incorporating them would be a lengthy extension outside what this paper evaluates. We therefore exclude the affected samples and keep only clean, single-structure digits with the expected loops. Crucially, to keep the comparison fair, \emph{all} baselines are trained and evaluated on this same filtered subset; no method sees data the others do not.

\emph{The filtering criterion.} Topology is measured through the model's own image-to-network front end, so that a loop which does not close under the model's binarization genuinely is not there for the model and the sample is dropped. An image is binarized (no Gaussian smoothing, threshold $0.5$) and its contours extracted with a nesting hierarchy, discarding contours shorter than $10$ pixels so that antialiasing speckle is not counted; contours at even nesting depth are outer boundaries (connected components) and those at odd depth are holes. A sample is kept if and only if
\begin{enumerate}\itemsep1pt
\item it has exactly \emph{one} connected component---more than one outer contour means
disconnected strokes, and the sample is dropped; \emph{and}
\item its hole count matches the digit, for those digits that are expected to enclose
regions: $0\!\rightarrow\!1$, $6\!\rightarrow\!1$, $9\!\rightarrow\!1$,
$8\!\rightarrow\!2$. Digits $1,2,3,4,5,7$ carry no hole constraint and may have any
number.
\end{enumerate}

\subsubsection{Stream and evaluation.} Ten classes ($0$--$9$) in fixed order; $20$ cycles; $5$ samples per class per cycle from disjoint indices; one learning step per sample; no re-presentation, replay, or task/cycle-boundary signal. Held-out evaluation uses $20$ images per class at a fixed disjoint offset. All Modeller results average $10$ seeds ($0$--$9$), each fixing the sample draw and the model's internal random state. A run is therefore exactly reproducible from its seed, and we verified this rather than assuming it: retraining all ten seeds from scratch reproduced both the final accuracy and the final \csv{} count of the original run in every case ($10/10$ exact matches). Results are additionally invariant to the order in which internal identifiers are assigned. Trained models can optionally be serialized (per cycle, per class block, or at the end of a run) so that read-out variants such as the maturity filter of Table~\ref{tab:maturity} can be re-scored without retraining.

\subsubsection{Representation.} Contours are extracted from the binarized image (outer boundaries and holes distinguished); each is traversed and orientation-change points read off the raw contour (typed by change axis, direction, convexity; no polygonal approximation). Nodes are wired into \emph{contour} and \emph{spatial} layers, horizontal and vertical variants, and coarsened by successive pair contraction with per-level recomputation of spatial edges; the edges surviving at every level are accumulated into one augmented network. The no-multi-scale ablation uses the finest level only.

\subsubsection{Learning hyperparameters (fixed across all runs).} Significance threshold $\esign=0.1$; removal rate $0.5$; reintegration rate $0.5$; spawn reliability threshold $0.9$; refinement reliability threshold $0.9$; stale-statistics coefficient $0.5$; refinement match threshold $0.5$ on combined node-and-edge coverage; depth-scaled significance enabled. Matching uses the spatial multi-network matcher over the horizontal (\texttt{contour\_h}, \texttt{spatial\_h}) and vertical (\texttt{contour\_v}, \texttt{spatial\_v}) layers.

\subsubsection{Read-out.} The local geometric read-out (Sec.~\ref{sec:readout}) uses position weight $w_p=3.0$ with a variance floor of $1.5$\,px on the position anchors, orientation weight $w_o=1.0$ with concentration $\kappa=2.0$, and add-$\tfrac12$ smoothing on the firing counters; a node or edge anchor contributes only once it has accumulated at least two observations. The alternative read-out of Sec.~\ref{app:readout} uses the binary-integration log-odds rule (Eq.~\ref{eq:readout}) with base-rate pivot, anchored chain-aware matching, and maturity filter $M=0$ unless stated. Both are deterministic given a trained model.

\subsubsection{Metrics.} Three quantities are reported. \emph{Accuracy} is the fraction of held-out images classified correctly, computed per class over the twenty held-out images of that class and then averaged over the ten classes, so every class contributes equally regardless of its frequency in MNIST. \emph{End-of-cycle} accuracy is that quantity measured at the last block of a cycle, and \emph{Early} its mean over cycles $0$--$2$. \emph{Forgetting} (retention loss) is computed per class as the difference between that class's highest accuracy at any checkpoint and its accuracy at the end of the run, then averaged over classes---so it measures what was reached and not held, and is zero for a class that never declines. Within-cycle retention is defined in Sec.~\ref{app:wcr}.

\subsubsection{Seeds and repetitions.} Headline Modeller results and all neural baselines average ten seeds ($0$--$9$); the multi-scale ablation (Sec.~\ref{app:ablation}) uses five seeds per cell, the threshold sweep (Sec.~\ref{app:sensitivity}) two seeds per configuration, the enlarged-buffer replay study (Sec.~\ref{app:largebuffer}) and the preservation audit (Sec.~\ref{app:preserve}) three seeds, and the class-count scaling sweep (Table~\ref{tab:scaling}) four seeds per class count. Each seed fixes both the sample draw and the model's internal random state, and results are invariant to the order in which internal identifiers are assigned.

\subsubsection{Baselines.} Run under the identical stream and held-out sets. Plain/replay: a fully connected network (Flatten--Dense$256$--Dense$256$--Dense$10$, softmax), batch $10$, $20$ epochs per class block, replay buffer ratios $r\in\{0,0.2,0.5,0.8\}$. CNN: two $3\times3$ conv--pool blocks of $32$ filters each, then Dense$128$--Dense$10$. Expert-AEC: per expert a predictor (Dense$128$) and an autoencoder (Dense$64$, sigmoid output), a new expert spawned when the minimum reconstruction error exceeds a $k$-scaled threshold, failing once more than $30$ experts are spawned; configurations vary $k$ and batch size. All hidden units are ReLU. Every classifier is trained with Adam at its framework defaults (learning rate $10^{-3}$) under sparse categorical cross-entropy, and every autoencoder with Adam under mean squared error; each presented batch is one gradient step, so the $20$ epochs per class block are $20$ such steps. Baselines run in an isolated TensorFlow/Keras environment on CPU.

\subsubsection{Compute.} All experiments run on a single machine: a $6$-core (12-thread) Intel Core i7-9750H at $2.6$\,GHz with $16$\,GB of RAM, under macOS $15.7$. No GPU is used by our system at any point. Seeds are run as independent single-threaded processes (BLAS threading pinned to one thread per process) with several in parallel.

Two software environments are used, and the separation is itself informative: our system requires no deep-learning framework, running on Python $3.13$ with NumPy $2.2$, NetworkX $3.5$, SciPy $1.16$ and OpenCV $4.12$ (the last only for contour extraction), while TensorFlow $2.16$ / Keras $3.15$ (Python $3.11$, NumPy $1.26$) appear \emph{only} in the neural baselines.

For wall-clock, a single twenty-cycle seed takes ${\sim}7$--$8$ minutes when the model is evaluated once at the end, and ${\sim}55$--$60$ minutes when it is evaluated after every one of the $200$ class blocks, as in the retention matrices. The difference makes the cost profile clear: the $1{,}000$ learning steps are the cheap part, and wall-clock is dominated by \emph{evaluation}---that is, by repeatedly matching every \csv{} against every held-out image---rather than by learning.

\renewcommand{\refname}{References}
\putbib
\end{bibunit}


% Generated by IEEEtranN.bst, version: 1.14 (2015/08/26)
\begin{thebibliography}{45}
\providecommand{\natexlab}[1]{#1}
\providecommand{\url}[1]{#1}
\csname url@samestyle\endcsname
\providecommand{\newblock}{\relax}
\providecommand{\bibinfo}[2]{#2}
\providecommand{\BIBentrySTDinterwordspacing}{\spaceskip=0pt\relax}
\providecommand{\BIBentryALTinterwordstretchfactor}{4}
\providecommand{\BIBentryALTinterwordspacing}{\spaceskip=\fontdimen2\font plus
\BIBentryALTinterwordstretchfactor\fontdimen3\font minus
  \fontdimen4\font\relax}
\providecommand{\BIBforeignlanguage}[2]{{%
\expandafter\ifx\csname l@#1\endcsname\relax
\typeout{** WARNING: IEEEtranN.bst: No hyphenation pattern has been}%
\typeout{** loaded for the language `#1'. Using the pattern for}%
\typeout{** the default language instead.}%
\else
\language=\csname l@#1\endcsname
\fi
#2}}
\providecommand{\BIBdecl}{\relax}
\BIBdecl

\bibitem[McCloskey and Cohen(1989)]{mccloskey1989catastrophic}
M.~McCloskey and N.~J. Cohen, ``Catastrophic interference in connectionist
  networks: The sequential learning problem,'' in \emph{Psychology of Learning
  and Motivation}.\hskip 1em plus 0.5em minus 0.4em\relax Academic Press, 1989,
  vol.~24, pp. 109--165.

\bibitem[van~de Ven et~al.(2025)van~de Ven, Soures, and
  Kudithipudi]{vandeven2024continual}
G.~M. van~de Ven, N.~Soures, and D.~Kudithipudi, ``Continual learning and
  catastrophic forgetting,'' in \emph{Learning and Memory: A Comprehensive
  Reference}, 3rd~ed., J.~T. Wixted, Ed.\hskip 1em plus 0.5em minus 0.4em\relax
  Academic Press, 2025, vol.~1, pp. 153--168.

\bibitem[Erden and Faltings(2026)]{erden2025foundations}
Z.~D. Erden and B.~Faltings, ``Foundations of a developmental design paradigm
  for integrated continual learning, deliberative behavior, and
  comprehensibility,'' \emph{IEEE Transactions on Emerging Topics in
  Computational Intelligence}, vol.~10, no.~2, pp. 1738--1752, 2026.

\bibitem[Erden(2025)]{erden2025thesis}
Z.~D. Erden, ``Foundations of a new learning paradigm in {AI} grounded in the
  principles of evolutionary developmental biology,'' Ph.D. dissertation, EPFL,
  2025.

\bibitem[Weng et~al.(2001)Weng, McClelland, Pentland, Sporns, Stockman, Sur,
  and Thelen]{weng2001autonomous}
J.~Weng \emph{et~al.}, ``Autonomous mental development by robots and animals,''
  \emph{Science}, vol. 291, no. 5504, pp. 599--600, 2001.

\bibitem[Marr and Nishihara(1978)]{marr1978representation}
D.~Marr and H.~K. Nishihara, ``Representation and recognition of the spatial
  organization of three-dimensional shapes,'' \emph{Proceedings of the Royal
  Society of London. Series B. Biological Sciences}, vol. 200, no. 1140, pp.
  269--294, 1978.

\bibitem[Biederman(1987)]{biederman1987recognition}
I.~Biederman, ``Recognition-by-components: A theory of human image
  understanding,'' \emph{Psychological Review}, vol.~94, no.~2, pp. 115--147,
  1987.

\bibitem[LeCun et~al.(1998)LeCun, Bottou, Bengio, and
  Haffner]{lecun1998gradient}
Y.~LeCun, L.~Bottou, Y.~Bengio, and P.~Haffner, ``Gradient-based learning
  applied to document recognition,'' \emph{Proceedings of the IEEE}, vol.~86,
  no.~11, pp. 2278--2324, 1998.

\bibitem[Cohen et~al.(2017)Cohen, Afshar, Tapson, and van
  Schaik]{cohen2017emnist}
G.~Cohen, S.~Afshar, J.~Tapson, and A.~van Schaik, ``{EMNIST}: Extending
  {MNIST} to handwritten letters,'' in \emph{International Joint Conference on
  Neural Networks (IJCNN)}, 2017, pp. 2921--2926.

\bibitem[Zenke et~al.(2017)Zenke, Poole, and Ganguli]{zenke2017si}
F.~Zenke, B.~Poole, and S.~Ganguli, ``Continual learning through synaptic
  intelligence,'' in \emph{Proceedings of the 34th International Conference on
  Machine Learning}, 2017, pp. 3987--3995.

\bibitem[Aljundi et~al.(2018)Aljundi, Babiloni, Elhoseiny, Rohrbach, and
  Tuytelaars]{aljundi2018mas}
R.~Aljundi, F.~Babiloni, M.~Elhoseiny, M.~Rohrbach, and T.~Tuytelaars, ``Memory
  aware synapses: Learning what (not) to forget,'' in \emph{Proceedings of the
  European Conference on Computer Vision (ECCV)}, 2018, pp. 139--154.

\bibitem[Li and Hoiem(2018)]{li2018lwf}
Z.~Li and D.~Hoiem, ``Learning without forgetting,'' \emph{IEEE Transactions on
  Pattern Analysis and Machine Intelligence}, vol.~40, no.~12, pp. 2935--2947,
  2018.

\bibitem[Buzzega et~al.(2020)Buzzega, Boschini, Porrello, Abati, and
  Calderara]{buzzega2020derpp}
P.~Buzzega, M.~Boschini, A.~Porrello, D.~Abati, and S.~Calderara, ``Dark
  experience for general continual learning: a strong, simple baseline,'' in
  \emph{Advances in Neural Information Processing Systems}, vol.~33, 2020, pp.
  15\,920--15\,930.

\bibitem[Caccia et~al.(2022)Caccia, Aljundi, Asadi, Tuytelaars, Pineau, and
  Belilovsky]{caccia2022erace}
L.~Caccia, R.~Aljundi, N.~Asadi, T.~Tuytelaars, J.~Pineau, and E.~Belilovsky,
  ``New insights on reducing abrupt representation change in online continual
  learning,'' in \emph{International Conference on Learning Representations},
  2022.

\bibitem[Erden et~al.(2025)Erden, Gasmi, and Faltings]{erden2025aec}
Z.~D. Erden, D.~Gasmi, and B.~Faltings, ``Continual reinforcement learning via
  autoencoder-driven task and new environment recognition,'' in \emph{Adaptive
  and Learning Agents (ALA) and Autonomous Robots and Multirobot Systems (ARMS)
  Workshops at the 24th International Conference on Autonomous Agents and
  Multiagent Systems (AAMAS)}, 2025.

\bibitem[Witkin(1983)]{witkin1983scale}
A.~P. Witkin, ``Scale-space filtering,'' in \emph{International Joint
  Conference on Artificial Intelligence (IJCAI)}, 1983, pp. 1019--1022.

\bibitem[Lindeberg(1998)]{lindeberg1998feature}
T.~Lindeberg, ``Feature detection with automatic scale selection,''
  \emph{International Journal of Computer Vision}, vol.~30, no.~2, pp. 77--116,
  1998.

\bibitem[Burt and Adelson(1983)]{burt1983laplacian}
P.~J. Burt and E.~H. Adelson, ``The {Laplacian} pyramid as a compact image
  code,'' \emph{IEEE Transactions on Communications}, vol.~31, no.~4, pp.
  532--540, 1983.

\bibitem[Mallat(1989)]{mallat1989theory}
S.~G. Mallat, ``A theory for multiresolution signal decomposition: The wavelet
  representation,'' \emph{IEEE Transactions on Pattern Analysis and Machine
  Intelligence}, vol.~11, no.~7, pp. 674--693, 1989.

\bibitem[Kropatsch(1995)]{kropatsch1995building}
W.~G. Kropatsch, ``Building irregular pyramids by dual-graph contraction,''
  \emph{IEE Proceedings---Vision, Image and Signal Processing}, vol. 142,
  no.~6, pp. 366--374, 1995.

\bibitem[Lowe(2004)]{lowe2004distinctive}
D.~G. Lowe, ``Distinctive image features from scale-invariant keypoints,''
  \emph{International Journal of Computer Vision}, vol.~60, no.~2, pp. 91--110,
  2004.

\bibitem[Belongie et~al.(2002)Belongie, Malik, and Puzicha]{belongie2002shape}
S.~Belongie, J.~Malik, and J.~Puzicha, ``Shape matching and object recognition
  using shape contexts,'' \emph{IEEE Transactions on Pattern Analysis and
  Machine Intelligence}, vol.~24, no.~4, pp. 509--522, 2002.

\bibitem[Marr(1982)]{marr1982vision}
D.~Marr, \emph{Vision: A Computational Investigation into the Human
  Representation and Processing of Visual Information}.\hskip 1em plus 0.5em
  minus 0.4em\relax W.H. Freeman, 1982.

\bibitem[Fukushima(1980)]{fukushima1980neocognitron}
K.~Fukushima, ``Neocognitron: A self-organizing neural network model for a
  mechanism of pattern recognition unaffected by shift in position,''
  \emph{Biological Cybernetics}, vol.~36, no.~4, pp. 193--202, 1980.

\bibitem[Riesenhuber and Poggio(1999)]{riesenhuber1999hierarchical}
M.~Riesenhuber and T.~Poggio, ``Hierarchical models of object recognition in
  cortex,'' \emph{Nature Neuroscience}, vol.~2, no.~11, pp. 1019--1025, 1999.

\bibitem[Messmer and Bunke(1998)]{messmer1998new}
B.~T. Messmer and H.~Bunke, ``A new algorithm for error-tolerant subgraph
  isomorphism detection,'' \emph{IEEE Transactions on Pattern Analysis and
  Machine Intelligence}, vol.~20, no.~5, pp. 493--504, 1998.

\bibitem[Siddiqi et~al.(1999)Siddiqi, Shokoufandeh, Dickinson, and
  Zucker]{siddiqi1999shock}
K.~Siddiqi, A.~Shokoufandeh, S.~J. Dickinson, and S.~W. Zucker, ``Shock graphs
  and shape matching,'' \emph{International Journal of Computer Vision},
  vol.~35, no.~1, pp. 13--32, 1999.

\bibitem[Felzenszwalb et~al.(2010)Felzenszwalb, Girshick, McAllester, and
  Ramanan]{felzenszwalb2010object}
P.~F. Felzenszwalb, R.~B. Girshick, D.~McAllester, and D.~Ramanan, ``Object
  detection with discriminatively trained part-based models,'' \emph{IEEE
  Transactions on Pattern Analysis and Machine Intelligence}, vol.~32, no.~9,
  pp. 1627--1645, 2010.

\bibitem[Kipf and Welling(2017)]{kipf2017semi}
T.~N. Kipf and M.~Welling, ``Semi-supervised classification with graph
  convolutional networks,'' in \emph{International Conference on Learning
  Representations (ICLR)}, 2017.

\bibitem[Lake et~al.(2015)Lake, Salakhutdinov, and Tenenbaum]{lake2015human}
B.~M. Lake, R.~Salakhutdinov, and J.~B. Tenenbaum, ``Human-level concept
  learning through probabilistic program induction,'' \emph{Science}, vol. 350,
  no. 6266, pp. 1332--1338, 2015.

\bibitem[Sabour et~al.(2017)Sabour, Frosst, and Hinton]{sabour2017dynamic}
S.~Sabour, N.~Frosst, and G.~E. Hinton, ``Dynamic routing between capsules,''
  in \emph{Advances in Neural Information Processing Systems}, 2017, pp.
  3856--3866.

\bibitem[Carpenter and Grossberg(1987)]{carpenter1987massively}
G.~A. Carpenter and S.~Grossberg, ``A massively parallel architecture for a
  self-organizing neural pattern recognition machine,'' \emph{Computer Vision,
  Graphics, and Image Processing}, vol.~37, no.~1, pp. 54--115, 1987.

\bibitem[Fisher(1987)]{fisher1987knowledge}
D.~H. Fisher, ``Knowledge acquisition via incremental conceptual clustering,''
  \emph{Machine Learning}, vol.~2, no.~2, pp. 139--172, 1987.

\bibitem[Fritzke(1995)]{fritzke1995growing}
B.~Fritzke, ``A growing neural gas network learns topologies,'' in
  \emph{Advances in Neural Information Processing Systems 7}, 1995, pp.
  625--632.

\bibitem[Kohonen(1982)]{kohonen1982self}
T.~Kohonen, ``Self-organized formation of topologically correct feature maps,''
  \emph{Biological Cybernetics}, vol.~43, no.~1, pp. 59--69, 1982.

\bibitem[Rebuffi et~al.(2017)Rebuffi, Kolesnikov, Sperl, and
  Lampert]{rebuffi2017icarl}
S.-A. Rebuffi, A.~Kolesnikov, G.~Sperl, and C.~H. Lampert, ``{iCaRL}:
  Incremental classifier and representation learning,'' in \emph{IEEE
  Conference on Computer Vision and Pattern Recognition (CVPR)}, 2017.

\bibitem[Kirkpatrick et~al.(2017)Kirkpatrick, Pascanu, Rabinowitz, Veness,
  Desjardins, Rusu, Milan, Quan, Ramalho, Grabska-Barwinska, Hassabis, Clopath,
  Kumaran, and Hadsell]{kirkpatrick2017overcoming}
J.~Kirkpatrick \emph{et~al.}, ``Overcoming catastrophic forgetting in neural
  networks,'' \emph{Proceedings of the National Academy of Sciences}, vol. 114,
  no.~13, pp. 3521--3526, 2017.

\bibitem[Aljundi et~al.(2019)Aljundi, Kelchtermans, and
  Tuytelaars]{aljundi2019taskfree}
R.~Aljundi, K.~Kelchtermans, and T.~Tuytelaars, ``Task-free continual
  learning,'' in \emph{IEEE/CVF Conference on Computer Vision and Pattern
  Recognition (CVPR)}, 2019, pp. 11\,246--11\,255.

\bibitem[Aljundi et~al.(2017)Aljundi, Chakravarty, and
  Tuytelaars]{aljundi2017expertgate}
R.~Aljundi, P.~Chakravarty, and T.~Tuytelaars, ``Expert gate: Lifelong learning
  with a network of experts,'' in \emph{IEEE Conference on Computer Vision and
  Pattern Recognition (CVPR)}, 2017.

\bibitem[De~Lange et~al.(2022)De~Lange, Aljundi, Masana, Parisot, Jia,
  Leonardis, Slabaugh, and Tuytelaars]{delange2021continual}
M.~De~Lange \emph{et~al.}, ``A continual learning survey: Defying forgetting in
  classification tasks,'' \emph{IEEE Transactions on Pattern Analysis and
  Machine Intelligence}, vol.~44, no.~7, pp. 3366--3385, 2022.

\bibitem[Parisi et~al.(2019)Parisi, Kemker, Part, Kanan, and
  Wermter]{parisi2019continual}
G.~I. Parisi, R.~Kemker, J.~L. Part, C.~Kanan, and S.~Wermter, ``Continual
  lifelong learning with neural networks: A review,'' \emph{Neural Networks},
  vol. 113, pp. 54--71, 2019.

\bibitem[Wang et~al.(2022{\natexlab{a}})Wang, Zhang, Lee, Zhang, Sun, Ren, Su,
  Perot, Dy, and Pfister]{wang2022l2p}
Z.~Wang \emph{et~al.}, ``Learning to prompt for continual learning,'' in
  \emph{Proceedings of the IEEE/CVF Conference on Computer Vision and Pattern
  Recognition (CVPR)}, 2022, pp. 139--149.

\bibitem[Wang et~al.(2022{\natexlab{b}})Wang, Zhang, Ebrahimi, Sun, Zhang, Lee,
  Ren, Su, Perot, Dy, and Pfister]{wang2022dualprompt}
------, ``{DualPrompt}: Complementary prompting for rehearsal-free continual
  learning,'' in \emph{Proceedings of the European Conference on Computer
  Vision (ECCV)}, 2022, pp. 631--648.

\bibitem[Zhuang et~al.(2024)Zhuang, Liu, He, Tong, Zeng, Chen, Wang, and
  Chau]{zhuang2024foal}
H.~Zhuang \emph{et~al.}, ``{F-OAL}: Forward-only online analytic learning with
  fast training and low memory footprint in class incremental learning,'' in
  \emph{Advances in Neural Information Processing Systems (NeurIPS)}, vol.~37,
  2024.

\bibitem[Wu et~al.(2021)Wu, Baek, You, and Ma]{wu2021ratereduction}
Z.~Wu, C.~Baek, C.~You, and Y.~Ma, ``Incremental learning via rate reduction,''
  in \emph{Proceedings of the IEEE/CVF Conference on Computer Vision and
  Pattern Recognition (CVPR)}, 2021, pp. 1125--1133.

\end{thebibliography}


% Generated by IEEEtranN.bst, version: 1.14 (2015/08/26)
\begin{thebibliography}{9}
\providecommand{\natexlab}[1]{#1}
\providecommand{\url}[1]{#1}
\csname url@samestyle\endcsname
\providecommand{\newblock}{\relax}
\providecommand{\bibinfo}[2]{#2}
\providecommand{\BIBentrySTDinterwordspacing}{\spaceskip=0pt\relax}
\providecommand{\BIBentryALTinterwordstretchfactor}{4}
\providecommand{\BIBentryALTinterwordspacing}{\spaceskip=\fontdimen2\font plus
\BIBentryALTinterwordstretchfactor\fontdimen3\font minus
  \fontdimen4\font\relax}
\providecommand{\BIBforeignlanguage}[2]{{%
\expandafter\ifx\csname l@#1\endcsname\relax
\typeout{** WARNING: IEEEtranN.bst: No hyphenation pattern has been}%
\typeout{** loaded for the language `#1'. Using the pattern for}%
\typeout{** the default language instead.}%
\else
\language=\csname l@#1\endcsname
\fi
#2}}
\providecommand{\BIBdecl}{\relax}
\BIBdecl

\bibitem[Erden and Faltings(2026)]{erden2025foundations}
Z.~D. Erden and B.~Faltings, ``Foundations of a developmental design paradigm
  for integrated continual learning, deliberative behavior, and
  comprehensibility,'' \emph{IEEE Transactions on Emerging Topics in
  Computational Intelligence}, vol.~10, no.~2, pp. 1738--1752, 2026.

\bibitem[Erden(2025)]{erden2025thesis}
Z.~D. Erden, ``Foundations of a new learning paradigm in {AI} grounded in the
  principles of evolutionary developmental biology,'' Ph.D. dissertation, EPFL,
  2025.

\bibitem[Zenke et~al.(2017)Zenke, Poole, and Ganguli]{zenke2017si}
F.~Zenke, B.~Poole, and S.~Ganguli, ``Continual learning through synaptic
  intelligence,'' in \emph{Proceedings of the 34th International Conference on
  Machine Learning}, 2017, pp. 3987--3995.

\bibitem[Aljundi et~al.(2018)Aljundi, Babiloni, Elhoseiny, Rohrbach, and
  Tuytelaars]{aljundi2018mas}
R.~Aljundi, F.~Babiloni, M.~Elhoseiny, M.~Rohrbach, and T.~Tuytelaars, ``Memory
  aware synapses: Learning what (not) to forget,'' in \emph{Proceedings of the
  European Conference on Computer Vision (ECCV)}, 2018, pp. 139--154.

\bibitem[Li and Hoiem(2018)]{li2018lwf}
Z.~Li and D.~Hoiem, ``Learning without forgetting,'' \emph{IEEE Transactions on
  Pattern Analysis and Machine Intelligence}, vol.~40, no.~12, pp. 2935--2947,
  2018.

\bibitem[Buzzega et~al.(2020)Buzzega, Boschini, Porrello, Abati, and
  Calderara]{buzzega2020derpp}
P.~Buzzega, M.~Boschini, A.~Porrello, D.~Abati, and S.~Calderara, ``Dark
  experience for general continual learning: a strong, simple baseline,'' in
  \emph{Advances in Neural Information Processing Systems}, vol.~33, 2020, pp.
  15\,920--15\,930.

\bibitem[Caccia et~al.(2022)Caccia, Aljundi, Asadi, Tuytelaars, Pineau, and
  Belilovsky]{caccia2022erace}
L.~Caccia, R.~Aljundi, N.~Asadi, T.~Tuytelaars, J.~Pineau, and E.~Belilovsky,
  ``New insights on reducing abrupt representation change in online continual
  learning,'' in \emph{International Conference on Learning Representations},
  2022.

\bibitem[Erden et~al.(2025)Erden, Gasmi, and Faltings]{erden2025aec}
Z.~D. Erden, D.~Gasmi, and B.~Faltings, ``Continual reinforcement learning via
  autoencoder-driven task and new environment recognition,'' in \emph{Adaptive
  and Learning Agents (ALA) and Autonomous Robots and Multirobot Systems (ARMS)
  Workshops at the 24th International Conference on Autonomous Agents and
  Multiagent Systems (AAMAS)}, 2025.

\bibitem[LeCun et~al.(1998)LeCun, Bottou, Bengio, and
  Haffner]{lecun1998gradient}
Y.~LeCun, L.~Bottou, Y.~Bengio, and P.~Haffner, ``Gradient-based learning
  applied to document recognition,'' \emph{Proceedings of the IEEE}, vol.~86,
  no.~11, pp. 2278--2324, 1998.

\end{thebibliography}
\end{document}